\documentclass[10pt,twocolumn,letterpaper]{article}

\usepackage[pagenumbers]{cvpr} 

\usepackage[T1]{fontenc}  
\usepackage{amsthm}
\usepackage{multirow}
\usepackage{tikz}
\usepackage{bbm}
\usetikzlibrary{arrows.meta,positioning,calc,fit}
\usepackage{pgfplots}
\pgfplotsset{compat=1.18}
\usetikzlibrary{shapes.geometric,backgrounds}
\definecolor{cbred}{HTML}{D55E00}\definecolor{cbpurple}{HTML}{CC78BC}

\usepackage{bibunits}  
\usepackage{placeins}  

\newif\iffigspecs
\figspecsfalse

\AtEndPreamble{%
  \crefname{equation}{Eq.}{Eqs.}
  \Crefname{equation}{Equation}{Equations}
  \crefname{figure}{Fig.}{Figs.}
  \Crefname{figure}{Figure}{Figures}
  \crefname{table}{Tab.}{Tabs.}
  \Crefname{table}{Table}{Tables}
  \crefname{section}{Sec.}{Secs.}
  \Crefname{section}{Section}{Sections}
  \crefname{appendix}{Sec.}{Secs.}
  \Crefname{appendix}{Section}{Sections}
  \crefname{subappendix}{Sec.}{Secs.}
  \Crefname{subappendix}{Section}{Sections}
  \crefname{proposition}{Prop.}{Props.}
  \Crefname{proposition}{Proposition}{Propositions}
}
\newtheorem{proposition}{Proposition}

\newcommand{\method}{CalToR\xspace}                 
\newcommand{\methodlong}{Calibrated Token Reduction\xspace}
\newcommand{\IN}{IN\xspace}            
\newcommand{\INV}{IN-V2\xspace}        
\newcommand{\INR}{IN-R\xspace}         
\newcommand{\INA}{IN-A\xspace}         
\newcommand{\INS}{IN-S\xspace}         
\newcommand{\PPS}{PPS\xspace}          
\newcommand{\KL}{\mathrm{KL}}
\newcommand{\pp}{\,pp\xspace}          
\newcommand{\gh}{GH200\xspace}
\newcommand{\ahun}{A100\xspace}
\newcommand{\sig}{$^\dagger$}          

\definecolor{cvprblue}{rgb}{0.21,0.49,0.74}
\definecolor{cbblue}{HTML}{0173B2}\definecolor{cborange}{HTML}{DE8F05}\definecolor{cbgreen}{HTML}{029E73}
\usepackage[pagebackref,breaklinks,colorlinks,allcolors=cvprblue]{hyperref}
\newcommand{\appxtag}{App.}          
\newcommand{\appxtags}{Apps.}        
\newcommand{\Appxtag}{Appendix}      
\newcommand{\Appxtags}{Appendices}   
\newcommand{\maintag}{Main}          
\newcommand{\Maintag}{Main}          
\newcommand{\appxsecstyle}{\arabic}  
\newif\ifappxrestart
\appxrestarttrue
\newif\ifappxSnumbers
\appxSnumbersfalse

\newif\ifinappendix
\protected\def\toappx{\ifinappendix\else\appxtag~\fi}
\protected\def\Toappx{\ifinappendix\else\Appxtag~\fi}
\protected\def\tomain{\ifinappendix\maintag~\fi}
\protected\def\Tomain{\ifinappendix\Maintag~\fi}
\protected\def\appxsec{\ifinappendix Sec.\else\appxtag\fi}
\protected\def\appxsecs{\ifinappendix Secs.\else\appxtags\fi}
\protected\def\Appxsec{\ifinappendix Section\else\Appxtag\fi}
\protected\def\Appxsecs{\ifinappendix Sections\else\Appxtags\fi}

\makeatletter
\newcommand{\appendixrefs}{%
  \global\inappendixtrue
  \renewcommand{\thesection}{\appxsecstyle{section}}
  \renewcommand{\thepage}{\texorpdfstring{\arabic{page}}{S\arabic{page}}}%
  \ifappxSnumbers\gdef\appx@numprefix{S}\else\global\let\appx@numprefix\@empty\fi
  \appx@renumber{figure}\appx@renumber{table}\appx@renumber{equation}\appx@renumber{algocf}%
  \appx@renumber{theorem}\appx@renumber{proposition}\appx@renumber{lemma}\appx@renumber{corollary}%
  \crefalias{section}{appendix}%
  \crefalias{subsection}{subappendix}%
  \crefalias{subsubsection}{subsubappendix}%
  \crefalias{figure}{appfigure}%
  \crefalias{table}{apptable}%
  \crefalias{equation}{appequation}%
  \crefalias{algocf}{appalgorithm}
  \crefalias{algorithm}{appalgorithm}
  \crefalias{theorem}{apptheorem}%
  \crefalias{proposition}{appproposition}%
  \crefalias{lemma}{applemma}%
  \crefalias{corollary}{appcorollary}}
\newcommand{\appx@renumber}[1]{%
  \@ifundefined{c@#1}{}{%
    \ifappxrestart
      \setcounter{#1}{0}%
      \@namedef{the#1}{\appx@numprefix\arabic{#1}}%
      \@namedef{theH#1}{S\arabic{#1}}%
    \fi}}

\newcommand{\appx@names}[6]{%
  \crefname{#1}{\tomain #3}{\tomain #4}\Crefname{#1}{\Tomain #5}{\Tomain #6}%
  \crefname{#2}{\toappx #3}{\toappx #4}\Crefname{#2}{\Toappx #5}{\Toappx #6}}
\AtEndPreamble{%
  \appx@names{section}{appendix}{Sec.}{Secs.}{Section}{Sections}%
  \appx@names{figure}{appfigure}{Fig.}{Figs.}{Figure}{Figures}%
  \appx@names{table}{apptable}{Tab.}{Tabs.}{Table}{Tables}%
  \appx@names{equation}{appequation}{Eq.}{Eqs.}{Equation}{Equations}%
  \appx@names{algorithm}{appalgorithm}{Alg.}{Algs.}{Algorithm}{Algorithms}
  \appx@names{theorem}{apptheorem}{Thm.}{Thms.}{Theorem}{Theorems}%
  \appx@names{proposition}{appproposition}{Prop.}{Props.}{Proposition}{Propositions}%
  \appx@names{lemma}{applemma}{Lem.}{Lems.}{Lemma}{Lemmas}%
  \appx@names{corollary}{appcorollary}{Cor.}{Cors.}{Corollary}{Corollaries}%
  \crefname{appendix}{\appxsec}{\appxsecs}\Crefname{appendix}{\Appxsec}{\Appxsecs}%
  \creflabelformat{appequation}{\textup{(#2#1#3)}}%
}
\makeatother

\makeatletter
\let\appx@newlabel\newlabel
\def\newlabel#1#2{\appx@newlabel{#1}{#2}\appx@newlabel{S-#1}{#2}\appx@newlabel{M-#1}{#2}}
\patchcmd{\@startbibunit}{\@input{\@bibunitname.aux}}
  {\makeatletter\let\@writefile\@gobbletwo\@input{\@bibunitname.aux}}{}{\@latex@warning@no@line{Patch of bibunits failed}}
\patchcmd{\@finishstartedbibunit}{\@input{\@bibunitname.aux}}
  {\makeatletter\let\@writefile\@gobbletwo\@input{\@bibunitname.aux}}{}{\@latex@warning@no@line{Patch of bibunits failed}}
\newcommand{\appx@bibunithooks}{%
  \def\hyper@natlinkstart##1{%
    {\let\@auxout\@mainaux\Hy@backout{S.##1}}%
    \hyper@linkstart{cite}{cite.S.##1}%
    \def\hyper@nat@current{##1}}%
  \def\hyper@natlinkbreak##1##2{\hyper@linkend##1\hyper@linkstart{cite}{cite.S.##2}}%
  \def\hyper@natanchorstart##1{\Hy@raisedlink{\hyper@anchorstart{cite.S.##1}}}%
  \let\appx@BR@backref\BR@backref
  \def\BR@backref##1{\appx@BR@backref{S.##1}}%
  \long\def\backrefxxx##1##2##3{\hyperlink{page.S##1}{##1}}}
\AtBeginEnvironment{bibunit}{\appx@bibunithooks}
\makeatother

\def\confName{CVPR}
\def\confYear{2027}

\title{Calibrate to the Clock, Not to FLOPs:\\Token Reduction Schedules for Vision Transformers}

\author{
\textbf{Magdalena Tr\k{e}dowicz}$^{1,2}$ \\
\texttt{\small magdalena.tredowicz@doctoral.uj.edu.pl} \\
\and
\textbf{Jacek Tabor}$^{1}$ \\
\texttt{\small jacek.tabor@uj.edu.pl} \\
\and
\textbf{\L{}ukasz Struski}$^{1}$ \\
\texttt{\small lukasz.struski@uj.edu.pl} \\
\\[-0.5em]
\parbox{\linewidth}{\small\raggedright
$\blacktriangleright\,^{1}$\,Faculty of Mathematics and Computer Science, Jagiellonian University, Poland; \quad
$\blacktriangleright\,^{2}$\,Doctoral School of Exact and Natural Sciences, Jagiellonian University, Poland 
}
\\[1em]
}

\begin{document}

\twocolumn[{%
\renewcommand\twocolumn[1][]{#1}%
\maketitle
\vspace{-1.3cm}
\begin{center}
  \captionsetup{type=figure}
  \centering
  \includegraphics[width=\textwidth]{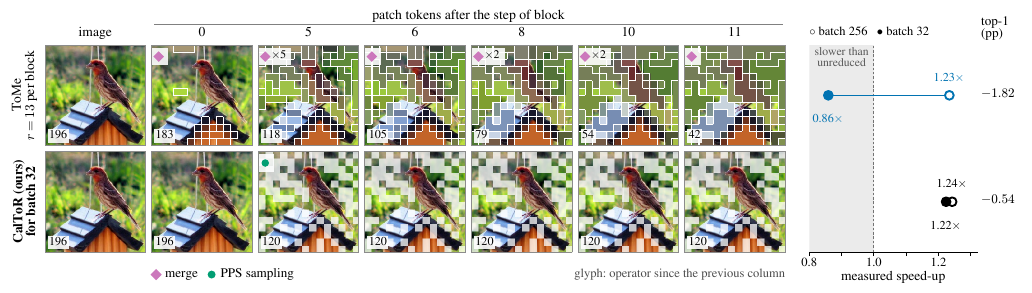}
  \captionof{figure}{\textbf{\method turns token reduction into a hardware-aware scheduling problem.}
A fixed ToMe policy ($r{=}13$) reduces tokens early and uniformly, but runs
\emph{slower} than the unreduced ViT ($0.86\times$). \method keeps early blocks
intact, then applies calibrated merges and PPS sampling later, reaching
$1.22\times$--$1.24\times$ speed-up with only $0.54$\,pp accuracy drop.
The result: a static, deployment-ready schedule that meets the throughput
target without sacrificing accuracy.}
  \label{fig:teaser}
\end{center}
\vspace{1em}
}]

\begin{abstract}
Vision Transformers (ViTs) are commonly sped up by removing tokens at inference. Most methods fix the policy in advance, for example by merging a constant number of tokens per block, and ignore what the reduction costs on the target hardware. A reduction that lowers FLOPs can therefore fail to cut wall-clock latency, especially at small batch sizes or in late blocks, where computing statistics, matching, sorting and gathering can take longer than the time saved. We present \method, which designs token-reduction schedules for frozen ViTs. Given a model, a GPU, a batch size and a throughput target, \method chooses which operator to apply, in which block, and how many tokens to remove. It first measures a latency with the fixed and per-token cost of each operator, then measures the damage each reduction does to the predictions of the unreduced model. A greedy Lagrangian allocation builds candidate schedules under the time budget, and a refinement stage improves the best ones. Where prior methods rely on a single operator, \method searches jointly over merging, pruning and proportional-probability sampling and models how their effects interact. It returns one static schedule per GPU and batch size. At equal measured throughput on a GH200 at batch 256, \method gains a median of 1.05\pp top-1 over ToMe and 1.70\pp over PiToMe, without retraining or fine-tuning.
\end{abstract}
\vspace{-0.1cm}

\section{Introduction}
\label{sec:intro}

Vision Transformers (ViTs) have become a cornerstone of modern computer
vision, but their computational cost remains a barrier to deployment at
scale. A natural way to reduce this cost is to remove tokens during the
forward pass: merge similar patches, prune less informative ones, or
sample a subset of the input. This idea, known as token reduction, has
driven a rich line of work~\cite{tome,evit,dynamicvit,pitome,tofu},
promising faster inference with minimal loss in accuracy.

The promise, however, rests on a hidden assumption that removing tokens
makes the model faster. On a GPU, this assumption often fails. Deciding what to remove is itself expensive. The model must compute statistics, match or sort tokens, and gather those that remain. When the
saved computation is smaller than this overhead, the reduced model can be slower than the unreduced one. On a GH200 at batch 32, 54 of 66 competitor configurations on DeiT-S are slower than the model they are meant to speed up (\cref{S-fig:supp:macs}).

Existing token-reduction methods rarely confront this gap. Most fix the operator, merging or pruning, and set the reduction amount by hand, or
optimize it under a FLOPs budget~\cite{diffrate}. Such a budget ignores
the fixed cost of each reduction step, which dominates at small batch
sizes and on smaller models.

We argue that token reduction should be treated as a resource-allocation problem under a measured-time budget, rather than as the design of a single reduction operator or a fixed token budget. Given a frozen ViT, a target GPU, a batch size, and a throughput target, \method automatically determines how to distribute token reductions across operators and blocks to minimize changes in the model's predictions while meeting the target.

While post-training pruning has used measured timings and small calibration sets to meet speed-up targets~\cite{spdy,kwon2022,ziplm}, token reduction poses additional challenges. Removing tokens saves computation not only in the current block but also in all subsequent blocks; at the same time, each reduction incurs a fixed overhead, and reductions across blocks can interact. \method accounts for these effects when allocating reductions, using only a few minutes of calibration on a single GPU, with no labels or weight updates. As \Cref{fig:teaser} illustrates for DeiT-B at batch size 32, optimizing for measured time rather than MACs also matters for predictive quality: MAC-based allocation incurs an additional 0.10--2.10\,pp of accuracy loss. Our contributions are as follows.
\begin{itemize}
  \item We show that MACs mislead token reduction on GPUs. Every reduction step has a fixed cost, so it pays off only above a break-even size that depends on depth and batch size.
  \item We propose \method, which allocates a budget of measured time across operators and blocks. It chooses merging, pruning or sampling for each block from measured latency and KL damage, accounts for how neighboring reductions interact, and refines the result on the true KL.  
  \item We establish gradient-free calibration as an effective alternative to existing training-free methods, maching or outperforming ten method families across diverse settings with only minutes of calibration on a single GPU.
\end{itemize}

\section{Related Work}
\label{sec:related}

\paragraph{Training-free token merging and pruning.}
Training-free methods differ mostly in what they do to a token. ToMe merges similar tokens by bipartite matching, with
a fixed number of merges per block~\cite{tome}. PiToMe protects informative tokens with an energy score~\cite{pitome},
and ToFu mixes merging with pruning~\cite{tofu}. PPT chooses, per image and layer, between pruning inattentive tokens
and pooling similar ones~\cite{ppt}. ATS samples tokens by the inverse transform of their significance~\cite{ats}. More recent work includes ``Later Is Better''~\cite{lib} and Prune-and-Merge~\cite{pruneandmerge},
which combines pruning by CLS attention with similarity-based merging. Our menu has both operators, but Prune-and-Merge
decides in advance where they act and how many tokens they remove. Trained methods fine-tune the network together with
the reduction~\cite{dynamicvit,evit,ltmp}.
Despite their differences, these methods make the same two choices. They fix their operators by design, or switch
between them with an input-adaptive rule, and they set the amounts by hand or for a FLOPs target. ToMe goes one step
further and compares a constant schedule with 15\,000 random ones on measured throughput. None of them recalibrates the
schedule for a new GPU or batch size.

\paragraph{Learned and searched rates.}
A second line of work learns or searches the amounts. DiffRate searches per-block prune and merge rates with gradients
under a FLOPs constraint, which takes 2.7 GPU-hours for ViT-B, and a variant adds the latency and power losses of a
simulated accelerator~\cite{diffrate}. TAP learns an exact per-layer budget with a trained selector~\cite{tap}, and
SPViT and HeatViT train selectors for the latency of mobile devices and FPGAs~\cite{spvit,heatvit}.

\paragraph{Latency and allocation under a budget.}
The closest work to ours is POMT~\cite{pomt}. It prunes once, at a quarter of the depth, and picks the amount by a
latency-accuracy grid search on edge devices and an A100. Other works agree that the speed-up does not follow from
FLOPs. It depends on the batch size~\cite{mergeover}, on kernel dispatch costs~\cite{raggedattn} and on the
near-constant cost of matching~\cite{toast}. Sorting and scattered writes can cancel the gain of merging on a
GPU~\cite{toma,mame,mpm}, and FLOPs and measured speed can disagree~\cite{misnomer}. Our allocation draws on compression. Spending a budget with Lagrange multipliers over per-item damage curves is
classical~\cite{everett1963,shoham1988}. Post-training pruning meets a speed-up target from per-layer
timings and a small calibration set. SPDY does so by dynamic programming and search~\cite{spdy}, Kwon et al.\ fit a
piecewise-linear latency with a constant overhead to a lookup table~\cite{kwon2022}, and ZipLM uses runtime tables of the
target environment~\cite{ziplm}.

\begin{figure*}[t]
  \centering
  \includegraphics[width=\textwidth]{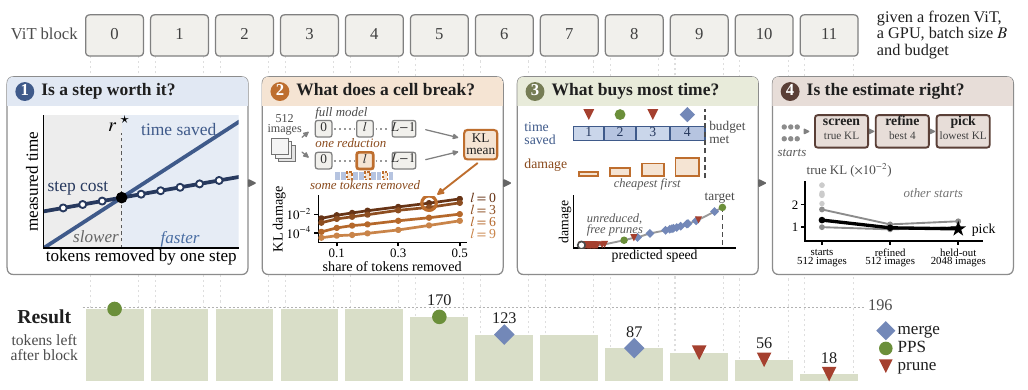}
 \caption{\textbf{Overview of \method.} Given a frozen ViT, a target GPU, batch size, and throughput budget, \method constructs a token reduction schedule in four stages. (1) Break-even analysis measures the latency of reduction steps to determine when token savings outweigh their execution cost. (2) Damage estimation measures the KL divergence caused by individual reductions across operators and blocks. (3) Schedule allocation greedily combines these measurements to prioritize reductions with the best trade-off between saved time and predicted damage, generating candidate schedules. (4) Refinement and selection evaluate the best candidates using the true KL divergence and select the final schedule on held-out images. The bottom illustrates the resulting schedule, showing the number of tokens remaining after each block and the selected reduction operators.}
  \label{fig:pipeline}
\end{figure*}

\section{\method: \methodlong}
\label{sec:method}

Reducing the number of tokens does not necessarily lower inference latency. The speedup depends on where tokens are removed, which operator is used, and how its cost interacts with the execution of the remaining network. Token reduction is therefore a hardware-dependent allocation problem. We must decide how many tokens to remove, where, and with which operator, while minimizing the change in the model's predictions.

\method first measures the runtime cost of each operator on the target hardware. It then estimates the distortion of reductions at different network locations, and finally combines both to construct and refine a schedule under a prescribed runtime budget.

\subsection{Problem Formulation}
\label{sec:problem}
Let a frozen ViT consist of $L$ transformer blocks. We consider a set of token reduction operators $\mathcal{O}$, such as token merging and token pruning. A reduction schedule specifies the number of tokens removed by each operator at each block. We represent a schedule as
$
s =~\{r_{o,l}\}_{o\in\mathcal{O},\,l\in\{0,\ldots,L-1\}},
$
where $r_{o,l}$ is the number of tokens removed by operator $o$ at block $l$. We refer to each operator--block pair $(o,l)$ as a \emph{cell}. A cell is active if $r_{o,l}>0$ and inactive otherwise. Given a target throughput improvement $\theta$, our objective is to find a schedule that minimizes the distortion relative to the original model while satisfying the corresponding runtime budget:
\begin{equation}
\begin{aligned}
\min_s \quad & \mathcal{K}(s) \qquad
\text{s.t.} \quad & T(s) \leq \frac{1}{\theta},
\end{aligned}
\label{eq:objective}
\end{equation}
where $\mathcal{K}(s)$ denotes the output distortion induced by the schedule and $T(s)$ is its predicted runtime, normalized by the runtime of the original model.

We use the KL divergence between the output distributions of the original and reduced models as our distortion measure. The original ViT remains frozen throughout the procedure: calibration data are used only to estimate distortion and construct the schedule, not to update model parameters.

\subsection{Hardware-Aware Latency Modeling}
\label{sec:latency}

The first component of \method is a latency model that captures the cost of executing token reduction on a particular hardware configuration. We profile each reduction operator at different transformer blocks and token counts, using the target GPU and batch size.

The latency of a reduction operator is not generally proportional to the number of removed tokens. In addition to the per-token processing cost, an operator incurs a fixed overhead associated with launching and executing its reduction procedure. We therefore model the cost of activating operator $o$ at block $l$ as
\begin{equation}
C_{o,l}(r_{o,l})
=
\kappa^0_o(n_l)\mathbbm{1}_{[r_{o,l}>0]}
+
\kappa^1_o(n_l)r_{o,l},
\label{eq:operator_cost}
\end{equation}
where $n_l$ is the number of tokens entering block $l$, $\kappa^0_o$ is the fixed activation cost, and $\kappa^1_o$ is the marginal cost per removed token. The indicator is important: the fixed cost is paid once for an active cell, independently of the number of tokens removed by that cell. Inactive cells incur no operator cost. The predicted runtime of a schedule is
\begin{equation}
\begin{aligned}
T(s) ={}&
t_{\mathrm{emb}}
+
\sum_{l=0}^{L-1}\tau_l(n_l)
+
t_{\mathrm{head}}\\
&+
\sum_{o\in\mathcal{O}}\sum_{l=0}^{L-1}
\left[
\kappa^0_o(n_l)\mathbbm{1}_{[r_{o,l}>0]}
+
\kappa^1_o(n_l)r_{o,l}
\right],
\end{aligned}
\label{eq:latency_model}
\end{equation}
where $t_{\mathrm{emb}}$ and $t_{\mathrm{head}}$ denote the embedding and prediction-head costs, respectively, and $\tau_l(n_l)$ captures the cost of transformer block $l$ at its current token count. The token count $n_l$ is determined by the reductions applied before block $l$.

We measure the latency of the transformer blocks and reduction operators separately and fit the corresponding cost models. The accounting convention assigns the cost of computing an operator's importance statistics to the relevant block execution, rather than counting it again as an independent operator cost. This avoids double-counting work shared with the block computation.

\paragraph{Break-even condition.}
The fixed activation cost explains why reducing more tokens in a cell can be substantially more efficient than activating several cells with small reductions. To make this effect explicit, consider a reduction at block $l$ that removes $r$ tokens and changes the token count of subsequent blocks. Under a locally linear approximation of block latency, let
\begin{equation}
S_l =
\partial_n \tau_l^{\mathrm{mlp}}
+
\sum_{l'>l}\partial_n\tau_{l'}
\label{eq:saving_slope}
\end{equation}
denote the total marginal runtime saving associated with removing one token at block $l$. Here, $\tau_l^{\mathrm{mlp}}$ captures the token-dependent MLP cost at the reduction block, while the remaining terms account for the effect on subsequent blocks. The fixed-cost term also accounts for any additional execution cost induced in later blocks by the reduction mechanism, where applicable.

\begin{proposition}[Break-even condition]
\label{prop:threshold}
Assume that the relevant block latencies are locally linear in the token count. A reduction of $r$ tokens by operator $o$ at block $l$ decreases the predicted runtime if and only if
\begin{equation}
S_l > \kappa^1_o
\qquad\text{and}\qquad
r > \frac{\kappa^0_o}{S_l-\kappa^1_o}.
\label{eq:break_even}
\end{equation}
\end{proposition}

\begin{proof}
Removing $r$ tokens saves $rS_l$ in the transformer blocks and costs $\kappa_o^0+r\kappa_o^1$ in the operator, so $\Delta T=\kappa_o^0-r(S_l-\kappa_o^1)$. This is negative exactly when $r(S_l-\kappa_o^1)>\kappa_o^0$. Since $\kappa_o^0>0$, this requires $S_l>\kappa_o^1$ and then gives $r>\kappa_o^0/(S_l-\kappa_o^1)$.
\end{proof}

The condition shows that the usefulness of a reduction depends on its location, operator, and magnitude. In particular, a reduction may be beneficial in an early block but unprofitable in a later one, even if it removes the same number of tokens. We use it to avoid spending the latency budget on reductions without a net runtime benefit.

\subsection{Estimating Single-Cell Distortion}
\label{sec:distortion}

We next estimate the distortion caused by applying a reduction operator at a particular block. For each cell $(o,l)$, we measure the distortion induced by removing different numbers of tokens while leaving the remaining schedule unchanged. This produces a distortion curve $D_{o,l}(r),$ which approximates the output KL divergence associated with removing $r$ tokens using operator $o$ at block $l$.

The curves are obtained using a calibration subset and the frozen reference model. For image classification, we use 512 calibration images sampled from the training subset, with 10 images per class. No labels are required for this image calibration: the reference model's output distribution provides the target for measuring KL divergence. For video, we use 1024 calibration clips (\cref{S-sec:supp:impl}). The resulting curves capture the fact that the same reduction can have different effects depending on both its operator and its position in the network. However, individual curves alone do not determine the distortion of a schedule: reductions at different cells can interact, particularly when they occur in neighboring blocks.

\subsection{Schedule Construction}
\label{sec:schedule}

\paragraph{Modeling interactions between reductions.}

To account for interactions between cells, we approximate the distortion of a complete schedule using a partially coherent model:
\begin{equation}
\hat{\mathcal{K}}(s)
=
(1-c)\sum_{o,l}D_{o,l}(r_{o,l})
+
c\left(
\sum_{o,l}\sqrt{D_{o,l}(r_{o,l})}
\right)^2,
\label{eq:coherent}
\end{equation}
where $c\in[0,1]$ controls the assumed degree of interaction between individual distortions. At $c=0$, the model reduces to an additive approximation. At $c=1$, the square roots of the individual distortions accumulate before being squared, corresponding to a setting in which errors reinforce one another. Intermediate values interpolate between these two regimes. This model provides an inexpensive estimate of schedule distortion and allows us to compare candidate allocation changes without repeatedly evaluating the full model on the calibration set.

\paragraph{Greedy allocation.}
Given the latency model and the damage curves, we start from the unreduced model and repeatedly increase the reduction of one operator--block cell. For a candidate change $a$ applied to the current schedule $s$, we select the feasible change that minimizes the predicted distortion per saved time,
\begin{equation}
\frac{\hat{\mathcal{K}}(s+a)-\hat{\mathcal{K}}(s)}
{T(s)-T(s+a)},
\label{eq:greedy}
\end{equation}
and discard changes that do not reduce predicted runtime. We repeat this until the predicted runtime satisfies the target budget $T(s)\leq 1/\theta$. The runtime saving is evaluated with respect to the current schedule. Activating an inactive cell therefore incurs its fixed cost $\kappa^0_o$, whereas further reductions within an already active cell do not, which makes the criterion sensitive to the amortization of operator overhead.

The candidate magnitudes are 1, 2, 4, or 8 times 2.5\% of the tokens entering the block, which accounts for the break-even behavior of \cref{prop:threshold}. The fixed activation costs turn the allocation into a knapsack problem with setup costs, so a single greedy run may miss useful schedules. We therefore build several starting schedules with different operator sets and coherence values (\cref{S-sec:supp:impl}).

\subsection{Schedule Refinement and Selection}
\label{sec:refinement}

The coherence model in \cref{eq:coherent} is an approximation and may not rank complete schedules accurately. We therefore refine the starting schedules on the true KL divergence of the calibration data. We fit the coherence parameter $c$ to the measured KL of the starts on the 512 calibration images, add the start built with the fitted $c$, and keep the four with the lowest measured KL.

Each retained schedule is refined by local moves. A move transfers tokens between two operator--block cells at equal predicted runtime, or empties the source cell entirely and thereby removes its fixed cost. In each iteration we rank eight random moves with \cref{eq:coherent}, evaluate the most promising one on the true KL, and accept it only if the measured KL decreases (60 iterations on images, 100 on video; move sizes in ~\cref{S-sec:supp:impl}).

Finally, we select the best refined schedule on 2048 additional training images, held out from construction and refinement, which reduces the risk of overfitting the calibration set. The resulting schedule is fixed for a given model, GPU, batch size, and throughput target.

\subsection{Token Reduction Operators}
\label{sec:operators}

\method supports different token reduction operators within a common scheduling framework. Each operator is characterized by its token-selection or token-matching rule, its single-cell distortion curve, and its measured hardware cost. This separation allows the schedule optimizer to choose both the reduction location and the operator without assuming that all reductions have the same computational behavior.

\paragraph{Token merging.}

For token merging, we identify pairs of similar tokens and combine them into a single representative. The matching cost accounts for both feature similarity and token importance:
\begin{equation}
d(i,j)
=
\bigl(1-\cos(v_i,v_j)\bigr)
\sqrt{u_i+u_j},
\label{eq:merge_metric}
\end{equation}
where $v_i$ and $v_j$ are token features and $u_i$ and $u_j$ are their importance values. This weighting makes dissimilarity more consequential for pairs with greater combined importance.

After merging, we preserve the token sizes, rescale the merged feature to the size-weighted mean norm and adjust the attention contributions proportionally to the represented sizes.

\paragraph{Importance-based pruning.}

For pruning, we estimate token importance and retain a subset of tokens according to their scores. To avoid concentrating the retained tokens in a small spatial region, we use probability-proportional-to-size sampling with a uniform component. Given importance scores $z_j$ for $n$ tokens, we define
\begin{equation}
q_j
=
\frac{\alpha}{n}
+
(1-\alpha)\operatorname{softmax}(z)_j,
\label{eq:pps_distribution}
\end{equation}
where $\alpha$ controls the uniform component. We choose inclusion probabilities
\begin{equation}
\pi_j=\min(1,\lambda q_j),
\qquad
\sum_{j=1}^{n}\pi_j=k,
\label{eq:inclusion_prob}
\end{equation}
where $k$ is the number of tokens to retain and $\lambda$ is chosen to satisfy the prescribed sum.

We use systematic sampling to select tokens with these inclusion probabilities. Tokens are ordered along a Morton traversal, and the sampling offset is derived deterministically from the importance scores.

\paragraph{Video-specific reduction.}

For video models, we additionally consider temporal information when selecting tokens. Before the first transformer block, we estimate temporal change from the input tubelets or spatial-temporal grid and use it as an importance signal. This allows the reduction procedure to preserve tokens associated with greater temporal variation. The operator set for video also includes token-norm, received-attention, grid-based, random, and temporal-change selection rules. The same hardware-aware schedule construction and refinement procedure is used to allocate reductions across these operators and transformer blocks.

\begin{table*}[ht!]
  \centering
  \footnotesize
  \caption{\method (ours) against ten competitor families at equal measured throughput (nine on the \ahun, without
    Prune-and-Merge, and on CLIP, without DiffRate). Each cell gives the number of pairs in which \method is more
    accurate than the competitor. In brackets: the significant wins (bold) and the significant losses of \method after
    Holm correction. Then the median paired difference $\Delta$ = top-1 of \method $-$ top-1 of the competitor (pp).}
  \label{tab:main}
  \begin{tabular}{@{}llccccc@{}}
    \toprule
    & & \multicolumn{5}{c}{\method (ours) against each competitor: better (wins/losses), median $\Delta$ (pp)} \\
    \cmidrule(l){3-7}
    setting & pairs & \IN & \INV & \INR & \INA & \INS \\
    \midrule
    \gh, batch 256, 5 models & 314 & 304 (\textbf{267}/1) $+0.96$ & 303 (\textbf{217}/0) $+1.31$ & 305 (\textbf{294}/1) $+1.93$ & 279 (\textbf{241}/6) $+2.89$ & 298 (\textbf{273}/7) $+1.26$ \\
    \gh, batch 32, 5 models & 321 & 310 (\textbf{266}/2) $+1.03$ & 311 (\textbf{235}/0) $+1.57$ & 311 (\textbf{296}/0) $+2.07$ & 295 (\textbf{248}/5) $+3.36$ & 304 (\textbf{278}/8) $+1.32$ \\
    \ahun, batch 256, 5 models & 279 & 278 (\textbf{241}/0) $+0.92$ & 276 (\textbf{195}/0) $+1.34$ & 273 (\textbf{269}/1) $+1.97$ & 263 (\textbf{232}/3) $+3.51$ & 267 (\textbf{251}/6) $+1.31$ \\
    \gh, batch 256, CLIP & 125 & 125 (\textbf{123}/0) $+0.91$ & 125 (\textbf{111}/0) $+1.36$ & 122 (\textbf{122}/3) $+2.15$ & 117 (\textbf{103}/3) $+3.23$ & 123 (\textbf{115}/1) $+1.52$ \\
    \bottomrule
  \end{tabular}
\end{table*}
\section{Experimental Setup}
\label{sec:exp:setup}
\paragraph{Models and data.} We use DeiT-S and DeiT-B~\cite{deit}, MAE ViT-B, ViT-L and ViT-H~\cite{mae} and CLIP
ViT-B/16 and ViT-L/14~\cite{clip}, all fine-tuned on ImageNet-1k. We test on ImageNet val (\IN) and on four shifted test sets, ImageNet-V2 (\INV)~\cite{imagenetv2}, ImageNet-R
(\INR)~\cite{imagenetr}, ImageNet-A (\INA)~\cite{imageneta} and ImageNet-Sketch (\INS)~\cite{imagenetsketch}. For video
we use VideoMAE-B on Kinetics-400 (K400) and Something-Something~v2 (SSv2), and VideoMAE-L on
K400~\cite{videomae}, on the full validation sets with the standard protocol.
\vspace{-0.2cm}
\paragraph{Throughput.}
We measure images (clips) per second on a \gh at batch 256 and 32, on an \ahun at batch 256, and for video on a \gh at
batch 8. Each configuration is measured in one job together with the unreduced model and our schedules. The speed-up
within a job is far more stable than the speed between jobs, so we pair methods inside one job
(\cref{S-sec:supp:impl}).
\vspace{-0.2cm}
\paragraph{Competitors.}
We compare against ToMe~\cite{tome}, ``Later Is Better''~\cite{lib}, DiffRate~\cite{diffrate}, a single grid prune, PaPr~\cite{papr},
PiToMe~\cite{pitome}, ToFu~\cite{tofu}, PPT~\cite{ppt}, POMT~\cite{pomt} and Prune-and-Merge~\cite{pruneandmerge}. We ported each method into our engine and kept the arithmetic of the official code
(\cref{S-sec:supp:fidelity}). For video we compare against RLT~\cite{rlt}, PaPr, STA~\cite{sta} and ToMe. Methods
without public code are compared by their published numbers (\cref{S-tab:published}).
\vspace{-0.2cm}
\paragraph{Pairing and statistics.}
For each competitor configuration, we take our slowest schedule that is at least as fast in the same job, if one
exists. We report the paired top-1 difference and test it with McNemar's exact test over the images. Holm's correction
over all pairs of one setting and test set keeps the family-wise error at 5\%. To cover competitors that no schedule of
ours matches closely, we add calibration targets at 1.05 times their measured speed. Competitor configurations that
are faster than all our schedules stay unpaired. They are few, and all of them lose 1.5--51\pp of top-1
(\cref{S-sec:supp:models}).

\vspace{-0.2cm}

\section{Results}
\label{sec:exp:main}

\Cref{tab:main} provides a comprehensive pairwise comparison of
\method against the evaluated baselines. On a \gh at batch size 256,
\method achieves higher accuracy on \IN in 304 of 314 comparisons,
with 267 improvements remaining significant after Holm correction.
The only significant loss is against Prune-and-Merge at keep rate
$K=0.95$ on MAE ViT-B; notably, this configuration is 12\% slower
than the unreduced model. The remaining nine losses are small,
each within 0.18\pp. The median accuracy gain is $+0.96$\pp on
\IN and increases under distribution shift, reaching $+1.31$\pp
on \INV, $+1.93$\pp on \INR, and $+2.89$\pp on \INA.

Of the 314 comparisons, 38 involve baselines that are slower than
the unreduced model. Restricting the comparison to the 276 faster
configurations, \method is more accurate in 270 cases, 239
significantly, with no significant loss. The gains are therefore not
driven by slow baselines.

Pairwise counts, however, treat all competitors equally, regardless
of how strong they are. To account for this, \cref{tab:envelope}
compares \method against a per-model competitor envelope. We define
the envelope using configurations that are faster than the
unreduced model and are not dominated by another competitor
configuration that achieves at least as high accuracy on \IN
at equal or greater measured speed.

The resulting gains vary across model families. At batch size 256,
\method improves median accuracy over the envelope by $+0.83$ to
$+1.51$\pp on DeiT and by $+0.45$ to $+0.77$\pp on CLIP. The gains
are smaller on MAE, ranging from $+0.04$ to $+0.14$\pp, where the
envelope is largely determined by grid pruning. Importantly,
\method incurs no statistically significant accuracy loss on
\IN against the envelope in any setting. On the more challenging
\INA benchmark, its advantage over the MAE envelope increases to
$+2.4$--$+3.4$\pp.

\begin{figure*}[t]
  \centering
  \includegraphics[width=\textwidth]{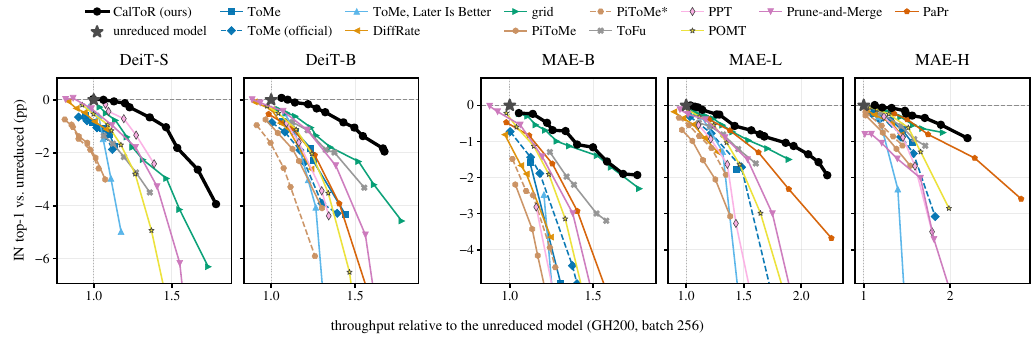}
  \caption{Top-1 on \IN relative to the unreduced model against the measured throughput relative to it (\gh, batch
    256). The two DeiT panels share one y axis, and so do the three MAE panels. Black, thick: \method (ours), the front
    of its schedules. Each competitor family is drawn through its operating points.}
  \label{fig:fronts}
  \vspace{-0.1cm}
\end{figure*}

\Cref{fig:fronts} provides a complementary view of the accuracy-throughput trade-off. Across the speed range covered by
\method, no competitor configuration achieves higher accuracy
than \method at a comparable measured speed.

\begin{figure}[t]
  \centering
  \includegraphics[width=\linewidth, trim=0.45cm 0 0 0, clip]{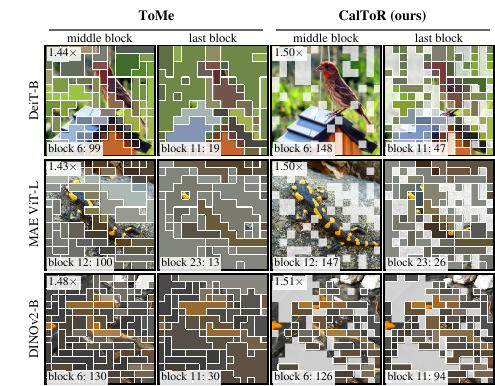}
  \caption{Tokens entering the middle and the last block under ToMe and \method (ours) at about the same measured
    speed (\gh, batch 256; speed-up at the top left, block and patch tokens at the bottom left), on one held-out
    training image per model. A merged token is drawn in the mean color of its patches and outlined; dropped patches
    are faded. On DINOv2, ToMe misclassifies this image and \method does not; on the other two images both are
    correct. More images: ~\cref{S-fig:supp:tokens,S-fig:supp:tokensL,S-fig:supp:tokensD}.}
  \label{fig:tokens}
\end{figure}
On the fronts at measured speed-ups of $1.25\times$ and $1.5\times$, \method is the most accurate method on every model
(\cref{S-tab:equalspeed}). Only three (DeiT-S) and five (DeiT-B) competitor variants reach $1.5\times$ at all.
Consider DeiT-B, with \method at $1.50\times$ and ToMe at $1.44\times$. \method keeps 148 tokens entering block~6, against 99
for ToMe, and 47 against 19 in the last block (\cref{fig:tokens}; \cref{S-fig:supp:tokens}). It samples once early
and reduces late, where removing a token does less damage. \Cref{fig:tokens} also shows MAE ViT-L and DINOv2 ViT-B/14
(\cref{S-tab:supp:dinov2}) at about the same speeds.

\begin{table}[t]
  \centering
  \footnotesize
  \setlength{\tabcolsep}{4pt}
  \caption{\method (ours) against the competitor envelope: the competitor configurations faster than the unreduced
    model that no other competitor configuration beats on \IN at the same or a higher measured speed, paired as in
    \cref{tab:main}. Median paired difference on \IN (pp). In brackets: significant wins after Holm correction / paired
    envelope configurations. There is no significant loss.}
  \label{tab:envelope}
  \begin{tabular}{@{}lccc@{}}
    \toprule
    model & \gh, 256 & \gh, 32 & \ahun, 256 \\
    \midrule
    DeiT-S & $+1.51$ (10/11) & $+1.13$ (7/8) & $+0.64$ (5/6) \\
    DeiT-B & $+0.83$ (10/10) & $+0.62$ (8/8) & $+0.81$ (12/12) \\
    MAE ViT-B & $+0.04$ (0/8) & $+0.22$ (3/7) & $+0.07$ (0/9) \\
    MAE ViT-L & $+0.14$ (3/12) & $+0.11$ (2/13) & $+0.18$ (5/12) \\
    MAE ViT-H & $+0.07$ (0/10) & $+0.10$ (1/11) & $+0.08$ (1/10) \\
    CLIP ViT-B/16 & $+0.77$ (18/18) & -- & -- \\
    CLIP ViT-L/14 & $+0.45$ (6/8) & -- & -- \\
    \bottomrule
  \end{tabular}
  \vspace{-0.2cm}
\end{table}
By family (\cref{S-tab:families}), \method has no significant loss on \IN against any family, except for the
Prune-and-Merge pair above. The margins are smallest against the grid prune ($+0.25$\pp median) and DiffRate
($+0.57$\pp). On \INA, \method has 6 significant losses, all on DeiT: 4 to DiffRate and 2 to Prune-and-Merge
(\cref{S-sec:supp:ina}).
\vspace{-0.1cm}
\paragraph{CLIP.} We added CLIP ViT-B/16 and ViT-L/14 after we froze the recipe. On them, \method is more accurate in
all 125 pairs on \IN at batch 256 on the \gh, 123 of them significantly. All seven significant losses on the shifted
sets are on ViT-B/16, to Prune-and-Merge and ToMe.
\vspace{-0.2cm}
\paragraph{Batch size and other GPUs.}
\label{sec:exp:batch}
At batch 32 on the \gh, \method wins 310 of 321 pairs on \IN, 266 of them significantly. Both significant losses are to Prune-and-Merge configurations that run 33--35\% slower than the unreduced model. Against the 185 configurations that are faster than the unreduced model, it wins 181, 151 significantly, with no significant loss. The official ToMe code and Later Is Better run there at $0.58$--$1.67\times$ the unreduced speed and lose all their 31 and 21 pairs, 51 of them significantly. On the \ahun at batch 256, \method wins 278 of 279 pairs on five models, 241 significantly, with no significant loss. On DeiT and MAE ViT-B, schedules calibrated at batch 256 do not transfer to batch 32. They keep a median of 72--90\% of their speed-up there and lose 0.37--1.67\pp. On MAE ViT-L and ViT-H, and between the \ahun and the \gh, they transfer within 0.11\pp (\cref{S-tab:supp:transfer}).
\vspace{-0.3cm}
\paragraph{Robustness of the comparison.}
\label{sec:exp:robust}
Speed varies by a few percent between jobs, and an optimized port of a competitor might run faster. When our schedule must be 10\% faster than the competitor, \method still has 215 significant wins on \IN at batch 256 against 3 losses, 234 against 6 at batch 32, and 193 against 3 on the \ahun (\cref{S-fig:margin}). On class-token models, most of our batch-256 schedules also prune after the attention of the last block. This step changes no prediction, and no competitor uses it. Without it, our schedules run up to 7.5\% slower, and the significant wins and losses on \IN stay at 267 and 1 on the \gh and 241 and 0 on the \ahun (\cref{S-sec:supp:lastblock}).
\vspace{-0.1cm}
\subsection{Video}
\label{sec:exp:video}
On a \gh at batch 8, \method is more accurate in 23 of the 28 video pairs, 12 of them significantly after Holm
correction, with no significant loss (\cref{S-tab:video}). Its three worse pairs are all against RLT, by
$-0.22\pm0.09$\pp (not significant after Holm correction) and $-0.07\pm0.06$\pp on K400, and by $-0.06\pm0.12$\pp on
SSv2. On K400, \method therefore does not beat RLT. Without the input-level prune, the RLT pair at 58.7 GFLOPs is a
significant loss of 0.66\pp. Schedules calibrated on an RTX~4090 miss the speed of RLT at 58.7 GFLOPs on the \gh, so we
calibrate per GPU.
\vspace{-0.1cm}
\subsection{Ablations and calibration cost}
\label{sec:exp:ablation}
\paragraph{Time cost and operators.} \Cref{tab:timecost} repeats the search of \method with two changes, one at a time and together. First, on
all five models, it counts the saving in MACs instead of measured time. Second, on DeiT-B and MAE ViT-L, it uses the
operators of the competitors: ToMe's merge (on the keys, without importance weighting and norm rescaling) and top-$k$
prunes by CLS attention, received attention or the token norm, without \PPS.
At equal measured speed on the \gh at batch 256, the time cost gains $+0.06$ to $+0.47$\pp on \IN (on the \ahun,
$+0.05$ to $+0.22$\pp), most on the smallest model. Our operators gain $+0.29$ (DeiT-B) and $+0.46$\pp (MAE ViT-L).
With both changes, the variant loses $0.88$ and $0.82$\pp.
At batch 32, the MAC cost fails on the smaller models. All 14 MAC-calibrated schedules on DeiT-S, 14 of 16 on DeiT-B
and 10 of 16 on MAE ViT-B run slower than the unreduced model. At equal MACs, the schedules of \method run 1.5--2.2
times as fast.

\begin{table}[t]
  \centering
  \footnotesize
  \setlength{\tabcolsep}{3pt}
  \caption{\method (ours) against variants of its own search at equal measured speed. Each value is the top-1 of
    \method on \IN at the measured speed of a variant schedule (interpolated, unreduced model at $1\times$) minus the top-1 of the variant, averaged over the variant's schedules within the speed range of
    \method (pp). In brackets: the speed-up of \method over the variant at equal MACs. $^\dagger$All 14 schedules run
    slower than the unreduced model; on DeiT-B at batch 32, only 2 of 16 are within the speed range of \method.}
  \label{tab:timecost}
  \resizebox{\linewidth}{!}{%
  \begin{tabular}{@{}llcc@{}}
    \toprule
    variant of the search & model & \gh, 256 & \gh, 32 \\
    \midrule
    MAC cost & DeiT-S & $+0.47$ (1.22$\times$) & all slower$^\dagger$ (2.16$\times$) \\
     & DeiT-B & $+0.42$ (1.13$\times$) & $+2.10$ (1.60$\times$) \\
     & MAE ViT-B & $+0.27$ (1.09$\times$) & $+1.19$ (1.45$\times$) \\
     & MAE ViT-L & $+0.12$ (1.07$\times$) & $+0.45$ (1.22$\times$) \\
     & MAE ViT-H & $+0.06$ (1.04$\times$) & $+0.10$ (1.07$\times$) \\
    \midrule
    competitor operators, & DeiT-B & $+0.29$ & $+0.37$ \\
    time cost & MAE ViT-L & $+0.46$ & $+1.02$ \\
    \midrule
    competitor operators, & DeiT-B & $+0.88$ (1.10$\times$) & $+2.58$ (1.34$\times$) \\
    MAC cost & MAE ViT-L & $+0.82$ (1.02$\times$) & $+1.25$ (1.18$\times$) \\
    \bottomrule
  \end{tabular}}
  \vspace{-0.5cm}
\end{table}
\vspace{-0.2cm}
\paragraph{Components.} We switch off one component of the search at a time on DeiT-B and MAE ViT-L (\gh, batch 256) and interpolate the top-1 of \method at the measured speed of each ablated schedule (\cref{S-tab:ablation}). \PPS helps on \IN ($+0.04$ and $+0.24$\pp) but costs on \INA ($-0.55\pm0.17$\pp on DeiT-B; \cref{S-sec:supp:ina}). The other components change \IN by at most 0.08\pp and \INA by less than 1.5 standard errors, except the fitted $c$ and the refinement moves on DeiT-B ($-0.28\pm0.09$ and $-0.31\pm0.09$\pp). Replacing the damage curves and the allocation by random sampling with the same number of true-KL evaluations is $0.17\pm0.04$\pp less accurate on DeiT-B and ties on MAE ViT-L, and random starts refined by our moves are $0.16\pm0.06$ and $0.24\pm0.09$\pp less accurate. The greedy allocation alone is $0.24\pm0.19$ and $0.67\pm0.22$\pp less accurate (\cref{S-tab:supp:stages}), while the start with the lowest true KL is within $0.04$\pp of \method. Flat or block-averaged damage curves and two other calibration sets change \IN by less than 0.2\pp.
\vspace{-0.1cm}
\paragraph{Calibration cost.}
\label{sec:exp:cost}
On one \gh, the latency table takes 1.1--4.6 minutes, and the 4--10 base targets take 1.8--15.9 minutes per model and
batch size (2.1--20.2 without our lossless caches; \cref{S-tab:cost}). For comparison, the DiffRate search takes
2.7 GPU-hours for ViT-B in its paper (0.9 for one epoch, at 0.18\pp less top-1). The six K400 targets take 1.5 and 2.8
hours for VideoMAE-B and -L.

\vspace{-0.1cm}
\section{Conclusion}
\label{sec:conclusion}
\vspace{-0.1cm}
Token reduction should be optimized for the hardware on which it runs. \method measures the fixed and marginal costs of each operator for a given GPU and batch size, then allocates merging and pruning across blocks according to their estimated KL distortion. This training-free process takes minutes. At matched measured throughput, \method is more accurate on ImageNet than competing methods across the evaluated models and operating points, with no significant losses to the strongest competitor. On an H100 at batch size 256, its median accuracy gains are 0.83--1.51\pp on DeiT, 0.45--0.77\pp on CLIP, and up to 0.14\pp on MAE. At batch size 32, its schedules run $1.03$--$2.65\times$ faster, while most competing configurations and MAC-calibrated schedules fail to outperform the unreduced model.
\vspace{-0.2cm}
\paragraph{Limitations.}
\method is not equally useful everywhere. On MAE, simple grid pruning is nearly as effective on \IN, and some configurations trade \INA for \IN accuracy. Because \method approximates the original model, it cannot correct its prediction errors. Our evaluation covers classification at $224\times224$, static schedules, and a limited set of image models and VideoMAE. Speedups are also modest at batch size 1, and on K400 \method matches rather than surpasses RLT. Extending the approach to dense prediction and developing more efficient small-batch kernels are promising directions for future work.

\section*{LLM Usage Statement}
We used Claude (Anthropic) as an assistant at some stages of this work. For the text, it edited and shortened passages of the manuscript and corrected grammar. For the code, it helped write and debug parts of the implementation and of the evaluation and analysis scripts behind the experiments. The idea, its mathematical formulation and proofs were developed by the authors and checked independently by them. The authors accept full responsibility for the content and accuracy.

{
  \small
  \bibliographystyle{ieeenat_fullname}
  \bibliography{references}
}

\clearpage
\setcounter{page}{1}
\maketitlesupplementary

\appendix
\appendixrefs  

\renewcommand{\floatpagefraction}{.85}
\renewcommand{\dblfloatpagefraction}{.85}

\noindent\textbf{Contents.} \Cref{sec:supp:results} gives additional results per model, test set, GPU and batch size, the ablations and their variants, the checks with a speed margin and without the step after the last attention, and the CLIP, DINOv2 and video results. \Cref{sec:supp:calib} covers the schedules, the latency table, the damage curves, the transfer between settings, batch 1, the link between KL and top-1, and the cost of calibration. \Cref{sec:supp:comp} compares with our ports of the competitors and with published numbers. \Cref{sec:supp:analyses} holds the analyses and qualitative results (interference maps, robustness on \INA, \PPS and token maps). \Cref{sec:supp:related} extends the related work, \cref{sec:supp:impl} gives the implementation details, and \cref{sec:supp:theory} states and proves the propositions.

\section{Additional results}
\label{sec:supp:results}
\begin{table}[t]
  \centering
  \caption{Unreduced models. Throughput is measured in our engine with CUDA graphs and bf16.}
  \label{tab:models}
  \resizebox{\linewidth}{!}{%
  \begin{tabular}{@{}lccccc|ccc@{}}
    \toprule
    & \multicolumn{5}{c|}{top-1 (\%)} & \multicolumn{3}{c}{throughput (img/s)} \\
    model & \IN & \INV & \INR & \INA & \INS & \gh, 256 & \gh, 32 & \ahun, 256 \\
    \midrule
    DeiT-S & 79.81 & 68.41 & 41.89 & 19.53 & 29.09 & 19\,303 & 17\,001 & 9\,156 \\
    DeiT-B & 81.82 & 70.99 & 44.64 & 27.87 & 31.98 & 8\,238 & 7\,852 & 3\,805 \\
    MAE ViT-B & 83.73 & 73.17 & 48.71 & 36.73 & 35.86 & 8\,178 & 7\,950 & 3\,799 \\
    MAE ViT-L & 85.94 & 76.60 & 60.30 & 56.73 & 45.59 & 2\,893 & 2\,592 & 1\,280 \\
    MAE ViT-H & 86.88 & 78.48 & 64.06 & 68.15 & 49.04 & 1\,153 & 1\,060 & 505 \\
    CLIP ViT-B/16 & 85.21 & 75.62 & 65.26 & 47.51 & 50.47 & 8\,001 & -- & -- \\
    CLIP ViT-L/14 & 87.59 & 79.50 & 81.39 & 72.49 & 61.77 & 2\,163 & -- & -- \\
    DINOv2 ViT-B/14 & 84.29 & 75.03 & 66.52 & 57.32 & 52.53 & 5\,605 & -- & -- \\
    \bottomrule
  \end{tabular}}
\end{table}
\begin{figure*}[t]
  \centering
  \includegraphics[width=\textwidth]{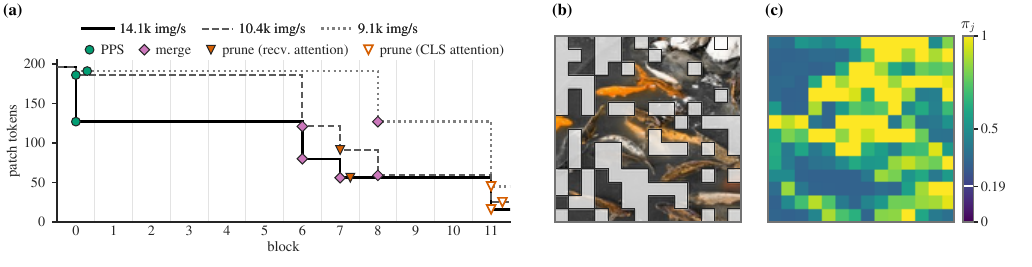}
  \caption{\textbf{\method on an example.} DeiT-B on a \gh at batch 256 (the steps of \method are shown in
    \cref{M-fig:pipeline}). (a)~Patch tokens in each block under the schedules calibrated for three throughput targets
    (14.1k, 10.4k and 9.1k img/s). A step runs after the attention of its block, so the MLP of that block already sees
    fewer tokens; glyphs mark the operator of each step. (b)~The 127 of 196 patches that the \PPS step in block~0 keeps
    on a held-out image (dropped patches faded). (c)~Its inclusion probabilities $\pi_j$: high where $z$ is large and
    never below $k\alpha/n=0.19$ (\cref{M-eq:pps_distribution,M-eq:inclusion_prob}; white mark on the color bar).}
  \label{fig:method}
\end{figure*}
\paragraph{Results per model and competitor family.}
\label{sec:supp:models}
\Cref{tab:supp:models} splits the pairs of \cref{M-tab:main} by model.
On \IN, all three significant losses are to Prune-and-Merge configurations that are slower than the unreduced model.
\INV has no significant loss.
On the shifted test sets, most significant losses are on DeiT. On \INS, DeiT-S has 4 in each setting and DeiT-B has 2--3.
\Cref{tab:supp:families} splits the batch-32, \ahun and CLIP pairs by competitor family, as \cref{tab:families} does
for batch 256.
At batch 32, PaPr accounts for 3 of the 5 significant \INA losses and DiffRate for 2. On the \ahun, all 3 are to
DiffRate.
Some configurations are measured in a second job, together with our final schedules and the unreduced model, and are
paired within that job. These are the official ToMe code and Later Is Better at batch 32, and the extra grid prunes of
every setting (keeping 110--180 of 196 patches on DeiT, 160--180 on MAE ViT-B/L and CLIP ViT-B/16, and 224 and 240 of
256 on CLIP ViT-L/14).

Some extreme competitor configurations are faster than all our schedules, so they stay unpaired.
At batch 256 on the \gh there are ten: the grid prune to 98 and 50 tokens, PaPr at 30\% (two models) and 40\%, and
Prune-and-Merge at $K=0.25$ (four models) and $K=0.4$ (DeiT-B). At batch 32 there are three: the grid prune to 98
(DeiT-B) and 50 tokens (MAE ViT-B), and PaPr at 30\% (MAE ViT-H). On the \ahun there are five: the grid prune to 98 and
110 tokens (DeiT-S), and PaPr at 30\% (MAE ViT-L and ViT-H) and 40\% (MAE ViT-H). On CLIP there are five: the grid prune
to 98 tokens and Prune-and-Merge at $K=0.25$ on ViT-B/16, and PaPr at 30\% and 40\% and Prune-and-Merge at $K=0.25$ on
ViT-L/14. All of them lose 1.5--51\pp of top-1 on \IN.

\begin{table*}[t]
  \centering
  \footnotesize
  \caption{\method (ours) against all competitors, per model. Better: pairs in which \method is more accurate on \IN.
    W/L: significant wins (bold) / losses of \method after Holm correction over all pairs of one setting and test set.
    $\tilde{\Delta}$: median paired difference, \method minus the competitor (pp).}
  \label{tab:supp:models}
  \begin{tabular}{@{}llrrcrccccrc@{}}
    \toprule
    & & & \multicolumn{3}{c}{\IN} & \INV & \INR & \multicolumn{2}{c}{\INA} & \INS \\
    \cmidrule(lr){4-6}\cmidrule(lr){9-10}
    GPU, batch & model & pairs & better & W/L & $\tilde{\Delta}$ & W/L & W/L & W/L & $\tilde{\Delta}$ & W/L \\
    \midrule
    \gh, 256 & DeiT-S & 66 & 65 & \textbf{62}/0 & $+1.13$ & \textbf{47}/0 & \textbf{58}/0 & \textbf{35}/4 & $+1.13$ & \textbf{45}/4 \\
     & DeiT-B & 66 & 65 & \textbf{63}/0 & $+1.27$ & \textbf{56}/0 & \textbf{62}/1 & \textbf{48}/2 & $+3.19$ & \textbf{58}/3 \\
     & MAE ViT-B & 62 & 58 & \textbf{51}/1 & $+1.65$ & \textbf{46}/0 & \textbf{59}/0 & \textbf{56}/0 & $+4.47$ & \textbf{57}/0 \\
     & MAE ViT-L & 63 & 62 & \textbf{48}/0 & $+0.52$ & \textbf{35}/0 & \textbf{60}/0 & \textbf{54}/0 & $+4.33$ & \textbf{58}/0 \\
     & MAE ViT-H & 57 & 54 & \textbf{43}/0 & $+0.50$ & \textbf{33}/0 & \textbf{55}/0 & \textbf{48}/0 & $+4.56$ & \textbf{55}/0 \\
    \midrule
    \gh, 32 & DeiT-S & 66 & 61 & \textbf{54}/2 & $+1.07$ & \textbf{50}/0 & \textbf{56}/0 & \textbf{41}/0 & $+1.75$ & \textbf{45}/4 \\
     & DeiT-B & 68 & 67 & \textbf{64}/0 & $+1.52$ & \textbf{61}/0 & \textbf{64}/0 & \textbf{50}/2 & $+3.67$ & \textbf{59}/3 \\
     & MAE ViT-B & 63 & 61 & \textbf{55}/0 & $+2.27$ & \textbf{46}/0 & \textbf{60}/0 & \textbf{53}/0 & $+5.28$ & \textbf{59}/0 \\
     & MAE ViT-L & 65 & 62 & \textbf{49}/0 & $+0.57$ & \textbf{38}/0 & \textbf{58}/0 & \textbf{54}/2 & $+4.35$ & \textbf{58}/1 \\
     & MAE ViT-H & 59 & 59 & \textbf{44}/0 & $+0.54$ & \textbf{40}/0 & \textbf{58}/0 & \textbf{50}/1 & $+4.44$ & \textbf{57}/0 \\
    \midrule
    \ahun, 256 & DeiT-S & 56 & 56 & \textbf{53}/0 & $+1.08$ & \textbf{40}/0 & \textbf{51}/0 & \textbf{37}/1 & $+1.45$ & \textbf{43}/4 \\
     & DeiT-B & 61 & 61 & \textbf{60}/0 & $+1.29$ & \textbf{56}/0 & \textbf{59}/1 & \textbf{48}/2 & $+3.97$ & \textbf{57}/2 \\
     & MAE ViT-B & 56 & 55 & \textbf{46}/0 & $+1.62$ & \textbf{40}/0 & \textbf{56}/0 & \textbf{51}/0 & $+5.32$ & \textbf{55}/0 \\
     & MAE ViT-L & 56 & 56 & \textbf{48}/0 & $+0.66$ & \textbf{34}/0 & \textbf{55}/0 & \textbf{51}/0 & $+5.04$ & \textbf{51}/0 \\
     & MAE ViT-H & 50 & 50 & \textbf{34}/0 & $+0.34$ & \textbf{25}/0 & \textbf{48}/0 & \textbf{45}/0 & $+3.89$ & \textbf{45}/0 \\
    \midrule
    \gh, 256 & CLIP ViT-B/16 & 62 & 62 & \textbf{62}/0 & $+1.01$ & \textbf{56}/0 & \textbf{59}/3 & \textbf{45}/3 & $+2.45$ & \textbf{55}/1 \\
     & CLIP ViT-L/14 & 63 & 63 & \textbf{61}/0 & $+0.85$ & \textbf{55}/0 & \textbf{63}/0 & \textbf{58}/0 & $+3.51$ & \textbf{60}/0 \\
    \bottomrule
  \end{tabular}
\end{table*}

\begin{table}[t]
  \centering
  \footnotesize
  \caption{\method (ours) against each competitor family on a \gh at batch 256 (all five models). W/L: significant
    wins (bold) / losses of \method after Holm correction; $\tilde{\Delta}$: median paired difference, \method minus the
    competitor (pp).}
  \label{tab:families}
  \begin{tabular}{@{}lrcrcr@{}}
    \toprule
    & & \multicolumn{4}{c}{\method (ours) against the family} \\
    \cmidrule(l){3-6}
    & & \multicolumn{2}{c}{\IN} & \multicolumn{2}{c}{\INA} \\
    family & pairs & W/L & $\tilde{\Delta}$ & W/L & $\tilde{\Delta}$ \\
    \midrule
    ToMe & 18 & \textbf{18}/0 & $+1.05$ & \textbf{17}/0 & $+4.44$ \\
    ToMe (official) & 31 & \textbf{31}/0 & $+1.05$ & \textbf{31}/0 & $+5.47$ \\
    Later Is Better & 21 & \textbf{20}/0 & $+2.70$ & \textbf{18}/0 & $+8.96$ \\
    DiffRate & 23 & \textbf{20}/0 & $+0.57$ & \textbf{11}/4 & $+0.95$ \\
    grid prune & 40 & \textbf{19}/0 & $+0.25$ & \textbf{39}/0 & $+3.91$ \\
    PaPr & 17 & \textbf{12}/0 & $+0.60$ & \textbf{8}/0 & $+1.07$ \\
    PiToMe & 24 & \textbf{24}/0 & $+1.70$ & \textbf{19}/0 & $+5.64$ \\
    PiToMe* & 24 & \textbf{24}/0 & $+1.31$ & \textbf{24}/0 & $+6.27$ \\
    ToFu & 24 & \textbf{22}/0 & $+0.94$ & \textbf{10}/0 & $+0.95$ \\
    PPT & 22 & \textbf{21}/0 & $+1.32$ & \textbf{19}/0 & $+5.29$ \\
    POMT & 35 & \textbf{29}/0 & $+0.90$ & \textbf{22}/0 & $+1.96$ \\
    Prune-and-Merge & 35 & \textbf{27}/1 & $+0.94$ & \textbf{23}/2 & $+2.56$ \\
    \bottomrule
  \end{tabular}
\end{table}

\begin{table*}[t]
  \centering
  \footnotesize
  \setlength{\tabcolsep}{3pt}
  \caption{\method (ours) against each competitor family at batch 32 on the \gh, at batch 256 on the \ahun (five models each),
    and on the two CLIP models (\gh, batch 256). Notation as in \cref{tab:families}: W/L are significant wins (bold) /
    losses of \method; $\tilde{\Delta}$ is the median of \method minus the competitor (pp).}
  \label{tab:supp:families}
  \begin{tabular}{@{}l|rcrcr|rcrcr|rcrcr@{}}
    \toprule
    & \multicolumn{5}{c|}{\gh, batch 32} & \multicolumn{5}{c|}{\ahun, batch 256} & \multicolumn{5}{c}{\gh, batch 256, CLIP} \\
    & & \multicolumn{2}{c}{\IN} & \multicolumn{2}{c|}{\INA} & & \multicolumn{2}{c}{\IN} & \multicolumn{2}{c|}{\INA} & & \multicolumn{2}{c}{\IN} & \multicolumn{2}{c}{\INA} \\
    family & pairs & W/L & $\tilde{\Delta}$ & W/L & $\tilde{\Delta}$ & pairs & W/L & $\tilde{\Delta}$ & W/L & $\tilde{\Delta}$ & pairs & W/L & $\tilde{\Delta}$ & W/L & $\tilde{\Delta}$ \\
    \midrule
    ToMe & 18 & \textbf{18}/0 & $+1.11$ & \textbf{18}/0 & $+4.08$ & 18 & \textbf{17}/0 & $+0.96$ & \textbf{18}/0 & $+4.21$ & 10 & \textbf{10}/0 & $+0.71$ & \textbf{10}/0 & $+3.76$ \\
    ToMe (official) & 31 & \textbf{31}/0 & $+1.18$ & \textbf{31}/0 & $+5.21$ & 31 & \textbf{30}/0 & $+1.01$ & \textbf{31}/0 & $+5.36$ & 15 & \textbf{15}/0 & $+0.81$ & \textbf{15}/0 & $+3.89$ \\
    Later Is Better & 21 & \textbf{20}/0 & $+2.71$ & \textbf{19}/0 & $+10.48$ & 21 & \textbf{21}/0 & $+2.61$ & \textbf{20}/0 & $+9.31$ & 10 & \textbf{10}/0 & $+2.06$ & \textbf{9}/0 & $+12.37$ \\
    DiffRate & 23 & \textbf{17}/0 & $+0.57$ & \textbf{11}/2 & $+0.96$ & 23 & \textbf{19}/0 & $+0.54$ & \textbf{12}/3 & $+0.99$ & -- & -- & -- & -- & -- \\
    grid prune & 40 & \textbf{19}/0 & $+0.24$ & \textbf{36}/0 & $+3.50$ & 40 & \textbf{20}/0 & $+0.23$ & \textbf{40}/0 & $+4.82$ & 16 & \textbf{15}/0 & $+0.71$ & \textbf{16}/0 & $+7.38$ \\
    PaPr & 19 & \textbf{13}/0 & $+0.51$ & \textbf{7}/3 & $+0.91$ & 17 & \textbf{14}/0 & $+0.70$ & \textbf{13}/0 & $+1.68$ & 8 & \textbf{7}/0 & $+0.93$ & \textbf{6}/0 & $+2.03$ \\
    PiToMe & 24 & \textbf{24}/0 & $+1.69$ & \textbf{19}/0 & $+5.97$ & 24 & \textbf{24}/0 & $+1.60$ & \textbf{20}/0 & $+4.94$ & 10 & \textbf{10}/0 & $+1.33$ & \textbf{7}/0 & $+1.89$ \\
    PiToMe* & 24 & \textbf{24}/0 & $+1.30$ & \textbf{24}/0 & $+6.15$ & 24 & \textbf{23}/0 & $+1.35$ & \textbf{24}/0 & $+6.47$ & 10 & \textbf{10}/0 & $+1.24$ & \textbf{10}/0 & $+4.69$ \\
    ToFu & 24 & \textbf{23}/0 & $+1.17$ & \textbf{17}/0 & $+2.06$ & 24 & \textbf{23}/0 & $+0.87$ & \textbf{12}/0 & $+1.37$ & 10 & \textbf{10}/0 & $+1.14$ & \textbf{5}/0 & $+1.48$ \\
    PPT & 22 & \textbf{19}/0 & $+1.52$ & \textbf{19}/0 & $+5.35$ & 22 & \textbf{21}/0 & $+1.31$ & \textbf{19}/0 & $+5.63$ & 9 & \textbf{9}/0 & $+0.79$ & \textbf{8}/0 & $+3.23$ \\
    POMT & 35 & \textbf{27}/0 & $+0.83$ & \textbf{19}/0 & $+1.37$ & 35 & \textbf{29}/0 & $+1.01$ & \textbf{23}/0 & $+1.85$ & 13 & \textbf{13}/0 & $+1.21$ & \textbf{8}/0 & $+1.87$ \\
    Prune-and-Merge & 40 & \textbf{31}/2 & $+1.33$ & \textbf{28}/0 & $+5.69$ & -- & -- & -- & -- & -- & 14 & \textbf{14}/0 & $+0.78$ & \textbf{9}/3 & $+1.24$ \\
    \bottomrule
  \end{tabular}
\end{table*}

\begin{table}[t]
  \centering
  \footnotesize
  \setlength{\tabcolsep}{3pt}
  \caption{Top-1 on \IN of every method at two measured speed-ups over the unreduced model (\gh, batch 256). Each value
    is read off the method's front of measured configurations by linear interpolation; --: the method's configurations
    do not bracket this speed-up. Best in bold, second underlined; grey: the unreduced model.}
  \label{tab:equalspeed}
  \resizebox{\linewidth}{!}{%
  \begin{tabular}{@{}lcccccccccc@{}}
  \toprule
  & \multicolumn{5}{c}{1.25$\times$} & \multicolumn{5}{c}{1.5$\times$} \\
  \cmidrule(lr){2-6} \cmidrule(lr){7-11}
  & \multicolumn{2}{c}{DeiT} & \multicolumn{3}{c}{MAE} & \multicolumn{2}{c}{DeiT} & \multicolumn{3}{c}{MAE} \\
  method & S & B & B & L & H & S & B & B & L & H \\
  \midrule
  unreduced & \textcolor{black!55}{79.81} & \textcolor{black!55}{81.82} & \textcolor{black!55}{83.73} & \textcolor{black!55}{85.94} & \textcolor{black!55}{86.88} & \textcolor{black!55}{79.81} & \textcolor{black!55}{81.82} & \textcolor{black!55}{83.73} & \textcolor{black!55}{85.94} & \textcolor{black!55}{86.88} \\
  \midrule
  ToMe & -- & 79.52 & 79.81 & -- & 86.54 & -- & -- & -- & -- & 86.03 \\
  ToMe (official) & -- & 78.68 & 80.94 & 85.05 & 86.43 & -- & -- & -- & 83.96 & 85.79 \\
  Later Is Better & -- & 78.12 & 78.87 & 85.01 & 85.88 & -- & -- & -- & 79.36 & 80.43 \\
  DiffRate & -- & 79.20 & 80.08 & 84.72 & 86.47 & -- & -- & -- & -- & -- \\
  grid prune & 78.00 & \underline{80.64} & \underline{82.88} & \underline{85.54} & \underline{86.72} & \underline{76.36} & \underline{79.55} & \underline{82.38} & \underline{85.04} & 86.44 \\
  PaPr & -- & 79.80 & 82.15 & 85.46 & -- & -- & 76.05 & 79.61 & 84.95 & \underline{86.49} \\
  PiToMe & -- & 78.42 & 77.84 & 83.90 & 86.03 & -- & -- & -- & -- & 85.31 \\
  PiToMe* & -- & 76.11 & 79.64 & 84.58 & 86.22 & -- & -- & -- & -- & -- \\
  ToFu & 77.22 & 80.44 & 82.38 & 85.27 & 86.59 & -- & 78.91 & 80.86 & 84.60 & 86.14 \\
  PPT & \underline{78.64} & 79.07 & 78.82 & 84.83 & 86.52 & -- & -- & 68.84 & 81.64 & 85.72 \\
  POMT & 77.19 & 79.67 & 81.63 & 85.35 & 86.67 & 71.32 & 73.70 & 76.93 & 84.03 & 86.28 \\
  Prune-and-Merge & 78.11 & 80.40 & 81.84 & 85.30 & 85.73 & 74.65 & 77.63 & 78.42 & 84.54 & 85.17 \\
  \midrule
  \method (ours) & \textbf{79.47} & \textbf{81.56} & \textbf{83.09} & \textbf{85.70} & \textbf{86.81} & \textbf{78.37} & \textbf{80.74} & \textbf{82.57} & \textbf{85.30} & \textbf{86.64} \\
  \bottomrule
  \end{tabular}}
\end{table}

\paragraph{Against the strongest competitor.}
\label{sec:supp:envelope}
\Cref{tab:supp:envelope} details \cref{M-tab:envelope}. A competitor configuration belongs to the envelope of a model if
two conditions hold: it is faster than the unreduced model, and no other competitor configuration of that model (paired
or not) is at least as fast and at least as accurate on \IN. The envelope is thus the front of the competitor
configurations on \IN. Each envelope configuration is paired with our slowest schedule that is at least as
fast, as in \cref{M-tab:main}. Those faster than all our schedules stay unpaired and are not counted.

On the \gh, the envelope consists mostly of grid prunes: 8 of 10 configurations on DeiT-B and all on MAE ViT-B at
batch 256, and all on DeiT at batch 32. On DeiT-S at batch 256, it holds 5 PPT and 2 Prune-and-Merge configurations
besides 4 grid prunes. POMT and PaPr join it on MAE ViT-L and ViT-H. On \IN, no envelope pair is a significant loss,
and single pairs are up to 0.32\pp worse on MAE. On \INA, the envelope has three significant losses, all at batch 32
on MAE ViT-L and ViT-H.

\begin{table}[t]
  \centering
  \footnotesize
  \setlength{\tabcolsep}{3pt}
  \caption{\method (ours) against the competitor envelope, per setting and model. $n$: number of envelope
    configurations. On \IN: pairs in which \method is more accurate, significant wins / losses after Holm correction
    (over all pairs of the setting, as in \cref{M-tab:main}), and the median paired difference with its range. On \INA:
    wins / losses and the median (pp).}
  \label{tab:supp:envelope}
  \resizebox{\linewidth}{!}{%
\begin{tabular}{@{}llrrrcrr@{}}
\toprule
& & & \multicolumn{3}{c}{\IN} & \multicolumn{2}{c}{\INA} \\
\cmidrule(lr){4-6}\cmidrule(l){7-8}
setting & model & $n$ & better & W/L & median [range] & W/L & median \\
\midrule
\gh, 256 & DeiT-S & 11 & 11 & 10/0 & $+1.51$ [$+0.11$, $+13.75$] & 7/0 & $+2.64$ \\
 & DeiT-B & 10 & 10 & 10/0 & $+0.83$ [$+0.39$, $+1.37$] & 8/0 & $+3.57$ \\
 & MAE ViT-B & 8 & 6 & 0/0 & $+0.04$ [$-0.18$, $+0.28$] & 8/0 & $+2.45$ \\
 & MAE ViT-L & 12 & 11 & 3/0 & $+0.14$ [$-0.02$, $+0.37$] & 10/0 & $+2.97$ \\
 & MAE ViT-H & 10 & 7 & 0/0 & $+0.07$ [$-0.15$, $+0.22$] & 6/0 & $+3.36$ \\
\midrule
\gh, 32 & DeiT-S & 8 & 8 & 7/0 & $+1.13$ [$+0.26$, $+2.77$] & 8/0 & $+4.94$ \\
 & DeiT-B & 8 & 8 & 8/0 & $+0.62$ [$+0.32$, $+1.09$] & 7/0 & $+4.22$ \\
 & MAE ViT-B & 7 & 6 & 3/0 & $+0.22$ [$-0.04$, $+0.51$] & 5/0 & $+2.92$ \\
 & MAE ViT-L & 13 & 10 & 2/0 & $+0.11$ [$-0.32$, $+1.02$] & 8/2 & $+2.64$ \\
 & MAE ViT-H & 11 & 11 & 1/0 & $+0.10$ [$+0.02$, $+0.33$] & 6/1 & $+2.27$ \\
\midrule
\ahun, 256 & DeiT-S & 6 & 6 & 5/0 & $+0.64$ [$+0.04$, $+1.71$] & 3/0 & $+0.71$ \\
 & DeiT-B & 12 & 12 & 12/0 & $+0.81$ [$+0.38$, $+1.69$] & 10/0 & $+4.49$ \\
 & MAE ViT-B & 9 & 8 & 0/0 & $+0.07$ [$-0.06$, $+0.22$] & 9/0 & $+3.12$ \\
 & MAE ViT-L & 12 & 12 & 5/0 & $+0.18$ [$+0.08$, $+1.13$] & 10/0 & $+2.94$ \\
 & MAE ViT-H & 10 & 10 & 1/0 & $+0.08$ [$+0.02$, $+0.26$] & 8/0 & $+3.59$ \\
\midrule
\gh, 256, CLIP & CLIP ViT-B/16 & 18 & 18 & 18/0 & $+0.77$ [$+0.26$, $+1.37$] & 14/0 & $+2.75$ \\
 & CLIP ViT-L/14 & 8 & 8 & 6/0 & $+0.45$ [$+0.04$, $+1.46$] & 7/0 & $+2.43$ \\
\bottomrule
\end{tabular}}
\end{table}

\paragraph{Rankings at equal MACs and at equal measured speed.}
\label{sec:supp:rankflip}
\Cref{tab:supp:rankflip} reads the front of every method at two kinds of operating points: at $1.5\times$ fewer MACs,
and at measured speed-ups of $1.25\times$ and $1.5\times$ (by linear interpolation, as in \cref{tab:equalspeed}).
At equal MACs, the most accurate competitor is DiffRate on four models and PaPr on MAE ViT-B. At equal measured speed,
this competitor either does not reach the speed-up or ranks fourth to tenth. There, the most accurate competitor is the
grid prune in all but three cells.

\begin{table*}[t]
  \centering
  \footnotesize
  \caption{The competitor that is most accurate on \IN at $1.5\times$ fewer MACs (top-1 in \%), its rank among all
    methods (\method included) at two measured speed-ups (--: its front does not reach the speed-up), and the most
    accurate competitor at those speed-ups (top-1 in \%).}
  \label{tab:supp:rankflip}
\begin{tabular}{@{}llllllll@{}}
\toprule
setting & model & \multicolumn{2}{l}{best at 1.5$\times$ fewer MACs} & \multicolumn{2}{l}{at 1.25$\times$ measured: rank, best competitor} & \multicolumn{2}{l}{at 1.5$\times$ measured: rank, best competitor} \\
\midrule
\gh, 256 & DeiT-S & DiffRate & 79.70 & -- & PPT 78.64 & -- & grid prune 76.36 \\
\gh, 256 & DeiT-B & DiffRate & 81.51 & 8 of 13 & grid prune 80.64 & -- & grid prune 79.55 \\
\gh, 256 & MAE ViT-B & PaPr & 83.11 & 4 of 13 & grid prune 82.88 & 4 of 7 & grid prune 82.38 \\
\gh, 256 & MAE ViT-L & DiffRate & 85.53 & 10 of 12 & grid prune 85.54 & -- & grid prune 85.04 \\
\gh, 256 & MAE ViT-H & DiffRate & 86.67 & 7 of 12 & grid prune 86.72 & -- & PaPr 86.49 \\
\gh, 32 & DeiT-S & DiffRate & 79.70 & -- & grid prune 76.60 & -- & -- \\
\gh, 32 & DeiT-B & DiffRate & 81.51 & -- & grid prune 80.32 & -- & grid prune 78.81 \\
\gh, 32 & MAE ViT-B & PaPr & 83.11 & 4 of 5 & grid prune 82.69 & -- & grid prune 82.06 \\
\gh, 32 & MAE ViT-L & DiffRate & 85.53 & -- & grid prune 85.56 & -- & grid prune 85.14 \\
\gh, 32 & MAE ViT-H & DiffRate & 86.67 & 5 of 12 & grid prune 86.70 & -- & PaPr 86.50 \\
\bottomrule
\end{tabular}
\end{table*}

\begin{figure*}[t]
  \centering
  \includegraphics[width=\textwidth]{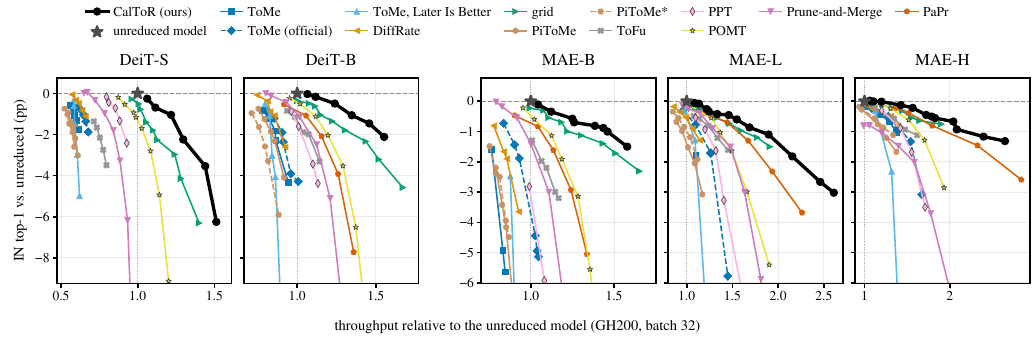}
  \caption{Fronts on \IN at batch 32 on a \gh, as in \cref{M-fig:fronts}; black, thick: \method (ours).}
  \label{fig:supp:fronts32}
\end{figure*}

\begin{figure*}[t]
  \centering
  \includegraphics[width=\textwidth]{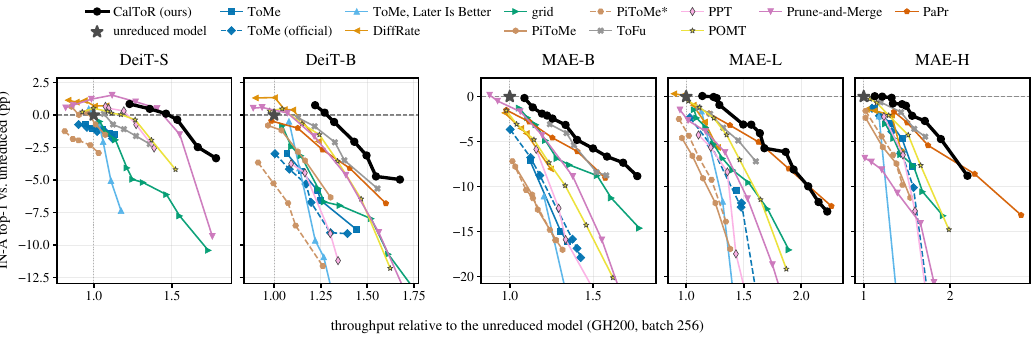}
  \caption{Fronts on \INA at batch 256 on a \gh; black, thick: \method (ours). The dashed line marks the unreduced
    model.}
  \label{fig:supp:frontsA}
\end{figure*}

\begin{figure*}[t]
  \centering
  \includegraphics[width=\textwidth]{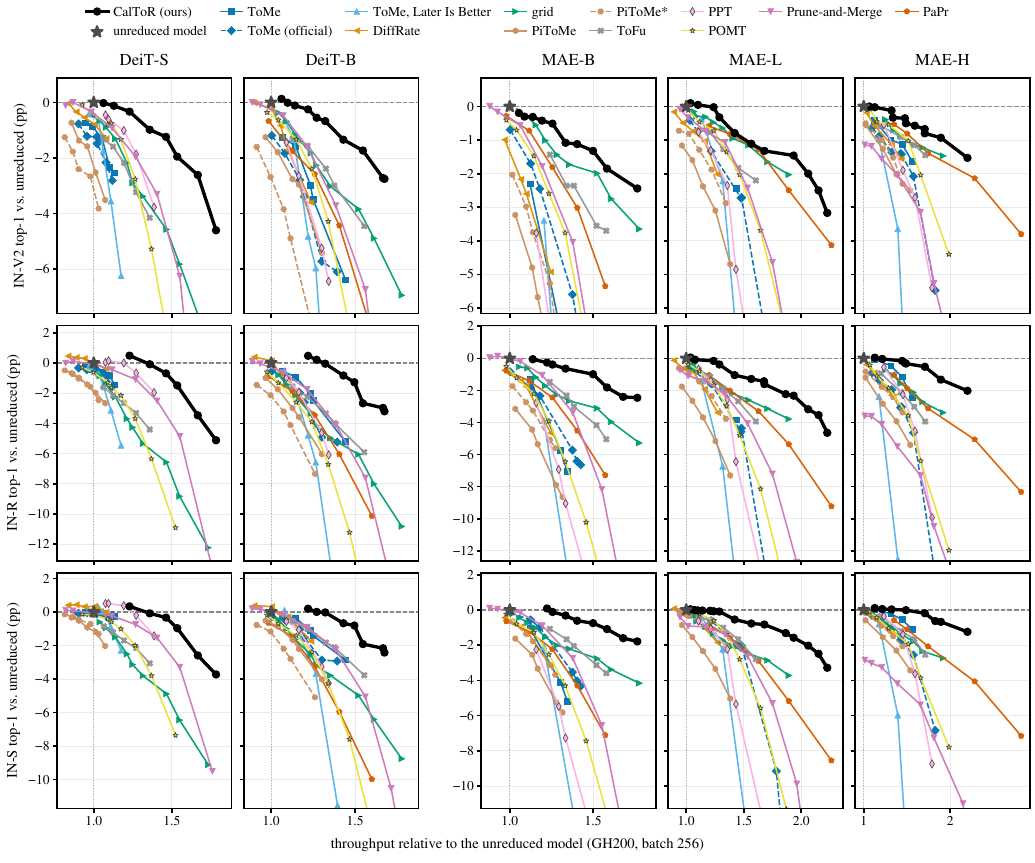}
  \caption{Fronts on \INV (top), \INR (middle) and \INS (bottom) at batch 256 on a \gh; black, thick: \method (ours).
    The dashed line marks the unreduced model.}
  \label{fig:supp:frontsood}
\end{figure*}

\begin{figure*}[t]
  \centering
  \includegraphics[width=\textwidth]{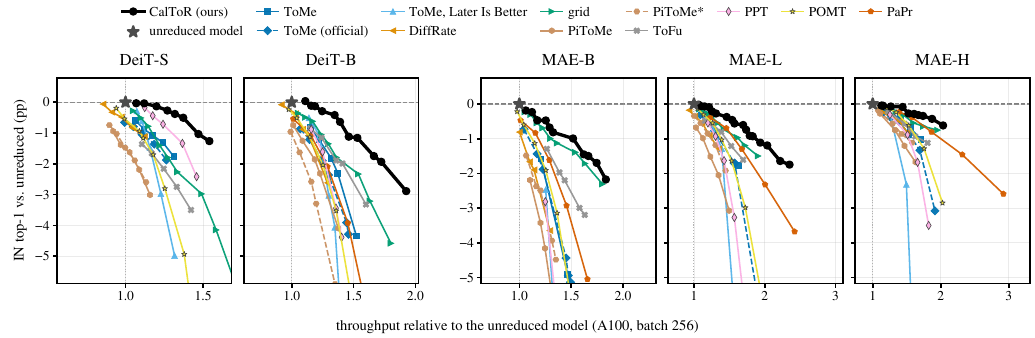}
  \caption{Fronts on \IN at batch 256 on an \ahun; black, thick: \method (ours).}
  \label{fig:supp:fronts100}
\end{figure*}

\paragraph{Fronts at batch 32, on the shifted test sets and on the \ahun.}
\label{sec:supp:fronts}
\Cref{fig:supp:fronts32,fig:supp:frontsA,fig:supp:frontsood,fig:supp:fronts100} extend \cref{M-fig:fronts}.
At batch 32, many competitor configurations run slower than the unreduced model, while all our schedules run faster
($1.03$--$2.65\times$).
On \INA, DiffRate on DeiT and Prune-and-Merge on DeiT-S lie above the front of \method at some speeds
(\cref{sec:supp:ina}). On the three MAE models, \method has no significant loss on \INA at batch 256.
On \INV, \INR and \INS (\cref{fig:supp:frontsood}), \method has no significant loss on the three MAE models, and none
at all on \INV. On DeiT, a few competitor configurations are significantly more accurate than their pair. On \INR,
it is one DiffRate configuration that is slower than the unreduced model. On \INS, they are two PPT configurations
on DeiT-S at $1.08$--$1.09\times$, and four DiffRate and one Prune-and-Merge configurations at $0.84$--$1.00\times$
(\cref{tab:supp:models}).

\begin{figure*}[t]
  \centering
  \includegraphics[width=\textwidth]{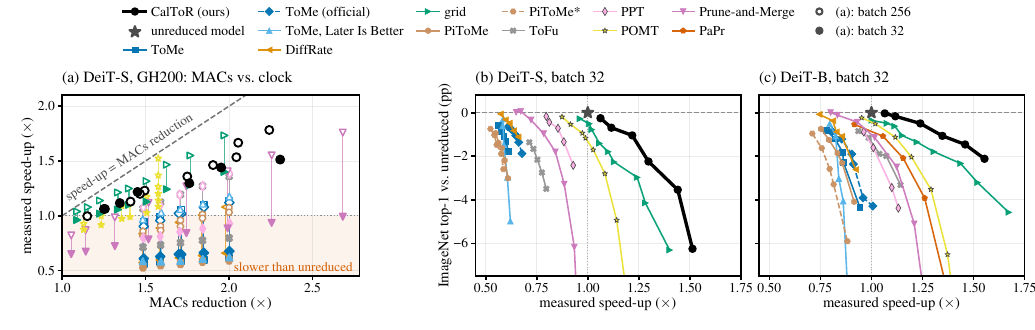}
  \caption{\textbf{MACs do not set the speed on a GPU; \method calibrates to the clock.} (a) Measured speed-up
    against the MACs reduction of each configuration on DeiT-S (\gh). Hollow markers: batch 256; filled markers: batch 32.
    A line joins the two measurements of one competitor configuration. At batch 32, 54 of 66 competitor configurations
    run slower than the unreduced model (shaded area), although they cut the MACs by up to $2.7\times$. (b, c) ImageNet
    top-1 relative to the unreduced model against the measured speed-up at batch 32, on shared axes. \method (ours,
    black) runs $1.06$--$1.51\times$ (DeiT-S) and $1.07$--$1.55\times$ (DeiT-B) faster than the unreduced model. In
    (b, c), every competitor configuration that is faster than the unreduced model is less accurate than a \method
    schedule that is at least as fast. The one exception is a grid prune of DeiT-B to 98 tokens: it is faster than all
    our schedules ($1.67\times$) and 4.6\pp below the unreduced model.}
  \label{fig:supp:macs}
\end{figure*}

\paragraph{Ablations.}
\label{sec:supp:ablation}
\Cref{tab:ablation} lists the ablations of \cref{M-sec:exp:ablation} with standard errors over the ablated schedules.
Without \PPS, the prunes keep the other scores of the menu. \emph{Several starts} compares with three starts: the
greedy allocations with all operators and with merging only, and the single prune in block~0. Each of them is refined
with one unranked proposal per step, without screening, fitted $c$ or final pick. \emph{Residual size} leaves \PPS
only the token norm. \emph{Fitted $c$} drops the start with the fitted coherence and the ranking of the proposals by
\cref{M-eq:coherent}; the starts with $c=0.2$ and $0.5$ stay. \emph{2048-image pick} compares with a pick on 512 other
images.

The two random searches get the same number of true-KL evaluations per target as \method. They spend as many true-KL evaluations on the
calibration images as \method spends on its starts, its refinement and its share of the damage curves (332 on DeiT-B,
512 on MAE ViT-L). They use the operators, per-block prune scores, latency table, targets, images and final pick of
\method. A random schedule picks 1--8 cells $(o,l)$ at random (operator $o$ in block $l$). It gives them shares of a
common reduction level, with the shares drawn from a flat Dirichlet distribution, and takes the smallest level whose
predicted time meets the target. Random sampling passes the four schedules with the lowest KL to the final pick. With
random starts, 10 random schedules are screened by their KL and the best four are refined with our moves. Since no
damage model ranks the proposals, each true-KL evaluation tests one proposal. Without the refinement moves, the ten
starts of \method are screened on the true KL and the best four go to the final pick.
The searches compute their KLs on an RTX~3080, using the prune scores and starts from the \gh calibration of \method.
When \method is also calibrated again on the RTX~3080, random sampling and random starts are $0.24\pm0.07$ and
$0.25\pm0.06$\pp less accurate than \method on DeiT-B at the same targets.

\begin{table}[t]
  \centering
  \footnotesize
  \setlength{\tabcolsep}{3pt}
  \caption{Ablations of \method on a \gh at batch 256. Each entry is the top-1 of \method minus that of the variant
    without the component, at equal measured speed (pp), given as mean $\pm$ standard error over 5--7 (DeiT-B) and 2--4
    (MAE ViT-L) ablated schedules; positive means the component helps. Middle: \method minus a random search that
    replaces the damage curves and the allocation, with the same number of true-KL evaluations (and our prune scores):
    random sampling, or random starts refined by our moves.
    Bottom: \method minus the same recipe on another set of calibration
    images. The first row is from an earlier run with 5 and 3 MAC-cost schedules; \cref{M-tab:timecost,tab:supp:timecost}
    report the full run. $^\ast$From a later run in which our schedules are measured again together with the variants.
    On DeiT-B, \method and most variants meet the slowest target with only a prune after the last attention, which
    changes no prediction. Without that target, the DeiT-B means change by at most 0.03\pp on \IN (\PPS on \INA:
    $-0.65$).}
  \label{tab:ablation}
  \resizebox{\linewidth}{!}{%
  \begin{tabular}{@{}lrrrr@{}}
    \toprule
    & \multicolumn{2}{c}{DeiT-B} & \multicolumn{2}{c}{MAE ViT-L} \\
    \cmidrule(lr){2-3}\cmidrule(l){4-5}
    component & \IN & \INA & \IN & \INA \\
    \midrule
    time cost instead of MACs & $+0.42\pm0.21$ & $+0.78\pm0.77$ & $+0.13\pm0.04$ & $-0.09\pm1.18$ \\
    \PPS & $+0.04\pm0.04$ & $-0.55\pm0.17$ & $+0.24\pm0.19$ & $-0.62\pm1.18$ \\
    several starts & $+0.07\pm0.05$ & $+0.24\pm0.38$ & $+0.04\pm0.05$ & $+1.15\pm2.04$ \\
    residual size & $0.00\pm0.00$ & $0.00\pm0.01$ & $+0.08\pm0.07$ & $+0.47\pm1.76$ \\
    fitted $c$ & $+0.01\pm0.02$ & $-0.28\pm0.09$ & $-0.03\pm0.05$ & $+0.02\pm0.26$ \\
    2048-image pick & $+0.01\pm0.01$ & $-0.12\pm0.09$ & $-0.04\pm0.04$ & $-0.51\pm0.49$ \\
    refinement moves$^\ast$ & $+0.03\pm0.06$ & $-0.31\pm0.09$ & $-0.01\pm0.04$ & $+0.85\pm1.07$ \\
    \midrule
    random sampling$^\ast$ & $+0.17\pm0.04$ & $+0.92\pm0.45$ & $-0.02\pm0.04$ & $+0.58\pm1.06$ \\
    random starts$^\ast$ & $+0.16\pm0.06$ & $+0.24\pm0.30$ & $+0.24\pm0.09$ & $+2.33\pm1.45$ \\
    \midrule
    calibration set 2 & $+0.01\pm0.02$ & $0.00\pm0.09$ & $-0.04\pm0.06$ & $-0.24\pm0.69$ \\
    calibration set 3 & $+0.06\pm0.04$ & $-0.10\pm0.11$ & $0.00\pm0.07$ & $-0.27\pm1.03$ \\
    \bottomrule
  \end{tabular}}
\end{table}

\paragraph{Steps of the search.}
\label{sec:supp:stages}
\Cref{tab:supp:stages} takes the search apart along its four steps (\cref{M-sec:method}) on DeiT-B and MAE ViT-L
(\gh, batch 256), at the targets of \method. Three variants are intermediate results of the calibration of \method: its
greedy start with all operators (steps 1--3); the start with the lowest true KL on the 512 calibration images (no
refinement, no pick); and the refined schedule with the lowest calibration KL (no pick). The fourth variant of the top block, the best
four starts with the 2048-image pick and no refinement, is the variant without refinement moves of \cref{tab:ablation}.
For step 2, we replace the measured damage curves in one of two ways: by $D_{o,l}(\rho)=\rho^2$ in every cell (flat), or
by each operator's curve averaged over the blocks (block-averaged). We keep the per-block prune scores of \method, which
were chosen on the measured curves, and rerun either the greedy allocation alone or the whole recipe. All schedules of a
model are measured in one job, together with those of \method. The row without refinement therefore differs from
\cref{tab:ablation} only through the speed measurement.

Screening the starts on the true KL brings almost all of the gain of step 4. The refinement and the pick change \IN by at
most $0.04$\pp, and on DeiT-B the refinement costs \INA accuracy. On \INR, the curves of \method gain $+0.30\pm0.16$ and
$+0.54\pm0.25$\pp over the flat and block-averaged curves on DeiT-B, and lose $0.25\pm0.24$ and $0.62\pm0.23$\pp on MAE
ViT-L. On DeiT-B, the greedy start alone is within 1.3 standard errors of \method on \IN. When run alone with flat or
block-averaged curves, the greedy allocation is more accurate on \IN of MAE ViT-L (by 0.16 and 0.39\pp), but much less
accurate on \INA on both models (by 1.0--2.4\pp). With these curves it prunes from block 0 onward and merges in 1 of its 24
schedules. With the measured curves it merges in 11 of 12, in blocks 5--9 of DeiT-B and 10--19 of MAE ViT-L.

\begin{table}[t]
  \centering
  \footnotesize
  \setlength{\tabcolsep}{3pt}
  \caption{The steps of the search on a \gh at batch 256. Each entry is the top-1 of \method minus that of the variant at
    equal measured speed (pp), as mean $\pm$ standard error over the 6--8 (DeiT-B) and 2--4 (MAE ViT-L) variant schedules
    within the speed range of the reference. Positive values mean that the omitted part helps. Top: the search stopped
    after the greedy allocation, after screening the starts, after the pick among the unrefined starts, and without the
    final pick. Middle and bottom: the measured damage curves against flat ($\rho^2$) and block-averaged curves, in the
    greedy allocation alone (middle) and in the whole recipe (bottom). $^\ast$Compared with the greedy start of \method on
    the measured curves instead of \method.}
  \label{tab:supp:stages}
  \resizebox{\linewidth}{!}{%
  \begin{tabular}{@{}lrrrr@{}}
    \toprule
    & \multicolumn{2}{c}{DeiT-B} & \multicolumn{2}{c}{MAE ViT-L} \\
    \cmidrule(lr){2-3}\cmidrule(l){4-5}
    variant & \IN & \INA & \IN & \INA \\
    \midrule
    greedy start alone (steps 1--3) & $+0.24\pm0.19$ & $0.00\pm0.15$ & $+0.67\pm0.22$ & $+0.05\pm1.19$ \\
    best start by true KL & $-0.02\pm0.02$ & $-0.35\pm0.07$ & $+0.04\pm0.00$ & $+0.01\pm0.66$ \\
    best four starts, pick & $-0.01\pm0.02$ & $-0.23\pm0.09$ & $+0.01\pm0.05$ & $+1.19\pm1.46$ \\
    refinement, no pick & $+0.01\pm0.01$ & $-0.06\pm0.07$ & $-0.03\pm0.04$ & $-0.51\pm0.44$ \\
    \midrule
    greedy start, flat curves$^\ast$ & $+0.24\pm0.15$ & $+1.47\pm0.22$ & $-0.16\pm0.03$ & $+1.04\pm0.39$ \\
    greedy start, block-averaged$^\ast$ & $+0.10\pm0.10$ & $+2.28\pm0.22$ & $-0.39\pm0.10$ & $+2.36\pm0.28$ \\
    \midrule
    flat curves & $+0.09\pm0.06$ & $+0.28\pm0.28$ & $-0.07\pm0.03$ & $-0.63\pm0.92$ \\
    block-averaged curves & $+0.16\pm0.05$ & $+0.71\pm0.37$ & $-0.06\pm0.05$ & $+0.44\pm0.61$ \\
    \bottomrule
  \end{tabular}}
\end{table}

\paragraph{Time cost and operators.}
\label{sec:supp:timecost}
Each variant in \cref{M-tab:timecost} changes the cost, the operators, or both, in the search of \method. With the MAC
cost (all five models), the allocation and the refinement count MACs (multiply-accumulate operations) instead of the
predicted time, and the prune score of a block is chosen at equal tokens. The targets are the GMACs of our final
schedules in the three settings, merged within 3\%. MACs have no fixed costs, so the allocation takes single steps of
2.5\% and no move empties a cell. The starts, the coherent model and the final pick are those of \method.

With the competitors' operators (DeiT-B and MAE ViT-L), the merge is ToMe's bipartite matching on the keys with
size-weighted averaging and proportional attention. It omits the importance weighting, the norm rescaling and the matching on
the token states of our merge. The prunes keep the top-$k$ tokens by the CLS attention, the received attention or (DeiT) the token
norm, without \PPS. All damage curves are measured afresh. We calibrate these operators to the time of the \gh at batch
256 and 32, at the targets of our final schedules, and to MACs. All variant schedules are measured in one job per model
and setting, together with our final schedules, and evaluated on all five test sets.

\Cref{tab:supp:timecost} adds \INA and the number of variant schedules slower than the unreduced model. On \INA, the
competitors' operators with the time cost are slightly better than ours on DeiT-B, within one standard error ($-0.14$
and $-0.33$\pp), and worse on MAE ViT-L ($+0.94$ and $+1.22$\pp). On the \ahun at batch 256, the MAC cost loses less
($0.05$--$0.22$\pp on \IN), and 4 of the 14 MAC-calibrated schedules of DeiT-S run slower than the unreduced model.

\begin{table*}[t]
  \centering
  \footnotesize
  \caption{\method (ours) against variants of its search, as in \cref{M-tab:timecost}. Each entry is the top-1 of \method at
    the measured speed of each variant schedule minus that of the variant, as mean $\pm$ standard error over the
    variant's schedules within the speed range of \method (pp). Slower: variant schedules slower than the unreduced
    model / all. On the \ahun, only the MAC cost was run.}
  \label{tab:supp:timecost}
  \resizebox{\textwidth}{!}{%
  \begin{tabular}{@{}llccccccccc@{}}
    \toprule
    & & \multicolumn{3}{c}{\gh, batch 256} & \multicolumn{3}{c}{\gh, batch 32} & \multicolumn{3}{c}{\ahun, batch 256} \\
    \cmidrule(lr){3-5}\cmidrule(lr){6-8}\cmidrule(l){9-11}
    variant of the search & model & \IN & \INA & slower & \IN & \INA & slower & \IN & \INA & slower \\
    \midrule
    MAC cost & DeiT-S & $+0.47\pm0.16$ & $+0.25\pm0.35$ & 6/14 & all slower & & 14/14 & $+0.22\pm0.09$ & $+0.04\pm0.16$ & 4/14 \\
     & DeiT-B & $+0.42\pm0.10$ & $+0.67\pm0.31$ & 4/16 & $+2.10\pm0.38$ & $+4.95\pm1.49$ & 14/16 & $+0.20\pm0.06$ & $+0.34\pm0.18$ & 1/16 \\
     & MAE ViT-B & $+0.27\pm0.04$ & $+0.83\pm0.17$ & 1/16 & $+1.19\pm0.17$ & $+4.38\pm0.70$ & 10/16 & $+0.15\pm0.04$ & $+0.51\pm0.12$ & 0/16 \\
     & MAE ViT-L & $+0.12\pm0.03$ & $-0.07\pm0.24$ & 1/21 & $+0.45\pm0.07$ & $+1.80\pm0.45$ & 4/21 & $+0.06\pm0.03$ & $-0.09\pm0.13$ & 0/21 \\
     & MAE ViT-H & $+0.06\pm0.02$ & $+0.44\pm0.20$ & 0/20 & $+0.10\pm0.02$ & $+0.29\pm0.17$ & 1/20 & $+0.05\pm0.02$ & $+0.33\pm0.15$ & 0/20 \\
    \midrule
    competitor operators, & DeiT-B & $+0.29\pm0.07$ & $-0.14\pm0.20$ & 0/12 & $+0.37\pm0.15$ & $-0.33\pm0.33$ & 0/7 & -- & -- & -- \\
    time cost & MAE ViT-L & $+0.46\pm0.13$ & $+0.94\pm0.63$ & 0/19 & $+1.02\pm0.36$ & $+1.22\pm0.54$ & 0/14 & -- & -- & -- \\
    \midrule
    competitor operators, & DeiT-B & $+0.88\pm0.21$ & $+1.41\pm0.38$ & 3/16 & $+2.58\pm0.48$ & $+4.40\pm0.97$ & 11/16 & -- & -- & -- \\
    MAC cost & MAE ViT-L & $+0.82\pm0.21$ & $+2.60\pm0.76$ & 0/21 & $+1.25\pm0.32$ & $+4.64\pm1.21$ & 1/21 & -- & -- & -- \\
    \bottomrule
  \end{tabular}}
\end{table*}

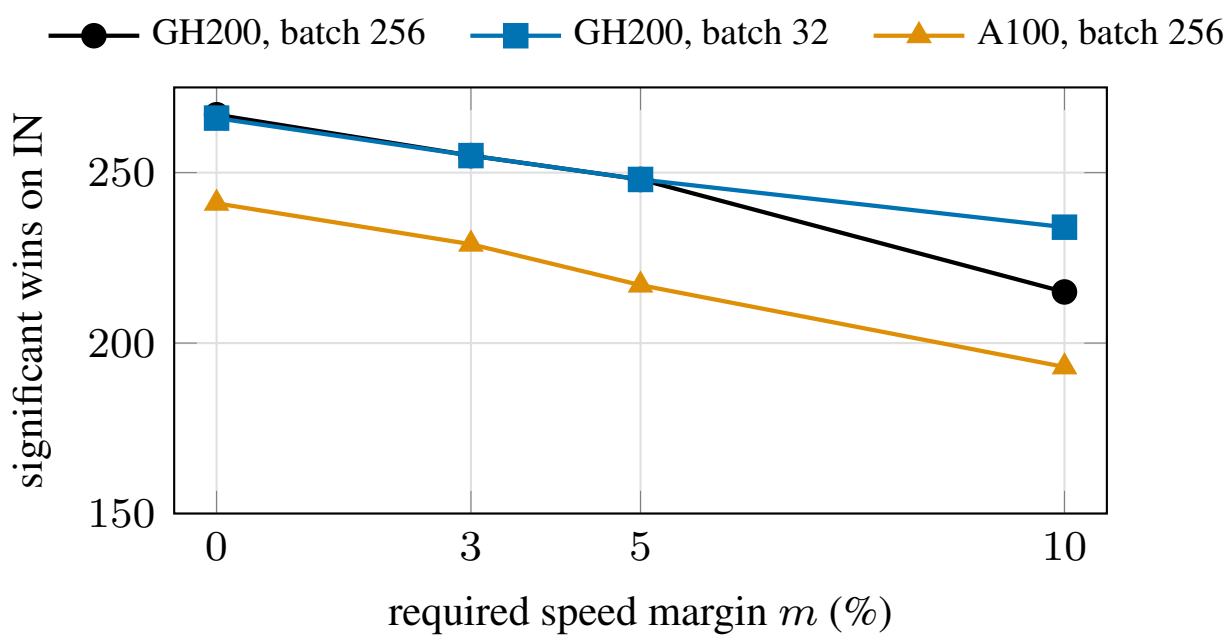
\begin{figure}[t]
  \centering
  \begin{tikzpicture}
    \begin{axis}[
        width=0.93\linewidth, height=4.4cm, xmin=-0.5, xmax=10.5, ymin=150, ymax=275,
        xlabel={required speed margin $m$ (\%)}, ylabel={significant wins on \IN},
        xlabel style={font=\footnotesize}, ylabel style={font=\footnotesize},
        xtick={0,3,5,10}, ticklabel style={font=\footnotesize}, grid=major, grid style={black!12},
        legend style={font=\scriptsize, at={(0.5,1.03)}, anchor=south, draw=none, legend columns=-1,
          /tikz/every even column/.append style={column sep=6pt}}, legend cell align=left]
      \addplot[thick, mark=*, black] coordinates {(0,267) (3,255) (5,248) (10,215)};
      \addplot[thick, mark=square*, cbblue] coordinates {(0,266) (3,255) (5,248) (10,234)};
      \addplot[thick, mark=triangle*, cborange] coordinates {(0,241) (3,229) (5,217) (10,193)};
      \legend{{\gh, batch 256}, {\gh, batch 32}, {\ahun, batch 256}}
    \end{axis}
  \end{tikzpicture}
  \caption{Significant wins of \method on \IN when our schedule must be faster than the competitor by a margin $m$.
    Significant losses at $m=0$, 3, 5 and 10\%: 1, 1, 2, 3 (\gh, batch 256); 2, 2, 4, 6 (\gh, batch 32); and 0, 0, 0, 3
    (\ahun).}
  \label{fig:margin}
\end{figure}

\paragraph{Pairing with a speed margin.}
\label{sec:supp:margin}
\Cref{fig:margin} repeats the pairing of \cref{M-tab:main}, but our schedule must now be faster than the competitor by a
margin $m$ (\cref{M-sec:exp:robust}). Our comparisons concern the implementations we measured. A faster kernel for any
operator would change the latency table, and the calibration would follow it.

\paragraph{The step after the last attention.}
\label{sec:supp:lastblock}
On class-token models, a step after the attention of the last block removes patch tokens before its MLP, since only the
CLS token's output reaches the head. The calibration uses this step in 12 of 12 DeiT-B schedules at batch 256 (and on the
\ahun), 6 of 10 on DeiT-S, 8 of 9 on CLIP ViT-B/16 and 12 of 12 on CLIP ViT-L/14, but in only one schedule at batch 32.
We measured our schedules without these steps in the same job as the originals. They run 1.0--7.5\% slower on DeiT-B,
0.6--6.3\% on CLIP ViT-B/16 and at most 1.1\% on DeiT-S and CLIP ViT-L/14. On the same GPU, they classify every image of
the five test sets in the same way as the originals (on \IN we also checked this for one \ahun DeiT-B schedule, at
$1.66\times$).

We then scale each schedule's speed in the main job by its measured speed ratio without the step and pair again. At batch 256 on the \gh, 32 of 314
pairs change partner. The significant wins and losses on \IN stay at 267 and 1, and those on \INA change from 241 and 6
to 237 and 5. On CLIP they change from 123 and 0 to 118 and 0 on \IN, and from 103 and 3 to 97 and 6 on \INA. On the
\ahun, the DeiT-B schedules run 0.5--2.8\% slower without the step; 18 of 279 pairs change partner, and the significant
wins and losses stay at 241 and 0 on \IN and at 232 and 3 on \INA. Against the envelope (\cref{M-tab:envelope}), the
median gain changes only on two models. On DeiT-B it goes from $+0.83$ to $+0.76$\pp on the \gh and from
$+0.81$ to $+0.68$\pp on the \ahun. On CLIP ViT-B/16 it goes from $+0.77$ to $+0.47$\pp, with 15 instead of 18
significant wins. There is still no significant loss.

\begin{figure*}[t]
  \centering
  \includegraphics[width=\textwidth]{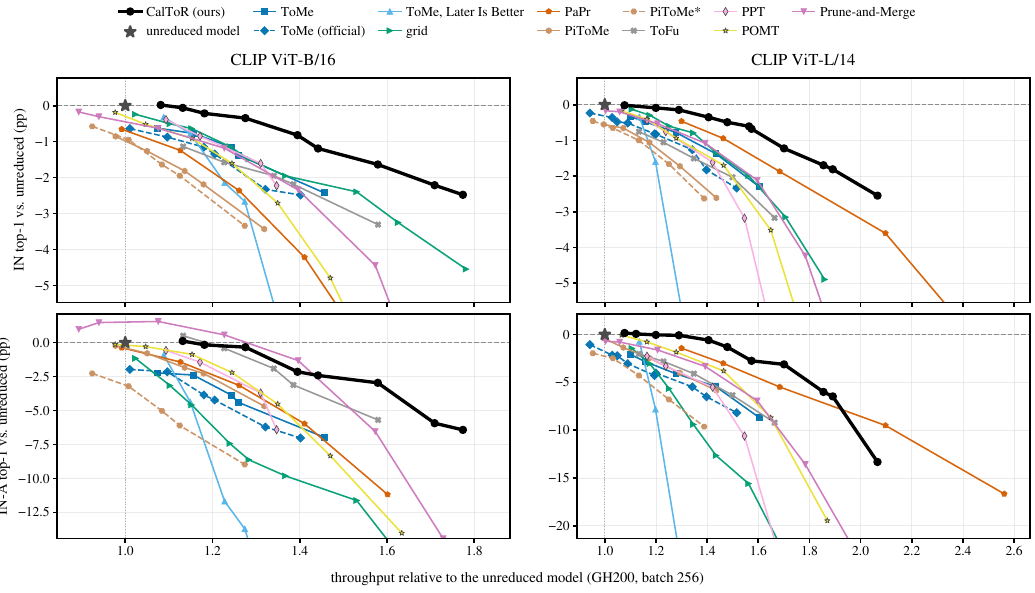}
  \caption{Fronts of CLIP ViT-B/16 and ViT-L/14 on \IN (top) and \INA (bottom), \gh at batch 256. Black, thick: \method
    (ours). The dashed line marks the unreduced model.}
  \label{fig:supp:frontsclip}
\end{figure*}

\paragraph{CLIP.}
\label{sec:supp:clip}
\Cref{fig:supp:frontsclip} shows the fronts of the two CLIP models on \IN and \INA. For ViT-B/16 we use the competitor
list of DeiT-B (the same depth, tokens and MACs). For ViT-L/14 we use ToMe, Later Is Better, PiToMe, ToFu and PPT at
50--75\% of its MACs, and the other families at 30--94\%. DiffRate is left out, as its rates come from a search per
model. PiToMe* is the DeiT variant, which is also the better of the two on CLIP (83.26 against 83.02\% on ViT-B/16 at
10\,GMACs). The CLIP inputs are normalized with CLIP's own mean and standard deviation; PaPr's ConvNet receives them
renormalized to the ImageNet statistics. On \INA, the three significant losses are to Prune-and-Merge configurations on
ViT-B/16 at $0.89$--$1.08\times$. ToFu at $1.13\times$ and Prune-and-Merge at $1.23$ and $1.40\times$ lie above our
front, without a significant difference.

\paragraph{DINOv2.}
\label{sec:supp:dinov2}
After CLIP, we add a self-supervised backbone, again with the recipe unchanged: DINOv2 ViT-B/14~\cite{dinov2} (timm,
LVD-142M) with the linear ImageNet-1k classifier from its release. This classifier is one linear layer on the final class
token concatenated with the mean of the final patch tokens. We run the model at $224\times224$, i.e.\ 256 patches of 14
pixels, with the position embedding resampled from $37\times37$. Its \IN top-1 of 84.29\% is close to the 84.5\%
reported for this classifier. Our engines and the competitor ports weight the patch mean by the token sizes after merges.

The competitor lists follow those of CLIP ViT-L/14, which also has 256 patches: ToMe, Later Is Better, PiToMe, ToFu and
PPT at 50--75\% of the 23.2\,GMACs of DINOv2 ViT-B/14, and the other families at the token counts of CLIP ViT-L/14,
again without DiffRate. The calibration uses the recipe of DeiT and CLIP and measures every damage curve afresh. Its
targets are 1.1--1.75 times the predicted unreduced speed and, as on CLIP, 1.05 times the speed of competitors without
a close pair. All schedules are measured in one frame on the \gh at batch 256, where the unreduced model runs at
5\,605\,img/s and ours at 1.05--1.78$\times$.

\Cref{tab:supp:dinov2} gives the pairs. \method is more accurate in all 60 pairs on \IN (55 significant wins, no loss,
median $+1.41$\pp). On \INA it is less accurate in two pairs, both against PaPr, one of them significantly. Four
configurations at 1.80--2.11$\times$ stay unpaired and lose 5.9--78.1\pp of \IN top-1: the grid prune to 128 tokens,
PaPr at 30\% and Prune-and-Merge at $K=0.4$ and $0.25$. \Cref{fig:supp:tokensD} shows the tokens of one pair (ToMe at
$1.48\times$, \method at $1.51\times$), as \cref{fig:supp:tokens,fig:supp:tokensL} do for DeiT-B and MAE ViT-L.

\begin{table}[t]
  \centering
  \footnotesize
  \setlength{\tabcolsep}{4pt}
  \caption{DINOv2 ViT-B/14 with its linear ImageNet-1k classifier, \gh at batch 256: \method (ours) against the
    competitors of CLIP at equal measured speed, as in \cref{M-tab:main}. Each competitor configuration is paired with our
    slowest schedule that is at least as fast. Significance: McNemar's exact test with Holm's correction over the 60 pairs of
    a test set.}
  \label{tab:supp:dinov2}
  \begin{tabular}{@{}lccccc@{}}
    \toprule
    & \IN & \INV & \INR & \INA & \INS \\
    \midrule
    unreduced top-1 (\%) & 84.29 & 75.03 & 66.52 & 57.32 & 52.53 \\
    \method better / pairs & 60/60 & 60/60 & 60/60 & 58/60 & 60/60 \\
    significant wins / losses & \textbf{55}/0 & \textbf{51}/0 & \textbf{59}/0 & \textbf{58}/1 & \textbf{60}/0 \\
    median difference (pp) & $+1.41$ & $+1.99$ & $+3.40$ & $+9.21$ & $+3.33$ \\
    \bottomrule
  \end{tabular}
\end{table}

\begin{table*}[t]
  \centering
  \footnotesize
  \renewcommand{\arraystretch}{0.9}
  \caption{\method (ours) against each video competitor on a \gh at batch 8, at equal measured throughput.
    $\Delta$ = top-1 of \method $-$ top-1 of the competitor (mean $\pm$ standard error over the validation videos);
    positive values favor \method. Bold, \sig: significant after Holm correction over all 28 pairs. The header gives the
    unreduced model's top-1 and throughput.}
  \label{tab:video}
  \begin{tabular}{@{}lr@{\hspace{14pt}}lr@{\hspace{14pt}}lr@{}}
    \toprule
    \multicolumn{2}{c}{K400, VideoMAE-B (81.39\%, 859 clips/s)} & \multicolumn{2}{c}{SSv2, VideoMAE-B (70.57\%, 861 clips/s)} & \multicolumn{2}{c}{K400, VideoMAE-L (85.17\%, 297 clips/s)} \\
    \cmidrule(r){1-2}\cmidrule(lr){3-4}\cmidrule(l){5-6}
    competitor & $\Delta$ (pp) & competitor & $\Delta$ (pp) & competitor & $\Delta$ (pp) \\
    & {\scriptsize \method$-$comp.} & & {\scriptsize \method$-$comp.} & & {\scriptsize \method$-$comp.} \\
    \midrule
    RLT, 58.7\,G & $-0.22\pm0.09$ & RLT, 58.7\,G & {\boldmath$+1.33\pm0.18$}\sig & PaPr (X3D-S), 30\% & {\boldmath$+0.69\pm0.13$}\sig \\
    RLT, 120\,G & $-0.07\pm0.06$ & RLT, 120\,G & $-0.06\pm0.12$ & PaPr (X3D-M), 30\% & {\boldmath$+0.67\pm0.12$}\sig \\
    PaPr (X3D-S), 35\% & {\boldmath$+1.04\pm0.14$}\sig & STA, $r{=}64$ & {\boldmath$+0.40\pm0.13$}\sig & PaPr (X3D-S), 249\,G & $+0.29\pm0.10$ \\
    PaPr (X3D-S), 45\% & {\boldmath$+0.45\pm0.12$}\sig & STA, $r{=}48$ & $+0.26\pm0.10$ & RLT, 274\,G & $+0.00\pm0.05$ \\
    STA, $r{=}64$ & {\boldmath$+0.46\pm0.12$}\sig & STA, $r{=}32$ & $+0.08\pm0.09$ & PaPr (X3D-S), 50\% & $+0.04\pm0.09$ \\
    STA, $r{=}48$ & $+0.25\pm0.10$ & ToMe, $r{=}64$ & $+0.00\pm0.09$ & STA, $r{=}64$ & $+0.14\pm0.10$ \\
    PaPr (X3D-M), 35\% & {\boldmath$+0.91\pm0.14$}\sig & PaPr (X3D-M), 35\% & {\boldmath$+1.62\pm0.17$}\sig & ToMe, $r{=}64$ & {\boldmath$+0.81\pm0.13$}\sig \\
    PaPr (X3D-M), 45\% & {\boldmath$+0.43\pm0.12$}\sig & PaPr (X3D-M), 45\% & {\boldmath$+0.77\pm0.15$}\sig & PaPr (X3D-M), 249\,G & $+0.13\pm0.10$ \\
    & & & & PaPr (X3D-M), 50\% & $+0.21\pm0.09$ \\
    & & & & STA, $r{=}48$ & $+0.04\pm0.08$ \\
    & & & & RLT, 405\,G & $+0.13\pm0.07$ \\
    & & & & STA, $r{=}32$ & $+0.10\pm0.07$ \\
    \bottomrule
  \end{tabular}
\end{table*}

\begin{figure*}[t]
  \centering
  \includegraphics[width=\textwidth]{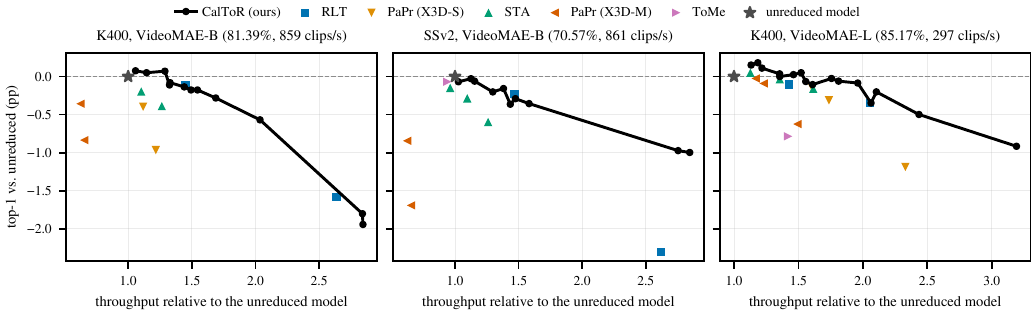}
  \caption{Video on a \gh at batch 8: top-1 on the full validation set relative to the unreduced model (one y axis for the
    three settings) against the throughput relative to it. In brackets: the unreduced model's top-1 and clips/s. Black,
    thick: \method (ours) with the input-level prune.}
  \label{fig:supp:video}
\end{figure*}

\paragraph{Video fronts.}
\label{sec:supp:video}
\Cref{fig:supp:video} shows top-1 relative to the unreduced model against the measured throughput, for the three
settings of \cref{tab:video}. After Holm correction, no competitor configuration is significantly more accurate than its
pair. Near RLT on K400, neither method is significantly better (\cref{tab:video}). The residual size measure of \PPS does
not help on video: after block~0 it trails the best fixed score (the temporal change on K400, the grid on SSv2) by
0.10--0.45\pp. We therefore leave it out of the video menu.

\section{Schedules, latency and damage}
\label{sec:supp:calib}
\begin{figure*}[t]
  \centering
  \includegraphics[width=\textwidth]{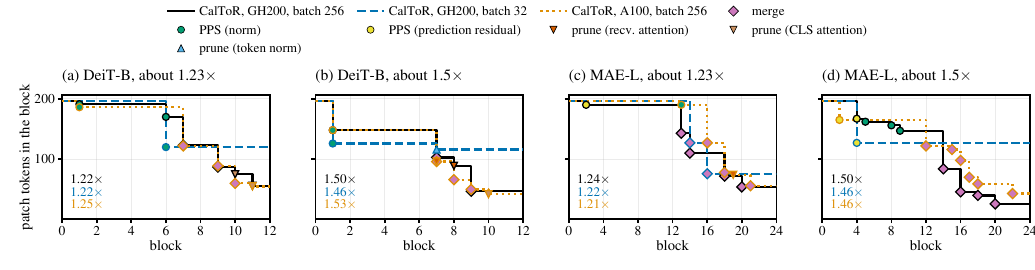}
  \caption{Patch tokens per block under the schedule of \method closest to $1.23\times$ and $1.5\times$ in each setting
    (measured speed-ups in the corner), with the operator of each step marked. DeiT-B (a, b) and MAE ViT-L (c, d).}
  \label{fig:supp:schedules}
\end{figure*}

\paragraph{Schedules across GPUs and batch sizes.}
\label{sec:supp:schedules}
\Cref{fig:supp:schedules} compares the schedules of \method in our three settings, each at about the same measured
speed-up. At batch 256, the calibration spreads the reduction over five to eight steps. These steps mix \PPS with
merges and, on DeiT-B, with attention prunes. At batch 32, a step costs two to five times more relative to a block
(\cref{tab:fixedcost}). At batch 32, every schedule in the figure has only one or two steps: a single \PPS step (DeiT-B at
$1.22\times$, MAE ViT-L at $1.46\times$), \PPS and a norm prune (DeiT-B at $1.46\times$), or two large merges (MAE
ViT-L at $1.22\times$). On the \ahun, the schedules at batch 256 resemble those on the \gh, but the steps in the middle
of the network differ. The schedule is thus a property of the GPU and the batch size (\cref{M-sec:intro}).

\begin{table}[t]
  \centering
  \caption{Kernel cost of one reduction step on a \gh, in units of the time of a full block, at the full token count
    (prunes keep 50\%). A step that reads an attention statistic pays for it in the attention of its block: 0.05--0.14
    of a block at batch 256 and 0.12--0.24 at batch 32 (CLS attention on DeiT, received attention on MAE). Costs grow
    at smaller batches and on smaller models.}
  \label{tab:fixedcost}
  \resizebox{\linewidth}{!}{%
  \begin{tabular}{@{}lccc@{}}
    \toprule
    batch, model & our merge ($\rho{=}25\%$) & grid prune & CLS-attn.\ prune \\
    \midrule
    256, DeiT-S & 0.59 & 0.13 & 0.09 \\
    256, DeiT-B / MAE-B & 0.36 & 0.06 & 0.05 \\
    256, MAE-H & 0.23 & 0.03 & 0.03 \\
    32, DeiT-S & 1.57 & 0.61 & 0.37 \\
    32, DeiT-B / MAE-B & 0.88 & 0.30 & 0.19 \\
    32, MAE-H & 0.44 & 0.11 & 0.08 \\
    \bottomrule
  \end{tabular}}
\end{table}

\begin{figure*}[t]
  \centering
  \includegraphics[width=\textwidth]{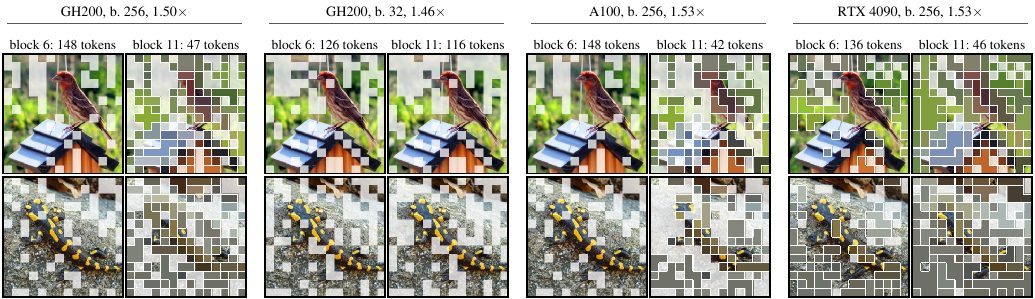}
  \caption{Tokens entering blocks 6 and 11 of DeiT-B under the schedule of \method (ours) closest to $1.5\times$ in each
    setting (measured speed-ups in the titles), drawn as in \cref{fig:supp:tokens}.}
  \label{fig:supp:settings}
\end{figure*}

\Cref{fig:supp:settings} shows the DeiT-B schedules of \cref{fig:supp:schedules} at about $1.5\times$, together with
the one calibrated on the RTX~4090 (\cref{sec:supp:transfer}), on two held-out training images. At batch 256 on the \gh
and the \ahun, \method samples 148 patches in block~0 and merges and prunes in blocks 6--9; 42--47 tokens enter the last
block. On the RTX~4090, it keeps 175 patches in block~0 and merges in blocks 4--8. It does not prune before the last
block, which 46 tokens enter. At batch 32, the same speed-up comes from two prunes (\PPS in block~0 and a norm prune in
block~6), and 116 tokens enter the last block. A competitor removes the same tokens on every GPU and at every batch
size. The exception is POMT, which picks its reduction amount per device (we sweep that amount).

\paragraph{Batch 1.}
\label{sec:supp:batch1}
In measurement jobs at batch 1 (one pass), we timed competitors, grid prunes and schedules from an earlier version of
our recipe. On the \gh, no configuration speeds up DeiT-S. The fastest configuration gains 2.2\% on DeiT-B, 0.5\% on MAE
ViT-B, 5.4\% on MAE ViT-L and 14.7\% on MAE ViT-H. On the \ahun, the fastest configuration gains 0.8\% (DeiT-S), 11.3\%,
11.9\%, 19.8\% and 46.2\% (MAE ViT-H), with the models in the same order. At batch 1, token reduction thus pays off
mainly on large models, and more on the \ahun than on the \gh.

\begin{figure*}[t]
  \centering
  \includegraphics[width=\textwidth]{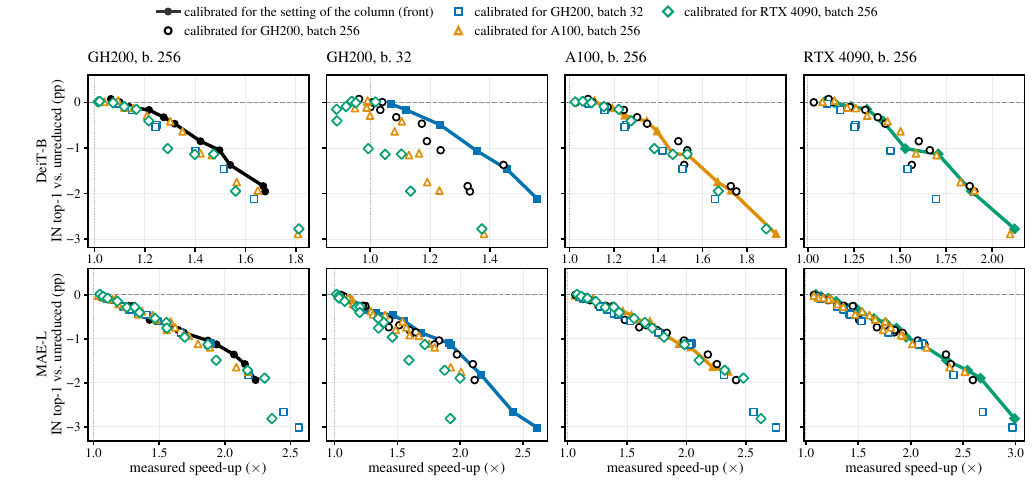}
  \caption{Schedules of \method calibrated for one setting (marker color) and measured in another (column): top-1 on
    \IN relative to the unreduced model against the measured speed-up there. The line is the front of the schedules
    calibrated for the setting of the column. DeiT-B (top) and MAE ViT-L (bottom).}
  \label{fig:supp:transfer}
\end{figure*}

\paragraph{Transfer between settings.}
\label{sec:supp:transfer}
We measure the final schedules calibrated for each setting in the jobs of the other settings, next to that setting's
own schedules (\cref{tab:supp:transfer,fig:supp:transfer}). For the RTX~4090, we calibrate the final recipe on its
latency table at the speeds that the schedules of the other settings reach there. A transferred schedule misses its
speed. At batch 32 on the \gh, schedules calibrated at batch 256 reach a median of 72--97\% of their speed-up (per
model), and schedules calibrated on the RTX~4090 reach 45--92\%. On the \ahun, schedules calibrated on the RTX~4090
reach 83--98\%. At the speed that a transferred schedule actually reaches, the schedules calibrated for the setting are
more accurate. The gap is up to 1.67\pp at batch 32. Against schedules from the RTX~4090, it is up to 0.37\pp on the
\gh at batch 256 and 0.22\pp on the \ahun. Against schedules from batch 32, it is up to 0.32\pp on the \gh at batch
256, 0.41\pp on the \ahun and 0.82\pp on the RTX~4090. Between the \ahun and the \gh at batch 256 (in both
directions), and from either of them to the RTX~4090, the schedules transfer within 0.11\pp. The effect shrinks as the
model grows: on MAE ViT-L and ViT-H it is below 0.4\pp everywhere.

\begin{table*}[t]
  \centering
  \footnotesize
  \setlength{\tabcolsep}{3pt}
  \caption{Schedules of \method run in another setting. Each entry is the top-1 on \IN of the schedules calibrated for
    that setting, interpolated at the measured speed of the transferred schedule, minus that of the transferred
    schedule (pp, mean over the transferred schedules). In brackets: the median ratio of the speed-up a transferred
    schedule reaches there to the speed-up it reached where it was calibrated. --: no transferred schedule within the
    range of the setting's own schedules.}
  \label{tab:supp:transfer}
  \begin{tabular}{@{}lccccc@{}}
  \toprule
  calibrated for $\to$ run on & DeiT-S & DeiT-B & MAE-B & MAE-L & MAE-H \\
  \midrule
  \gh, 32 $\to$ \gh, 256 & $+0.32$ (1.16) & $+0.24$ (1.03) & $-0.11$ (1.02) & $+0.02$ (1.00) & $+0.01$ (1.01) \\
  \ahun, 256 $\to$ \gh, 256 & $+0.10$ (0.96) & $+0.11$ (0.97) & $+0.04$ (0.98) & $+0.02$ (0.94) & $+0.01$ (0.97) \\
  RTX 4090, 256 $\to$ \gh, 256 & $+0.37$ (0.79) & $+0.19$ (0.86) & $+0.23$ (0.89) & $+0.03$ (0.89) & $+0.04$ (0.92) \\
  \addlinespace
  \gh, 256 $\to$ \gh, 32 & $+1.67$ (0.72) & $+0.46$ (0.87) & $+0.37$ (0.90) & $+0.08$ (0.97) & $+0.03$ (0.97) \\
  \ahun, 256 $\to$ \gh, 32 & -- (0.70) & $+1.01$ (0.80) & $+0.51$ (0.87) & $+0.13$ (0.91) & $+0.05$ (0.95) \\
  RTX 4090, 256 $\to$ \gh, 32 & -- (0.45) & $+1.45$ (0.65) & $+0.90$ (0.75) & $+0.39$ (0.80) & $+0.06$ (0.92) \\
  \addlinespace
  \gh, 256 $\to$ \ahun, 256 & $-0.03$ (1.03) & $-0.04$ (1.02) & $-0.03$ (1.02) & $-0.03$ (1.05) & $-0.02$ (1.02) \\
  \gh, 32 $\to$ \ahun, 256 & $+0.41$ (1.15) & $+0.22$ (1.03) & $-0.08$ (1.03) & $+0.02$ (1.05) & $-0.01$ (1.02) \\
  RTX 4090, 256 $\to$ \ahun, 256 & $+0.22$ (0.83) & $+0.07$ (0.91) & $+0.12$ (0.92) & $-0.01$ (0.94) & $-0.01$ (0.98) \\
  \addlinespace
  \gh, 256 $\to$ RTX 4090, 256 & $+0.07$ (1.18) & $-0.01$ (1.08) & $-0.06$ (1.08) & $+0.04$ (1.10) & $0.00$ (1.04) \\
  \gh, 32 $\to$ RTX 4090, 256 & $+0.82$ (1.31) & $+0.41$ (1.05) & $+0.02$ (1.05) & $+0.17$ (1.07) & $+0.03$ (1.03) \\
  \ahun, 256 $\to$ RTX 4090, 256 & $+0.01$ (1.13) & $0.00$ (1.07) & $-0.02$ (1.06) & $+0.04$ (1.04) & $0.00$ (1.02) \\
  \bottomrule
  \end{tabular}
\end{table*}

\begin{figure*}[t]
  \centering
  \includegraphics[width=\textwidth]{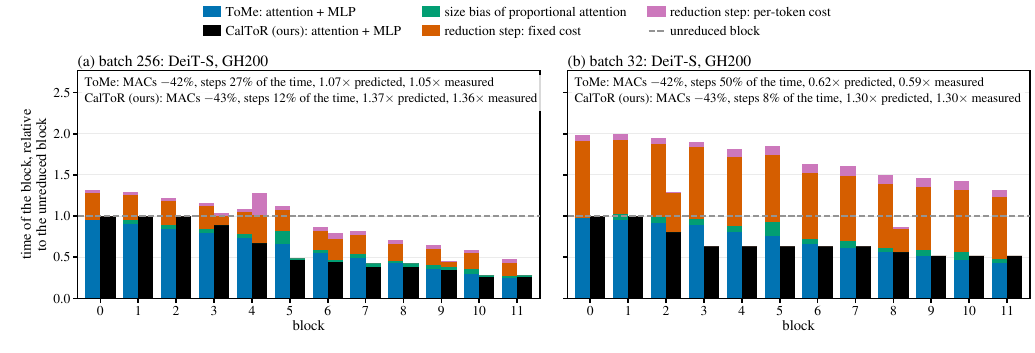}
  \caption{The forward pass of DeiT-S on a \gh, block by block, as the latency table prices it. ToMe and \method (ours)
    run at about the same MACs. Each block is shown relative to the same block of the unreduced model (dashed line) and
    split into attention and MLP, the size bias of proportional attention and the reduction steps (fixed cost: the cost
    of the smallest step; per-token cost: the rest). Embedding and readout are not drawn. Each panel gives the MACs, the
    share of the steps in the predicted time, and the predicted and measured speed-ups.}
  \label{fig:supp:time}
\end{figure*}

\paragraph{Where the time goes.}
\label{sec:supp:time}
\Cref{fig:supp:time} splits the forward pass of DeiT-S on the \gh into the parts that the latency table prices:
attention, MLP, the reduction steps and the size bias of proportional attention. ToMe and \method run at about the
same MACs, 42--43\% fewer than the unreduced model. At batch 256, the steps of ToMe take 27\% of its time, and ToMe
runs $1.05\times$ faster than the unreduced model. The steps of \method take 12\% of its time, and it runs
$1.36\times$ faster. At batch 32, matching and merging take half of the time of ToMe, mostly as fixed cost, and ToMe
runs at $0.59\times$. \method removes its tokens in two prunes that take 8\% of its time, and it runs at $1.30\times$.
Block by block, each merge of ToMe at batch 32 costs 0.84--1.00 of an unreduced block. As a result, all blocks of ToMe
take 1.3--2.0 times as long as those of the unreduced model. The two prunes of \method cost 0.49 and 0.31 of a block,
and its last three blocks run in 0.52 of the unreduced time. The table predicts these speed-ups within 1.2\%, and that
of ToMe at batch 32 within 6\%.

\begin{figure*}[t]
  \centering
  \includegraphics[width=\textwidth]{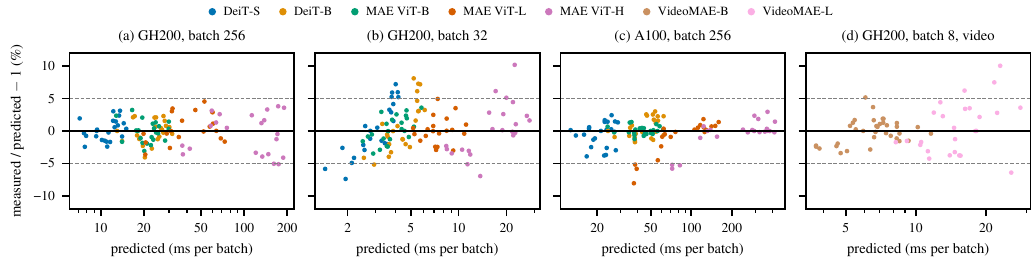}
  \caption{Error of the latency table on check schedules timed end to end, after one global scale per table fitted on the
    same schedules. Dashed: $\pm5\%$.}
  \label{fig:supp:lut}
\end{figure*}

\begin{figure*}[t]
  \centering
  \includegraphics[width=\textwidth]{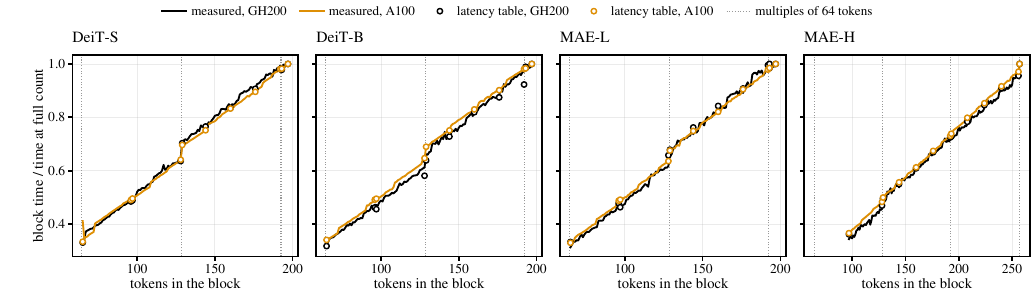}
  \caption{Time of one block against the token count at batch 256, relative to the time at the full count. Lines:
    measured at every count (the first point of each scan, 64 tokens or 96 on MAE ViT-H, is a warm-up and is left out).
    Circles: the latency table at its grid points.}
  \label{fig:supp:tiles}
\end{figure*}

\paragraph{Accuracy of the latency table.}
\label{sec:supp:lut}
We check each latency table on held-out schedules: competitor configurations, earlier schedules of ours and random
schedules. \Cref{fig:supp:lut} shows the relative error after one global scale factor.
The mean absolute error is 1.3--2.7\% on the \gh at batch 256, 1.5--3.5\% at batch 32, 0.5--1.8\% on the \ahun and
1.5--3.3\% for video. Single schedules deviate by up to 5.1\%, 10.2\%, 8.1\% and 10.1\%, respectively.
The pairing of our schedules with competitor configurations uses measured speeds, so these errors only move a schedule
away from its target. They do not enter the comparison.
The time of a block is a step function of the token count (\cref{fig:supp:tiles}). Going from 128 to 129 tokens costs
6.5--9.3\% of the block time on DeiT-S, DeiT-B and MAE ViT-L, and 3.5--4.4\% on MAE ViT-H. On MAE ViT-H, the one token
above 256 (257 tokens) also costs 2.5\% (\ahun) and 5.7\% (\gh). The table therefore has grid points on both sides of
64, 96, 128 and 192 tokens. Measured in a separate run, its block times differ from a scan of every token count by a
median of 0.1--1.9\%.

\paragraph{Calibration cost.}
\label{sec:supp:cost}
\Cref{tab:cost} lists the cost of the latency table and of the calibration of the base targets for the DeiT and MAE
models, at both batch sizes on the \gh (\cref{M-sec:exp:cost}). Our calibration uses caches. They keep the predictions of
the unreduced model on each image set. In the refinement, they also let the forward pass of a candidate start at the
first block whose operators differ from those of the current schedule, from the stored tokens entering that block. The
engine is deterministic. The 22 image calibrations behind our cost numbers therefore give the same 1331 schedules, with
the same KLs and predicted times, with and without the caches, and the same as in our experiments. With the caches,
the fast targets (added above the fastest base target of DeiT-S, MAE ViT-L and MAE ViT-H) add 1.0--10.6 minutes
(2.0--13.1 without). The extra targets at the competitors' speeds (\cref{M-sec:exp:setup}) add 0.8--11.2 minutes
(1.8--14.6 without). Over the 22 calibrations, the caches save up to 52\% of the time; one calibration, MAE ViT-B at
batch 32, is 10\% slower. On video, the targets at the competitors' speeds add 1.8 and 4.5 hours on K400 (VideoMAE-B
and -L) and 1.3 hours on SSv2, whose five base targets take 1.8 hours.

\begin{table}[t]
  \centering
  \footnotesize
  \caption{Cost of \method on one \gh, in minutes: the latency table (measurement and check) and the calibration of the
    base targets (their number in brackets), with / without the lossless caches of our code. The two variants were run
    one after the other on the same GPU and give the same schedules. \method needs no training.}
  \label{tab:cost}
  \begin{tabular}{@{}lcccc@{}}
    \toprule
    & \multicolumn{2}{c}{batch 256} & \multicolumn{2}{c}{batch 32} \\
    model & table & calibration & table & calibration \\
    \midrule
    DeiT-S & 1.7 & 1.8 / 2.9 (6) & 1.1 & 2.1 / 2.1 (6) \\
    DeiT-B & 2.0 & 2.1 / 3.6 (8) & 1.3 & 2.2 / 2.5 (6) \\
    MAE ViT-B & 1.9 & 2.2 / 2.2 (4) & 1.3 & 2.3 / 2.1 (6) \\
    MAE ViT-L & 2.5 & 4.1 / 4.7 (4) & 1.4 & 3.8 / 5.6 (6) \\
    MAE ViT-H & 4.6 & 15.9 / 20.2 (10) & 2.2 & 12.0 / 13.9 (6) \\
    \bottomrule
  \end{tabular}
\end{table}

\begin{figure*}[t]
  \centering
  \includegraphics[width=\textwidth]{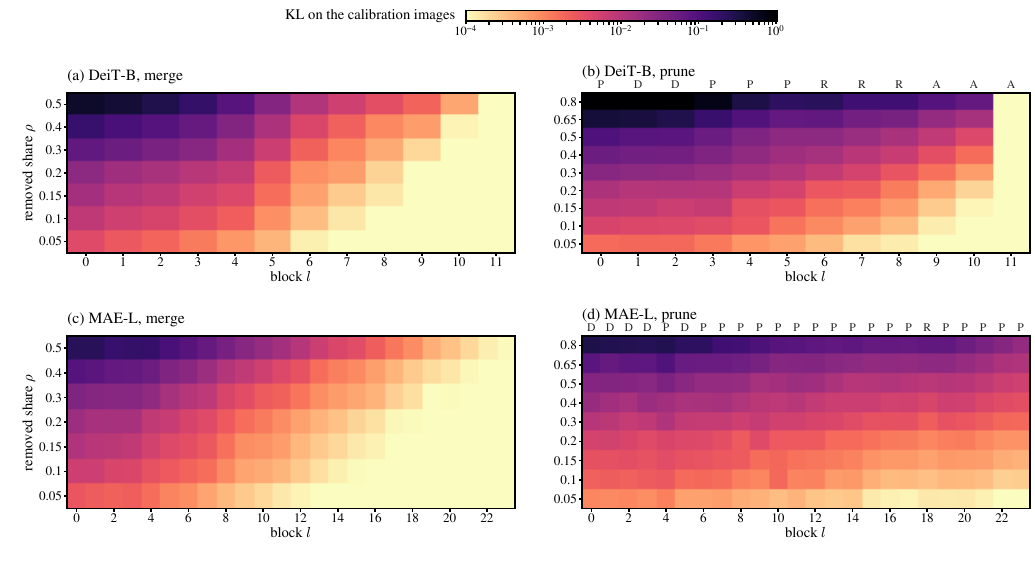}
  \caption{Damage curves of \method on a \gh at batch 256: KL on the 512 calibration images after removing a share
    $\rho$ of the tokens in one block, by a merge (a, c) or by the prune of the calibrated menu (b, d). Letters above
    (b, d) give the prune score picked for each block (P: \PPS with the norm, D: \PPS with the prediction residual, R:
    received attention, A: CLS attention); a merge has no score. The head of DeiT-B reads only the CLS token, so a
    reduction after the attention of the last block leaves its output unchanged (KL 0, \cref{sec:supp:lastblock}). MAE
    ViT-L averages the tokens.}
  \label{fig:supp:damage}
\end{figure*}

\paragraph{Damage curves.}
\label{sec:supp:damage}
\Cref{fig:supp:damage} shows the damage curves of the final calibration of DeiT-B and MAE ViT-L on the \gh at batch
256. The damage is the KL on the calibration images after removing a share $\rho$ of the tokens in one block.
At $\rho=0.2$, the damage of a merge falls by more than two orders of magnitude from block~0 to the second-to-last
block, on both models. The damage of a prune falls 54-fold on DeiT-B but only 6-fold on MAE ViT-L. There, an early
\PPS step costs 3.5 times less than an early merge. The letters in the figure give the prune score that the
calibration picked for each block, at equal predicted time saving. On DeiT-B, it picks \PPS in blocks 0--5, the
received attention in blocks 6--8 and the CLS attention in blocks 9 and 10 (in block~11 every score has damage 0). In
the first blocks, no cheap score predicts the influence of a token, so sampling beats a top-$k$ selection; the
attention becomes informative only with depth. On MAE ViT-L, it picks the prediction-residual size in the first
blocks. The curves are convex, and at $\rho=0.3$ they lie two to four times below those of a random prune
(\cref{prop:random}).

\begin{figure*}[t]
  \centering
  \includegraphics[width=\textwidth]{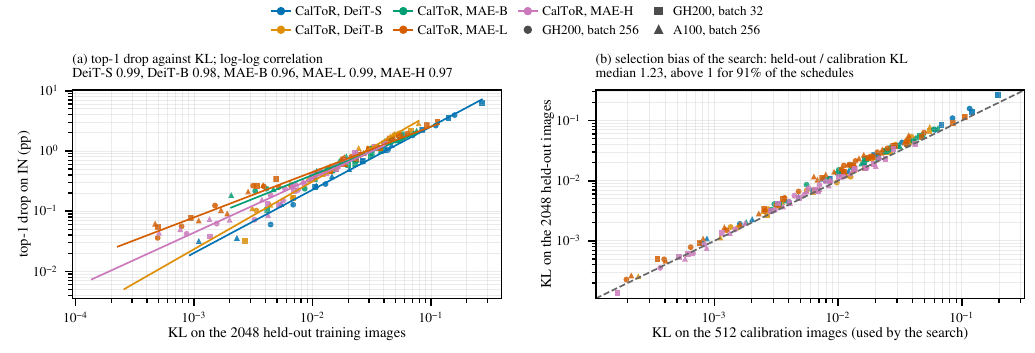}
  \caption{(a) Top-1 drop on \IN of every final schedule of \method that loses more than 0.02\pp (171 of 186), against
    its KL on the 2048 held-out training images of the final pick (log-log; lines: per-model fits). (b) That KL against
    the KL on the 512 calibration images used by the search; dashed: equality.}
  \label{fig:supp:kl}
\end{figure*}

\paragraph{Calibration objective and test accuracy.}
\label{sec:supp:kl}
\Cref{fig:supp:kl} relates the KL that \method minimizes to the top-1 that we report. The KL is measured on the 2048
images of the final pick. We take all 171 final schedules of the three settings that lose more than 0.02\pp. For these,
the top-1 drop on \IN follows a power law in the KL, with a log-log correlation of 0.96--0.99 per model.
\Cref{prop:pinsker} bounds the drop by the KL but does not imply this law. The KL that the search sees on its 512
calibration images is optimistic: on the 2048 images of the final pick, it is higher by a median factor of 1.23
(\cref{prop:holdout}).

\section{Competitor ports and published numbers}
\label{sec:supp:comp}
\paragraph{Fidelity of the competitor ports.}
\label{sec:supp:fidelity}
\Cref{tab:supp:fidelity} evaluates our ports at the operating points and in the numerical precision of each paper.
ToFu, PPT and POMT match their papers within 0.6\pp, and DiffRate within 0.14\pp.
The official PiToMe code falls short of its paper by 0.72\pp on DeiT-S, 1.86\pp on MAE ViT-L and 1.01\pp on DeiT-T.
Besides the official code, we tried up to nine variants of it. The best variant per model, PiToMe*, uses the margin $m_0=0.9$ with proportional attention
on DeiT, and bipartite soft matching (BSM) in every block on MAE ViT-L. It still falls 0.44--0.53\pp short.
Prune-and-Merge reports results only on ImageNet-100. At its keep rates 0.9--0.5, our port loses 0.24--1.90\pp relative to the unreduced
DeiT-S and 0.24--3.36\pp relative to the unreduced DeiT-B; its paper reports losses of 0.32--2.02 and 0.08--3.92\pp.

\begin{table*}[t]
  \centering
  \footnotesize
  \caption{Our ports of the competitors (not \method) at the operating points of their papers. We report \IN top-1 on the full
    validation set, in the precision of the official evaluation. POMT reports the loss of top-1, so its published top-1 is our unreduced model
    minus that loss. Prop.\ attention: proportional attention. DiffRate: the rates of its official release with its
    explicit attention; our tables use the same rates with fused attention (within 0.14\pp).}
  \label{tab:supp:fidelity}
  \begin{tabular}{@{}lllrrrr@{}}
    \toprule
    model & method & variant & GMACs & top-1 & published & $\Delta$ \\
    \midrule
    DeiT-T & ToFu & PiToMe code & 0.78 & 69.01 & 69.60 & $-$0.59 \\
     & PiToMe & official code & 0.79 & 69.79 & 70.80 & $-$1.01 \\
     & & BSM in all blocks & 0.79 & 69.14 & 70.80 & $-$1.66 \\
     & & protection in all blocks & 0.79 & 69.01 & 70.80 & $-$1.79 \\
     & & protection everywhere, prop. attention & 0.79 & 68.78 & 70.80 & $-$2.02 \\
     & & $m_0=0.9$ & 0.79 & 69.81 & 70.80 & $-$0.99 \\
     & & $m_0=0.9$, BSM everywhere & 0.79 & 69.19 & 70.80 & $-$1.61 \\
     & & $m_0=0.9$, protection in all blocks & 0.79 & 68.84 & 70.80 & $-$1.96 \\
     & & $m_0=0.9$, protection everywhere, prop. attention & 0.79 & 68.88 & 70.80 & $-$1.92 \\
     & & $m_0=0.9$, prop. attention & 0.79 & 70.36 & 70.80 & $-$0.44 \\
    \midrule
    DeiT-S & ToFu & PiToMe code & 2.82 & 78.00 & 77.80 & +0.20 \\
     & PiToMe & official code & 2.82 & 78.38 & 79.10 & $-$0.72 \\
     & & BSM in all blocks & 2.82 & 77.96 & 79.10 & $-$1.14 \\
     & & protection in all blocks & 2.82 & 77.67 & 79.10 & $-$1.43 \\
     & & protection everywhere, prop. attention & 2.82 & 77.21 & 79.10 & $-$1.89 \\
     & & $m_0=0.9$ & 2.82 & 78.25 & 79.10 & $-$0.85 \\
     & & $m_0=0.9$, BSM everywhere & 2.82 & 77.92 & 79.10 & $-$1.18 \\
     & & $m_0=0.9$, protection in all blocks & 2.82 & 77.64 & 79.10 & $-$1.46 \\
     & & $m_0=0.9$, protection everywhere, prop. attention & 2.82 & 77.26 & 79.10 & $-$1.84 \\
     & & $m_0=0.9$, prop. attention & 2.82 & 78.65 & 79.10 & $-$0.45 \\
     & PiToMe & official code, prop. attention & 2.82 & 78.64 & 79.10 & $-$0.46 \\
     & PPT & official code, fp16 & 2.94 & 79.48 & 79.50 & $-$0.02 \\
    \midrule
    MAE-L & ToFu & PiToMe code & 29.79 & 83.86 & 83.80 & +0.06 \\
     & PiToMe & official code & 30.07 & 82.74 & 84.60 & $-$1.86 \\
     & & BSM in all blocks & 30.07 & 84.07 & 84.60 & $-$0.53 \\
     & & protection in all blocks & 30.07 & 82.61 & 84.60 & $-$1.99 \\
     & & protection everywhere, prop. attention & 30.07 & 80.40 & 84.60 & $-$4.20 \\
     & & $m_0=0.9$ & 30.07 & 82.62 & 84.60 & $-$1.98 \\
     & & $m_0=0.9$, BSM everywhere & 30.07 & 83.97 & 84.60 & $-$0.63 \\
     & & $m_0=0.9$, protection in all blocks & 30.07 & 82.49 & 84.60 & $-$2.11 \\
     & & $m_0=0.9$, protection everywhere, prop. attention & 30.07 & 80.54 & 84.60 & $-$4.06 \\
     & & $m_0=0.9$, prop. attention & 30.07 & 80.88 & 84.60 & $-$3.72 \\
     & PiToMe & official code, prop. attention & 30.07 & 80.93 & 84.60 & $-$3.67 \\
    \midrule
    DeiT-S & POMT & $R=77$, max over 2 images & 3.27 & 78.60 & 78.59 & +0.01 \\
     & POMT & $R=77$, max over the batch & 3.27 & 78.60 & 78.59 & +0.01 \\
    \midrule
    DeiT-B & POMT & $R=70$, max over 2 images & 12.99 & 80.64 & 80.64 & 0.00 \\
     & POMT & $R=70$, max over the batch & 12.99 & 80.73 & 80.64 & +0.09 \\
    \midrule
    DeiT-S & DiffRate & official rates and arithmetic & 2.33 & 78.68 & 78.76 & $-$0.08 \\
     & & & 2.90 & 79.53 & 79.58 & $-$0.05 \\
    DeiT-B & DiffRate & official rates and arithmetic & 8.73 & 79.12 & 78.98 & +0.14 \\
     & & & 11.52 & 81.52 & 81.50 & +0.02 \\
    MAE-B & DiffRate & official rates and arithmetic & 8.71 & 80.00 & 79.96 & +0.04 \\
     & & & 11.52 & 82.80 & 82.91 & $-$0.11 \\
    MAE-L & DiffRate & official rates and arithmetic & 31.00 & 84.59 & 84.65 & $-$0.06 \\
     & & & 42.29 & 85.61 & 85.56 & +0.05 \\
    MAE-H & DiffRate & official rates and arithmetic & 92.11 & 86.36 & 86.40 & $-$0.04 \\
     & & & 103.34 & 86.64 & 86.72 & $-$0.08 \\
    \bottomrule
  \end{tabular}
\end{table*}

\paragraph{Comparison with published numbers.}
\label{sec:supp:published}
Zero-TPrune~\cite{zerotprune}, CATIS~\cite{catis}, RAPID~\cite{rapid}, ATM~\cite{atm}, Token
Transforming~\cite{tokentransforming}, SAD-TM~\cite{sadtm} and AdaMerge~\cite{adamerge} release no code that we could
find (the SAD-TM repository holds only a README). We therefore compare with the training-free numbers reported in their papers.
We match the computational budget, not the measured speed, because we cannot reproduce their speed measurements.
Where a paper gives GFLOPs, we scale its budgets by the ratio of the unreduced models' counts (\eg 4.60 in ours, 4.55
in Zero-TPrune). RAPID removes $r$ tokens per block, like ToMe; as its budget, we take the GMACs of ToMe with the same $r$ in our engine.
Where we can check SAD-TM's numbers, they are on our scale. It reports ToMe at 80.58, 82.32 and 86.01\% on DeiT-B,
MAE ViT-B and ViT-H, and DiffRate at 86.40\% on MAE ViT-H. At the same budgets, the official ToMe code and DiffRate's official rates and
arithmetic give 80.60, 82.30, 86.00 and 86.36\% in our engine.

For each budget, we take our schedule with the most GMACs within it, over all GPUs and batch sizes.
This rule was fixed in advance and uses no test accuracy. Taking the most accurate schedule instead would select on the test set.
It would change the schedule in 10 of the 30 compared rows and change $\Delta$ there by 0.01 to 1.07\pp. The largest changes are on DeiT-B against SAD-TM,
where $\Delta$ would become $-0.07$, $-0.09$ and $+0.17$\pp at 11.3, 10.5 and 10.1\,G.
ATM adapts the number of merged tokens to each batch, so its budget is an average.
CATIS reports results only on ViT, DeiT-3 and DINOv3 models, which we do not evaluate, and on VideoMAE on UCF-101.
AdaMerge reports only a ViT-B/16 whose weights it does not name (unreduced top-1 85.19\%).

\begin{table*}[t]
  \centering
  \footnotesize
  \setlength{\tabcolsep}{4pt}
  \caption{Published training-free ImageNet top-1 of methods without public code, at their computational budget, against the
    schedule of \method (ours) with the most GMACs within that budget (its GMACs in brackets). $\Delta$ = \method $-$
    published; bold: the higher top-1. Zero-TPrune budgets scaled to our count: 4.01, 3.00 and 2.53\,G. RAPID budgets: GMACs of ToMe with the same $r$ in
    our engine. SAD-TM: the Pareto front of its two variants (SAD-TM-DM defers merging). Its unreduced models reach 79.87,
    81.81, 83.68, 85.98 and 86.89\% (ours: 79.81, 81.82, 83.73, 85.94 and 86.88\%). None of our schedules is below
    7.14\,G on DeiT-B.}
  \label{tab:published}
  \resizebox{\textwidth}{!}{%
  \begin{tabular}{@{}llcccc|llcccc@{}}
    \toprule
    method (published) & model & budget (G) & top-1 & \method (ours) & $\Delta$ & method (published) & model & budget (G) & top-1 & \method (ours) & $\Delta$ \\
    \midrule
    Zero-TPrune-c & DeiT-S & 3.97 & 79.8 & \textbf{79.84} (4.00) & $+0.04$ & RAPID, $r{=}11$ & DeiT-B & 11.46 & 81.25 & \textbf{81.40} (11.45) & $+0.15$ \\
    Zero-TPrune-a & DeiT-S & 2.97 & 79.4 & \textbf{79.54} (2.96) & $+0.14$ & RAPID, $r{=}14$ & DeiT-B & 9.83 & 80.52 & \textbf{80.77} (9.70) & $+0.25$ \\
    Zero-TPrune-b & DeiT-S & 2.50 & \textbf{79.1} & 78.77 (2.42) & $-0.33$ & RAPID, $r{=}17$ & DeiT-B & 8.31 & 78.77 & \textbf{78.93} (7.58) & $+0.16$ \\
    ATM & DeiT-S & 3.0 & \textbf{79.7} & 79.54 (2.96) & $-0.16$ & RAPID, $r{=}20$ & DeiT-B & 7.14 & 72.03 & -- & -- \\
    Token Transforming & DeiT-S & 3.0 & \textbf{79.7} & 79.54 (2.96) & $-0.16$ & SAD-TM-DM & DeiT-B & 12.6 & \textbf{81.75} & 81.65 (12.46) & $-0.10$ \\
    SAD-TM-DM & DeiT-S & 3.3 & \textbf{79.78} & 79.68 (3.27) & $-0.10$ & SAD-TM-DM & DeiT-B & 11.5 & 81.36 & \textbf{81.40} (11.45) & $+0.04$ \\
    SAD-TM-DM & DeiT-S & 3.1 & \textbf{79.74} & 79.53 (3.08) & $-0.21$ & SAD-TM-DM & DeiT-B & 11.3 & \textbf{81.25} & 80.36 (10.98) & $-0.89$ \\
    SAD-TM-DM & DeiT-S & 3.0 & \textbf{79.71} & 79.54 (2.96) & $-0.17$ & SAD-TM-DM & DeiT-B & 10.5 & \textbf{81.05} & 80.70 (10.13) & $-0.35$ \\
    SAD-TM-DM & DeiT-S & 2.9 & \textbf{79.65} & 79.43 (2.86) & $-0.22$ & SAD-TM & DeiT-B & 10.1 & \textbf{80.60} & 79.70 (9.90) & $-0.90$ \\
    SAD-TM-DM & DeiT-S & 2.8 & \textbf{79.53} & 79.14 (2.63) & $-0.39$ & SAD-TM-DM & MAE ViT-L & 38.0 & 85.37 & \textbf{85.48} (36.32) & $+0.11$ \\
    SAD-TM & DeiT-S & 2.6 & \textbf{79.27} & 78.77 (2.42) & $-0.50$ & SAD-TM-DM & MAE ViT-L & 37.5 & 85.34 & \textbf{85.48} (36.32) & $+0.14$ \\
    SAD-TM & DeiT-S & 2.5 & \textbf{79.15} & 78.77 (2.42) & $-0.38$ & SAD-TM-DM & MAE ViT-L & 37.0 & 85.24 & \textbf{85.48} (36.32) & $+0.24$ \\
    ATM & MAE ViT-B & 11.5 & 82.5 & \textbf{83.02} (11.33) & $+0.52$ & Token Transforming & MAE ViT-H & 92.9 & 86.1 & \textbf{86.53} (91.29) & $+0.43$ \\
    SAD-TM & MAE ViT-B & 11.5 & 82.46 & \textbf{83.02} (11.33) & $+0.56$ & SAD-TM-DM & MAE ViT-H & 102.1 & 86.58 & \textbf{86.65} (101.74) & $+0.07$ \\
    SAD-TM & MAE ViT-B & 11.0 & 82.32 & \textbf{82.62} (10.71) & $+0.30$ & SAD-TM & MAE ViT-H & 97.3 & 86.43 & \textbf{86.56} (94.70) & $+0.13$ \\
     &  &  &  &  &  & SAD-TM-DM & MAE ViT-H & 92.6 & 86.42 & \textbf{86.53} (91.29) & $+0.11$ \\
    \bottomrule
  \end{tabular}}
\end{table*}

\section{Analyses and qualitative results}
\label{sec:supp:analyses}
\begin{figure}[t]
  \centering
  \includegraphics[width=\linewidth]{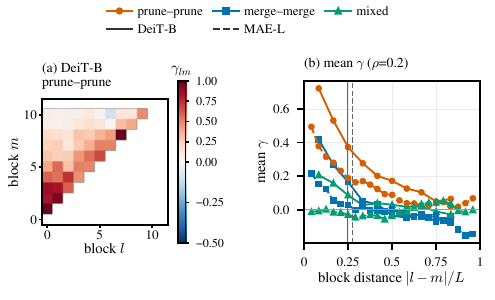}
  \caption{Measured coherence $\gamma_{lm}$ of two reductions of our menu at $\rho=0.2$. (a) Two prunes on DeiT-B.
    (b) Mean $\gamma$ against the block distance for DeiT-B and MAE ViT-L; vertical lines mark the fitted coherence length. Maps of every pair kind: \cref{fig:supp:h1}.}
  \label{fig:h1}
\end{figure}

\begin{figure*}[t]
  \centering
  \includegraphics[width=0.85\textwidth]{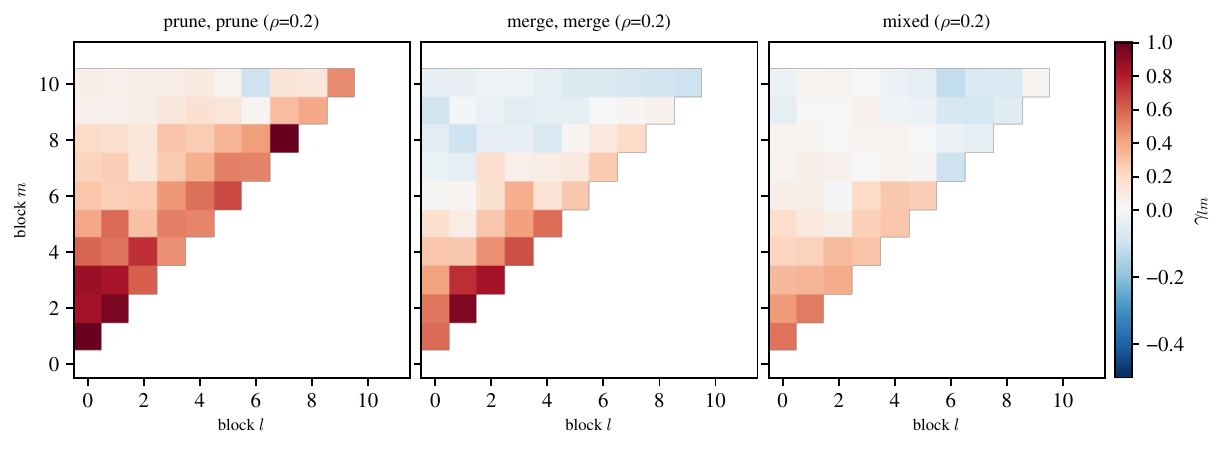}\\[2pt]
  \includegraphics[width=0.85\textwidth]{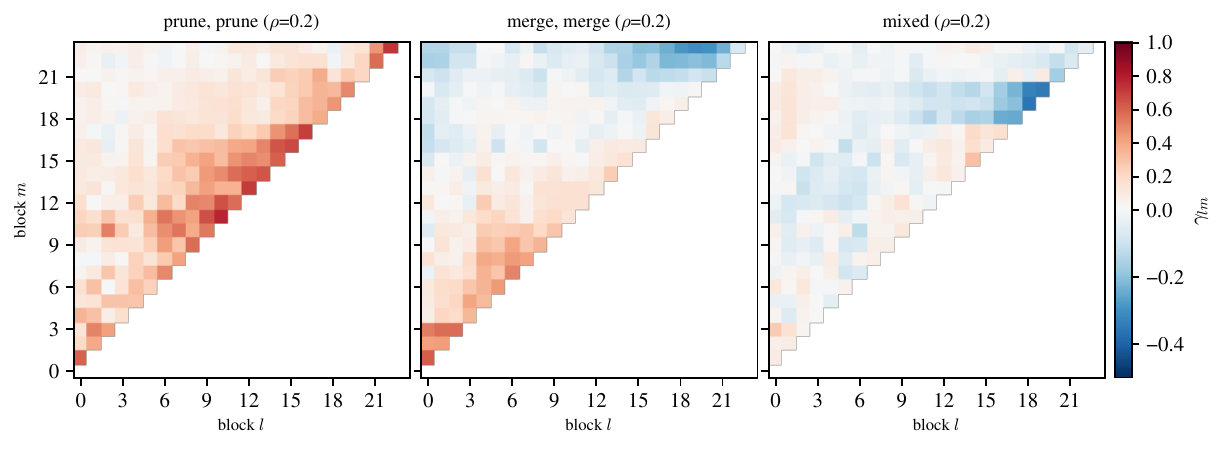}
  \caption{Measured coherence $\gamma_{lm}$ of two reductions at $\rho=0.2$. Top: DeiT-B. Bottom: MAE ViT-L.}
  \label{fig:supp:h1}
\end{figure*}

\paragraph{Interference maps.}
\label{sec:supp:h1}
\Cref{fig:h1} summarizes the coherence of two prunes on DeiT-B and how it decays with the block distance. \Cref{fig:supp:h1} shows the measured coherence $\gamma_{lm}$ for every pair of reduction steps of
DeiT-B and MAE ViT-L at $\rho=0.2$: two prunes, two merges, and a merge with a prune. On DeiT-B, a reduction after the
attention of the last block cannot change the CLS read-out. The 135 pairs whose second reduction lies there therefore have no
defined $\gamma$ and are left out (the empty top rows). The prunes use the per-block scores of an earlier calibration,
without the residual size. A structured model of $\gamma$
(by operator pair and block distance) predicts the pair KL with 60\% and 56\% lower log error than one scalar $c$.
\method uses the scalar $c$ of \cref{M-eq:coherent} only to build starting points for the search. The search then checks them on the true
KL and refines them.

\paragraph{When token reduction changes robustness.}
\label{sec:supp:ina}
On DeiT, DiffRate can exceed the unreduced model on \INA (29.2\% against 27.87\% on DeiT-B), while \method stays close to
it. Let $a$ be the unreduced accuracy, $f$ the share of its errors that a method fixes, and $b$ the share of its hits that
it breaks. The accuracy then changes by $(1-a)f-ab$, so a method gains iff $f/b>a/(1-a)$ (\cref{prop:accuracy}). This
threshold is 0.24 on \INA for DeiT-S but 4--6 on \IN, where changes almost always cost accuracy. Our changes are as well
directed as DiffRate's: $f/b$ is $0.38\pm0.02$ for \method and $0.36\pm0.03$ for DiffRate on DeiT-S, against
$0.21\pm0.01$ for the other families. However, DiffRate changes 5.3\% of the predictions and \method 2.1\%. The larger number of changes lifts \INA
and costs \IN.

\begin{figure*}[t]
  \centering
  \includegraphics[width=\textwidth]{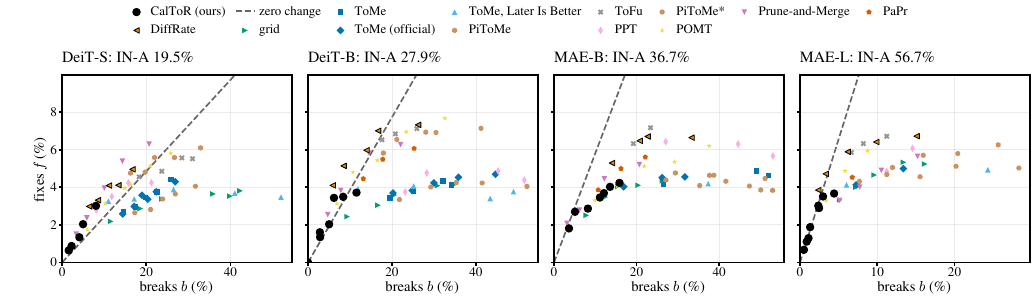}
  \caption{Fixed errors $f$ against broken hits $b$ of the unreduced model on \INA for four models, with one point per
    configuration up to $1.35\times$ (\gh, batch 256); \method (ours) in black. A configuration above the dashed line $f=b\,a/(1-a)$
    beats the unreduced model.}
  \label{fig:supp:ina}
\end{figure*}

\Cref{fig:supp:ina} shows the four models with an \INA analysis and every competitor family.
The threshold $a/(1-a)$ is 0.24 on DeiT-S, 0.39 on DeiT-B, 0.58 on MAE ViT-B and 1.31 on MAE ViT-L. On MAE ViT-B and
ViT-L the threshold is high, and every method loses \INA on average; ours loses least. \PPS makes the changes less well
directed, not more frequent. Without it, $f/b$ on DeiT-B rises from 0.40 to 0.54, at
about the same 5.1--5.5\% of changed predictions. The difference, paired by target, is $+0.14\pm0.04$ over
the 7 targets of the ablation whose schedules change a prediction, including those outside the speed range used in
\cref{tab:ablation}. At the slowest target, both variants only prune after the last attention, so $f/b$ is undefined there.

\begin{figure*}[t]
  \centering
  \includegraphics[width=\textwidth]{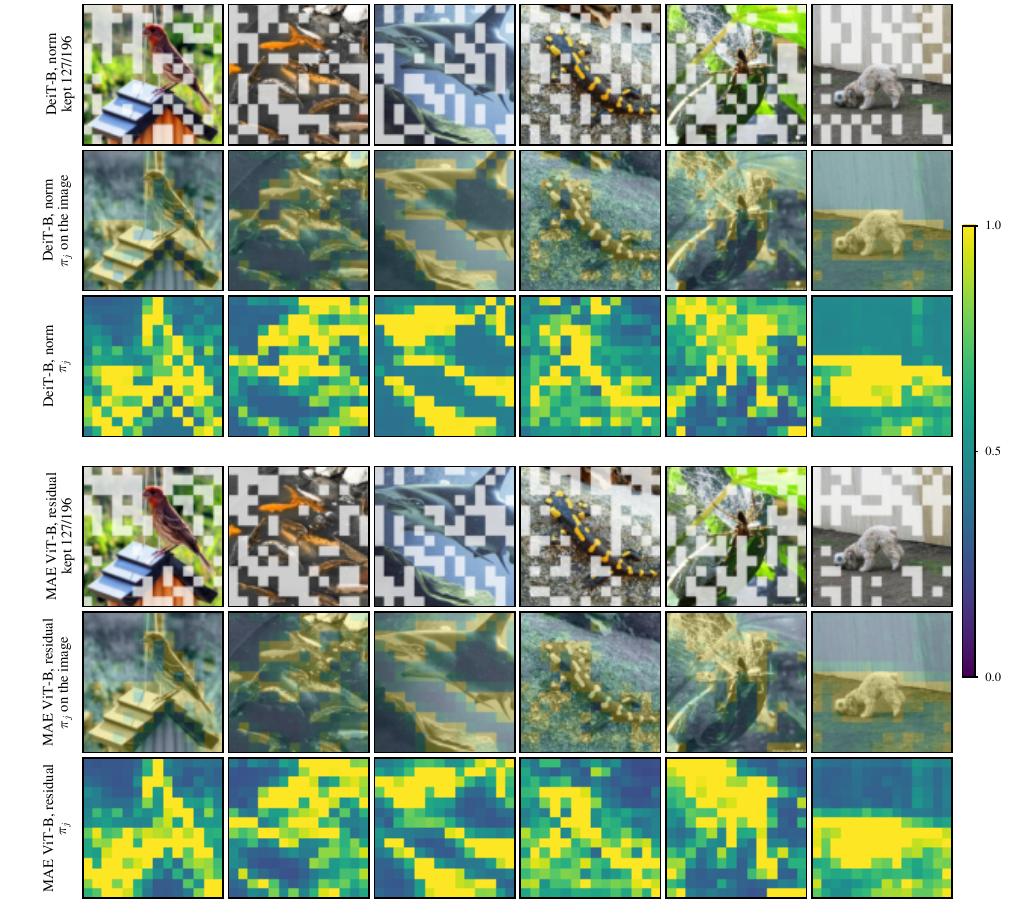}
  \caption{\PPS in block~0 on held-out training images: the kept patches (127 of 196; dropped ones faded) and the
    inclusion probabilities $\pi_j$, over the (gray) image and alone. Top: DeiT-B with the token-norm size. Bottom: MAE
    ViT-B with the prediction-residual size. $\pi_j$ is high on textured regions and edges and lower, but at least
    $k\alpha/n=0.19$, on flat background.}
  \label{fig:supp:pps}
\end{figure*}

\paragraph{Examples of \PPS.}
\label{sec:supp:pps}
\Cref{fig:supp:pps} shows the sample drawn by one \PPS step in block~0 on six held-out training images.
The step keeps 127 of 196 patches. The systematic sweep spreads the kept patches over the image.
The size measure decides where the inclusion probabilities are high. Elsewhere, the defensive term keeps them at
$k\alpha/n=0.19$ or more (\cref{sec:supp:proofs}).

\begin{figure*}[t]
  \centering
  \includegraphics[width=\textwidth]{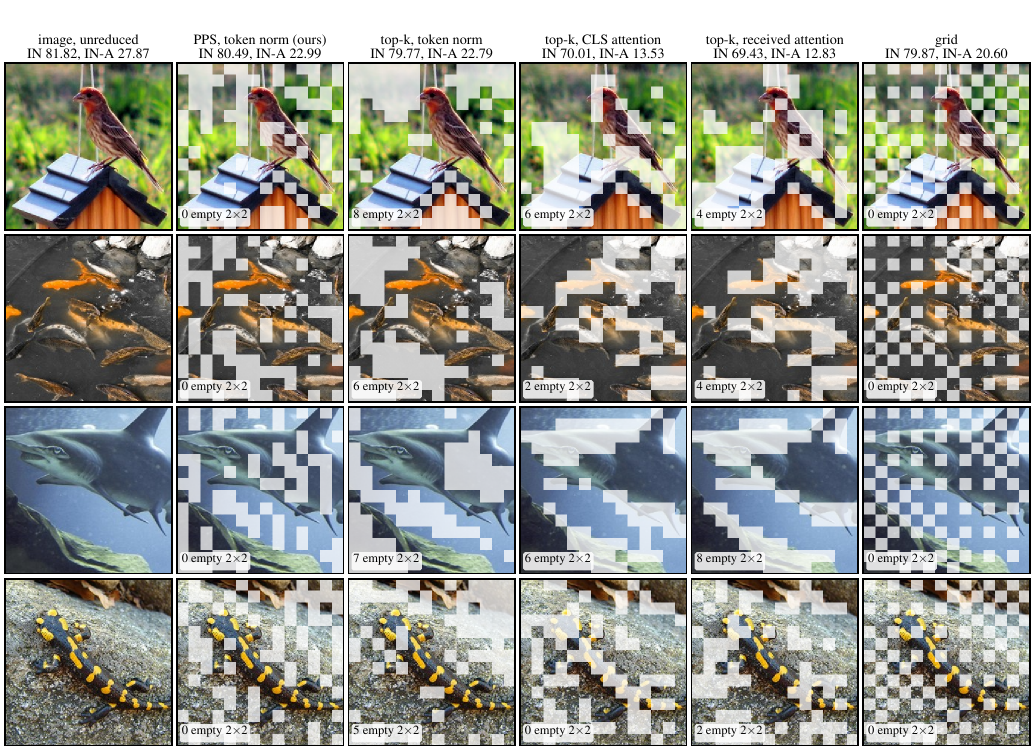}
  \caption{One prune of DeiT-B in block~0 that keeps 127 of 196 patches, by \PPS with the token-norm size (ours), by
    top-$k$ on the token norm, the CLS attention and the received attention, and by the grid, on held-out training images
    (dropped patches faded); the first column shows the image. Titles: top-1 on \IN and \INA of the unreduced model
    (first column) and of each single prune. Corner: aligned $2\times2$ squares of the patch grid without a kept patch.}
  \label{fig:supp:ppstopk}
\end{figure*}

\paragraph{\PPS against top-$k$.}
\label{sec:supp:ppstopk}
\Cref{fig:supp:ppstopk} keeps 127 of 196 patches of DeiT-B in block~0, either by \PPS, by top-$k$ on one of three scores, or
by the grid. Top-$k$ leaves holes. On the four images, 5--8 aligned $2\times2$ squares of the patch grid are empty
after top-$k$ on the token norm, and 0--8 after top-$k$ on an attention score; none are empty after \PPS or the grid. The
attention scores carry little information in block~0, and top-$k$ on them costs 11.8 and 12.4\pp of top-1 on \IN.
As a single step, \PPS costs 1.33\pp on \IN, the grid 1.94\pp and top-$k$ on the norm 2.05\pp. On \INA, they cost 4.9,
7.3 and 5.1\pp.

\PPS and ATS~\cite{ats} both sample tokens in proportion to a score; they differ mostly in the score. Keeping 98 of 196
patches after block~0, ATS reaches 70.88\% \IN top-1 on DeiT-S and 77.01\% on DeiT-B. We then turn ATS into our \PPS one change at a time
(numbers for DeiT-S and DeiT-B, respectively). Drawing exactly $k$ tokens without
replacement adds 0.27 and 0.15\pp. Using the softmax of the token norm as the size measure adds 3.07 and 0.93\pp. The Morton order
with the offset from the image gives $+0.09$ and $-0.07$\pp. The defensive mixture adds 0.46\pp on both. Our \PPS thus reaches 74.77 and 78.49\%.
After block~3, where the attention is informative, the token norm is no better than the ATS score
($-0.21$ and $-0.55$\pp on \IN).

\begin{figure*}[t]
  \centering
  \includegraphics[width=\textwidth]{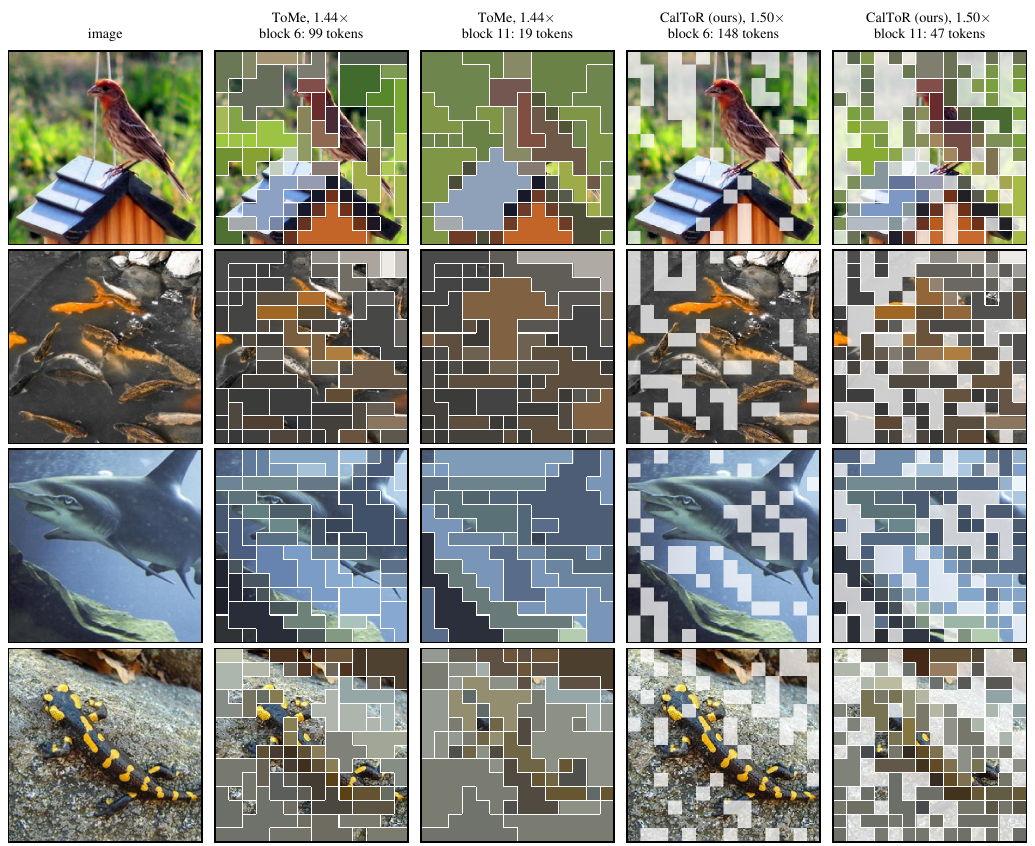}
  \caption{Tokens entering blocks 6 and 11 of DeiT-B under ToMe ($1.44\times$) and \method (ours, $1.50\times$) on
    held-out training images (\gh, batch 256). A merged token is drawn in the mean color of its patches and outlined, a
    single-patch token as it is; dropped patches are faded. The unreduced model, ToMe and \method classify all four
    images correctly.}
  \label{fig:supp:tokens}
\end{figure*}

\begin{figure*}[t]
  \centering
  \includegraphics[width=\textwidth]{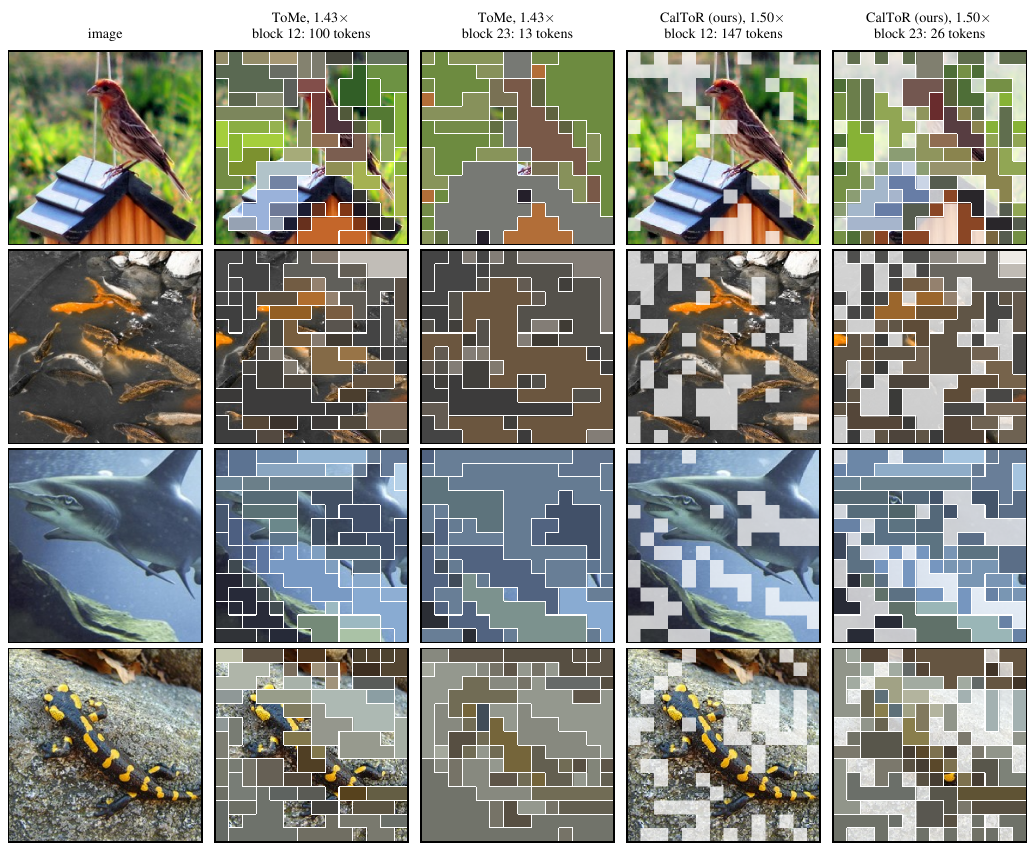}
  \caption{Tokens entering blocks 12 and 23 of MAE ViT-L under ToMe ($1.43\times$, 8 merges per block) and \method
    (ours, $1.50\times$) on the images of \cref{fig:supp:tokens}, drawn as there. The unreduced model, ToMe and \method
    classify all four images correctly.}
  \label{fig:supp:tokensL}
\end{figure*}

\begin{figure*}[t]
  \centering
  \includegraphics[width=\textwidth]{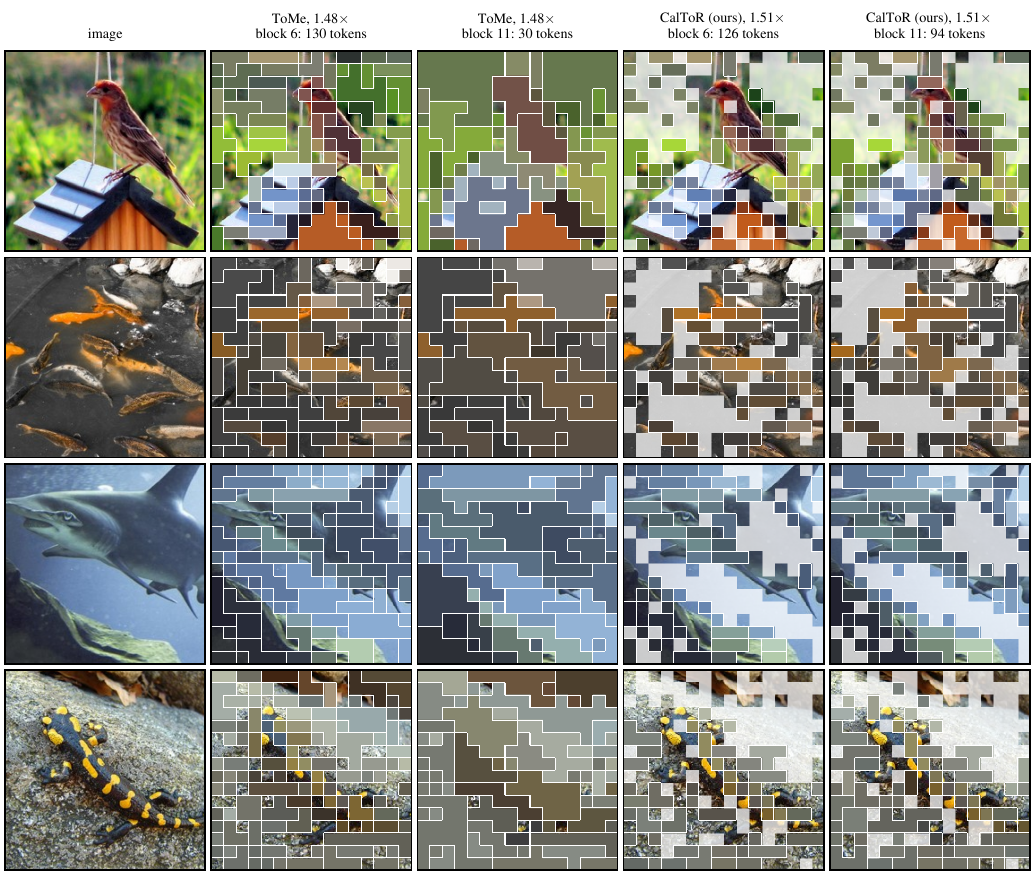}
  \caption{Tokens entering blocks 6 and 11 of DINOv2 ViT-B/14 under ToMe ($1.48\times$, 20--21 merges per block) and
    \method (ours, $1.51\times$) on the images of \cref{fig:supp:tokens}, drawn as there. The unreduced model and
    \method classify all four images correctly, ToMe three of them.}
  \label{fig:supp:tokensD}
\end{figure*}

\paragraph{Token maps.}
\label{sec:supp:tokens}
\Cref{fig:supp:tokens} shows which image regions the tokens of ToMe and \method cover at about the same measured speed
on DeiT-B (\gh, batch 256). We compare ToMe with 16--17 merges per block at $1.44\times$ with its paired schedule
of \method from the main tables, at $1.50\times$. Before block~6, ToMe has already merged half of the tokens, including parts of the objects. \method
keeps the image nearly intact up to the middle of the network. Its \PPS step in block~0 drops 48 patches spread over
the image, mostly background. It reduces the tokens in the later blocks, where merging and pruning cost less
(\cref{fig:supp:damage}). Its last block still processes 47 tokens, against 19 for ToMe.
MAE ViT-L shows the same pattern (\cref{fig:supp:tokensL}). ToMe at $1.43\times$ merges 8 tokens per block. It keeps
100 tokens entering block~12 and 13 entering block~23. The paired schedule of \method at $1.50\times$ samples in blocks 3--8 and
merges from block~13 on, so it keeps 147 and 26 tokens.
On DINOv2 ViT-B/14 (\cref{fig:supp:tokensD}), ToMe at $1.48\times$ merges 20--21 tokens per block. It keeps 130 tokens
entering block~6 and 30 entering block~11. The paired schedule of \method at $1.51\times$ has a \PPS step in block~1, a
prune by CLS attention in block~2 and merges in blocks 5 and 7, so it keeps 126 and 94 tokens. Unlike on DeiT-B and
MAE ViT-L, it does not keep more tokens than ToMe in the middle of the network; in the last block it keeps 94 against 30.

\section{Extended related work}
\label{sec:supp:related}
\paragraph{Further training-free methods.} Zero-TPrune ranks tokens on the attention graph~\cite{zerotprune}, and PaPr prunes patches with a lightweight proposal network~\cite{papr}.

\paragraph{Further calibrated methods.} In multimodal LLMs, pruning layers for visual tokens are also calibrated on a few samples~\cite{fitprune,vtw}.

\paragraph{Further latency-aware and compression work.} Outside token reduction, hardware-aware architecture search~\cite{fbnet,proxylessnas,ofa,hat} and latency-aware channel pruning~\cite{netadapt,halp} choose a network from measured latency or a latency predictor. Model compression uses measured sensitivities to choose bit widths~\cite{hawqv2,hawqv3} or channel counts~\cite{halp}. LANA picks distilled per-layer operations under a measured latency budget~\cite{lana}.

\section{Implementation details}
\label{sec:supp:impl}
\paragraph{Models and data.} DeiT-S and DeiT-B are the timm checkpoints \texttt{deit\_\{small,base\}\_patch16\_224.fb\_in1k}.
For MAE ViT-B, ViT-L and ViT-H we use the official fine-tuned MAE checkpoints, with average pooling.
Our calibration pool is a fixed subset of the ImageNet training set with 10 images per class. The calibration uses 512
of these images, and the final pick uses 2048 others. We evaluate on the 50\,000 validation images, on \INV (10\,000),
\INR (30\,000) and \INA (7\,500), with the logits restricted to their 200 classes, and on \INS (50\,889), all as
$224\times224$ center crops.
For video we use VideoMAE-B fine-tuned on K400 and on SSv2, and VideoMAE-L fine-tuned on K400 (the MCG-NJU
checkpoints). The calibration takes the center view of 1024 training clips. Their reference distribution is a mixture of the model output and the one-hot label, with probability 0.75 for the label and 0.25 for the model output.

\paragraph{Latency table.} We time the attention and the MLP of one block (all blocks of a ViT have the same shape) on a
grid of token counts from 8 to the full count (17 points for 197 tokens). The grid has points on both sides of 64, 96,
128 and 192 tokens. Each operator is timed at every token count and at four amounts: merges remove 1, 10, 25 or 50\% of
the tokens, and prunes keep 20, 50, 80 or 98\%. The table interpolates linearly between these points. The cost of a
step is therefore piecewise linear in $r$, and $\kappa^0_o$, $\kappa^1_o$ in \cref{M-eq:latency_model} describe it to first
order. When an operator needs attention statistics, we price them through the variant of the block's attention that
computes them. After the first merge, every attention also pays for the size bias. One global scale corrects the sum:
the median ratio of measured to predicted time on 20--35 check schedules timed end to end. The errors we report are on
the same schedules. The table times \PPS after a merge, on reordered tokens. When \PPS is the first reduction of a
schedule, it runs on tokens in raster order, and the step is 1.1--2.3 times faster than the table predicts (RTX~3080).
The table therefore overprices schedules that start with \PPS.

\paragraph{Damage curves.} A merge curve is measured at $\rho\in\{0.05,0.1,0.15,0.2,0.3,0.4,0.5\}$, and a prune curve
also at 0.65 and 0.8. Each point removes the share $\rho$ of the tokens that enter the step. All targets of a model share
the same merge curves; the prune curves are measured again for the final menu.

\paragraph{Operators.} The merge cost of a pair is $(1-\cos)\,(u_i+u_j)^{0.5}$, where $u$ is the importance: CLS
attention on DeiT and CLIP, and received attention on MAE and video. The cosine is computed on the token features,
except for MAE ViT-B and video, where it is computed on the keys. A merged token is the size-weighted mean of its sources, rescaled to their
size-weighted mean norm. This is a variant of ToFu's norm-preserving merge, which keeps the largest norm~\cite{tofu}. On
images, the received attention of a merged token is the sum of those of its members, and the later steps of the same
block use it. Attention
adds the $\log$ of the token size to the logits. On video, background sources (saliency below the mean) enter a merge
with weight 0, as in vid-TLDR~\cite{vidtldr}. \PPS uses $\alpha=0.3$. Its residual size is the log norm of the
difference between a token and the mean of its grid neighbors that are still present. Received attention is the
attention a token receives, averaged over the queries and heads. CLS attention is the attention of the CLS query,
averaged over the heads. As in ToMe, the matching splits the tokens alternately into two sets, and the CLS token is
never merged.

\paragraph{Search.} The base targets are 4--10 throughputs per model. On DeiT and MAE at batch 256, they are fixed
values at 1.03--2.04 times the predicted speed of the unreduced model; at batch 32, they are 1.05, 1.1, 1.2, 1.3, 1.4
and 1.5 times this speed. On CLIP, they are 1.1--1.75 (ViT-B/16) and 1.1--1.9 times (ViT-L/14). As the prune score of a
block we take the candidate with the lowest mean damage, measured at the time savings that the candidate with the
largest savings reaches at $\rho=0.1$--$0.5$. Allocation steps remove $2.5\%$ of the tokens entering the step, times 1,
2, 4 or 8 (with the MAC cost, once). A larger step that overshoots the target counts only the time it saves down to the
target. The coherence $c$ is the value on a grid of step 0.01 that minimizes the squared error of $\log\hat{\mathcal{K}}$
against the log of the true KL of the screened starts. These starts are pooled with those of the targets calibrated
earlier in the same run, so a schedule can depend on the order of the targets. The refinement runs 60 steps on images
and 100 on video, and each step ranks 8 moves proposed at random. A move shifts 2, 4, 6, 16 or 32 tokens (on video also
64, 128 and 256) between two cells at equal predicted time. With probability 0.3, the move instead empties the source
cell, which also removes its fixed cost. The starts are the additive allocation with all operators ($c=0$), merging only, pruning only, the allocations with $c=0.2$ and $0.5$, and single prunes in blocks 0--3. The search then adds the start with the fitted $c$, keeps the four starts with the lowest KL on the 512 calibration images, refines them (\cref{M-sec:refinement}) and picks the final schedule on 2048 other images. Video adds
a start with a single input-level prune. On video, the refinement averages the KL over batches of 4 and 8 clips, because in bf16
the merge decisions change with the batch size.

\paragraph{Throughput.} Each job measures configurations of one model, GPU and batch size, and always includes the
unreduced model and our final schedules. It makes two passes over the list, the second in reverse order (one pass on
video and at batch 1). Each measurement takes five repeats of 20 iterations at batch 256, of 50 at batch 32 and of 200
at batch 1, and we keep the best repeat.
We measure throughput with random weights and Gaussian inputs. Pretrained weights and real images change it by at most
0.6\% and the speed-ups by at most 0.4\% (RTX~3080; DeiT-S, DeiT-B and ToMe at batch 256 and 32). All runs use bf16 and
CUDA graphs, with torch~2.13, CUDA~13.2, Triton~3.7 and timm~1.0.28 on every cluster GPU.

\paragraph{Evaluation.} The offset of \PPS is a hash of the sizes, $u=\operatorname{frac}(12.9898\sum_j|z_j|+0.5)\cdot0.98+0.01$.
The forward pass is therefore deterministic for a given GPU, batch size and kernel. A different batch size changes the
bf16 rounding and can change the tokens that later steps keep. Top-1 is evaluated once per configuration, at batch 250
in every setting, and not necessarily on the GPU of the throughput job (our \ahun schedules were evaluated on RTX~4090
nodes). Compared with batch 256, one schedule of MAE ViT-B keeps the same final tokens for 97.9\% of 512 images at
batch 250, with no changed prediction, and for 3.5\% at batch 32, with 3 changed predictions. One schedule of DeiT-B
keeps them for 99.8\% (batch 250) and 100\% (batch 32) of the images, with no changed prediction. Re-evaluated on an RTX~3080 (torch~2.11), 16
configurations of DeiT-S and DeiT-B change their top-1 on \IN by at most 0.08\pp, even though the change of GPU and
software flips the hit or miss of 0.1--5.8\% of the images.

\section{Theory}
\label{sec:supp:theory}
This section states the mathematical facts behind the design of \method and proves them. When a fact makes a
measurable prediction, we check it against our measurements. \Cref{M-prop:threshold} (break-even size) is proved in
\cref{M-sec:latency}. We write $\mathcal{K}(s)$ for the expected KL of \cref{M-eq:objective}, $D_{o,l}(\rho)$ for the
damage curves and, as in \cref{M-prop:threshold}, $S_l$ for the time saved per token removed after the attention of block
$l$.

\paragraph{Damage of a schedule.}
\label{sec:supp:theory:damage}
\begin{proposition}[Second-order damage]
\label{prop:coherent}
Let $z(x)$ be the logits of the unreduced model and $p=\mathrm{softmax}(z)$. Let $\delta_i(x)$ be the change of the
logits caused by step $i$ of a schedule $s$ when this step is applied alone. Let
$D_i=\tfrac12\mathbb{E}_x[\delta_i^\top F\delta_i]$ be the damage of step $i$ alone to second order, with
$F=\mathrm{diag}(p)-pp^\top$, and assume $D_i>0$. Assume that the effects add, so that
the reduced logits are $z+\sum_i\delta_i$, and keep the KL to second order. Then
\begin{equation}
  \begin{aligned}
    \mathcal{K}(s)&=\sum_i D_i+\sum_{i\neq j}\gamma_{ij}\sqrt{D_iD_j},\\
    \gamma_{ij}&=\frac{\mathbb{E}_x[\delta_i^\top F\delta_j]}{2\sqrt{D_iD_j}}\in[-1,1].
  \end{aligned}
  \label{eq:secondorder}
\end{equation}
Hence (i) the measured coherence $\gamma_{lm}$ of \cref{sec:supp:h1} is the Fisher cosine $\gamma_{ij}$ of two steps;
(ii) $\mathcal{K}(s)\le(\sum_i\sqrt{D_i})^2$, which is the coherent model with $c=1$; and (iii) if $\gamma_{ij}=c$ for
all $i\neq j$, \cref{M-eq:coherent} holds exactly.
\end{proposition}
\begin{proof}
With $q=\mathrm{softmax}(z+\delta)$ we have $\log p_y-\log q_y=-\delta_y+\mathrm{LSE}(z+\delta)-\mathrm{LSE}(z)$, so
$\KL(p\,\|\,q)=\mathrm{LSE}(z+\delta)-\mathrm{LSE}(z)-p^\top\delta$. The gradient of the log-sum-exp is $p$ and its
Hessian is $F$. Hence $\KL(p\,\|\,q)=\tfrac12\delta^\top F\delta+O(\|\delta\|^3)$. Substituting $\delta=\sum_i\delta_i$
and taking the expectation over $x$ gives $\mathcal{K}(s)=\sum_i D_i+\tfrac12\sum_{i\neq j}\mathbb{E}_x[\delta_i^\top F\delta_j]$,
which is \cref{eq:secondorder}. Since $F$ is positive semi-definite, $\langle u,v\rangle=\mathbb{E}_x[u^\top Fv]$ is a
semi-inner product on random logit changes, and the Cauchy--Schwarz inequality gives $|\gamma_{ij}|\le1$. (i) For two
steps, $\mathcal{K}_{lm}=D_l+D_m+2\gamma_{lm}\sqrt{D_lD_m}$, which is the definition used in \cref{sec:supp:h1}.
(ii) Take the seminorm $\|u\|=\langle u,u\rangle^{1/2}$. Then $\|\delta_i\|=\sqrt{2D_i}$, and the triangle inequality
gives $\mathcal{K}(s)=\tfrac12\|\sum_i\delta_i\|^2\le\tfrac12(\sum_i\sqrt{2D_i})^2$. (iii) $\sum_{i\neq j}c\sqrt{D_iD_j}=
c[(\sum_i\sqrt{D_i})^2-\sum_iD_i]$.
\end{proof}
\noindent\emph{Check.} Mixed pairs (a merge and a prune) are much less coherent than two prunes. Their median $\gamma$ is
$+0.04$ and $-0.01$ over all pairs, and at most $+0.41$ (DeiT-B) and $+0.15$ (MAE ViT-L) in neighboring blocks, against
$+1.33$ and $+0.81$ for two prunes (for a merge and a prune in the same block, it reaches up to $+0.82$ and $+0.29$). A
merge and a prune thus perturb the logits in directions that are closer to Fisher-orthogonal, and mixing operators keeps
the damage closer to additive. Two prunes in neighboring blocks have a median $\gamma$ of 1.33 on DeiT-B at $\rho=0.35$
(single pairs reach up to 2.47). This is impossible for additive effects in the second-order regime: the second prune
acts on the tokens that the first one kept and removes the next ones in line. This is why we use the coherent model only
to build starts, which the search then checks on the true KL and refines. Over the 186 targets of the main settings, the
coherence fitted by the calibration has a median of 0.41 (interquartile range 0.35--0.54). Starts built with a fixed or
fitted coherence win 121 of the 185 targets with a nonzero KL (at the remaining target, five starts tie at KL 0). On
video, the fitted coherence is 1 at all 38 targets: there the coherent model underestimates the measured damage even at
$c=1$.

\begin{proposition}[Random pruning]
\label{prop:random}
Let one block keep a simple random sample of $k=(1-\rho)n$ of its $n$ tokens, with no token that is always kept, and let
every later operation run on the kept tokens. To first order in the sampling error, the expected KL is
$D(\rho)=A\,\rho/(1-\rho)$, where $A\ge0$ does not depend on $\rho$.
\end{proposition}
\begin{proof}
Consider one query with attention weights $w_j$ and values $v_j$. Its output on the sample $S$ satisfies
$\hat o-o=\sum_{j\in S}w_j(v_j-o)/\sum_{j\in S}w_j$. The denominator has mean $k/n$. Linearizing it at its mean gives
$\hat o-o\approx\frac1k\sum_{j\in S}y_j$ with $y_j=n\,w_j(v_j-o)$ and $\sum_jy_j=0$. A pooled readout is a sample mean
of the same form. Stacking these $y_j$ over the heads, queries and later blocks gives one vector $Y_j$ per token. To
first order, the change of the logits is then $\delta=J\,\bar Y_S$, where $\bar Y_S$ is the sample mean of $Y_j$. For a
simple random sample, $\mathrm{Cov}(\bar Y_S)=\frac{1-k/n}{k}\,\Sigma_Y$ with $\Sigma_Y=\frac1{n-1}\sum_jY_jY_j^\top$,
and $\frac{1-k/n}{k}=\frac{\rho}{(1-\rho)n}$. By \cref{prop:coherent},
$D(\rho)=\tfrac12\mathbb{E}_x\,\mathrm{tr}(J^\top FJ\,\Sigma_Y)\,\rho/((1-\rho)n)$.
\end{proof}
\noindent If a token is always kept, like the CLS token in all our image models (MAE reads out the mean of the patches
but keeps the CLS token in the sequence), the sampling error no longer has mean zero. A query that gives the weight
$w_0$ to this token is shifted on average by $\rho\,w_0(v_0-o)/(1-\rho(1-w_0))$, and $D$ grows faster than
$\rho/(1-\rho)$.
\noindent\emph{Check.} We fit $D(\rho)=A\,(\rho/(1-\rho))^{\beta}$ to those damage curves of the final calibrations on
the \gh at batch 256 that have at least three points above $10^{-4}$: 90 prune and 82 merge curves of the five models
(\cref{fig:supp:damage}). The median exponent is $\beta=1.31$ (interquartile range 1.24--1.45) for the prunes of the
calibrated menu and $1.52$ (1.46--1.66) for merges. A random prune (a fixed pseudo-random order of the positions),
measured in the same way in every block, has $\beta=1.38$ (1.29--1.41) on DeiT and $1.30$ (1.27--1.36) on MAE. So over
the measured range of $\rho$, up to 0.8, random pruning is not in the first-order regime of \cref{prop:random}, with
either a CLS readout or a mean readout. What the calibrated menu changes is the level: at $\rho=0.3$, its prunes cost a
median of 0.28--0.41 times the KL of the random prune, per model. Of the curves, 96\% (prunes) and 99\% (merges) are
convex up to a tolerance of 2\% of their range; convexity is a condition of \cref{prop:greedy}.

\begin{proposition}[Merging]
\label{prop:merge}
Consider the blocks after a reduction step, with no further steps, and let a token of size $n_j$ count as $n_j$
copies: attention adds $\log n_j$ to its logit, and a pooled readout weights it by $n_j$. (i) Replacing $k$ identical
tokens by one token of size $k$ does not change the output. (ii) Replacing a group
$G$ by one token of size $n_G=\sum_{i\in G}n_i$ at the size-weighted mean $\bar x$ changes the logits by
$\Delta z=-\frac12\sum_{i\in G}n_i(x_i-\bar x)^\top\nabla^2\psi(\bar x)(x_i-\bar x)+O(\sigma_G^3)$, where $\sigma_G$ is
the spread of the group and $\psi(x)$ is the change of the logits per unit of size added at $x$. (iii) If the merged
token gets size 1 instead of $n_G$, the merge also removes $n_G-1$ units of size. To first order,
$\Delta z=-(n_G-1)\psi(\bar x)$, which is a first-order change, like a prune.
\end{proposition}
\begin{proof}
(i) $k$ identical keys add $k\,e^{s}=e^{s+\log k}$ to every softmax and carry identical values, so no attention output
changes. The $k$ tokens stay identical in all later blocks, and the readout counts them with weight $k$. Hence the
output is a function $z=\Phi(\mu)$ of the token measure $\mu=\sum_jn_j\delta_{x_j}$, smooth in the sizes, and $\psi$ is
its derivative in the direction $\delta_x$. (ii) The merge changes $\mu$ by
$\nu=n_G\delta_{\bar x}-\sum_{i\in G}n_i\delta_{x_i}$. This $\nu$ has zero mass and, by the choice of $\bar x$, zero
first moment. To first order in $\nu$,
$\Delta z=n_G\psi(\bar x)-\sum_in_i\psi(x_i)$, and a Taylor expansion of $\psi$ around $\bar x$ cancels the constant and
the linear terms. The terms of second order in $\nu$ are $O(\sigma_G^4)$. (iii) Now $\nu$ has mass $1-n_G$, so the
constant term remains.
\end{proof}
\noindent Pruning a token is a first-order change, $-n_j\psi(x_j)$. Without an always-kept token, however, the errors
of a random sample have mean zero (\cref{prop:random}). A merge is a second-order change, and no choice of the sample
can average it out. Its size is set by the curvature of the rest of the network, which shrinks with depth. Rescaling
the merged token to the members' mean norm moves it by $O(\sigma_G^2)$, so the error stays second order.
\emph{Check.} At $\rho=0.2$, a merge in block~0 costs 2.1 (DeiT-B) and 3.5 (MAE ViT-L) times the KL of the calibrated
prune, but 3.2--6.0 times less in blocks 6--9 of DeiT-B and 4--21 times less in blocks 12--20 of MAE ViT-L
(\cref{fig:supp:damage}). Accordingly, the calibrated schedules merge in the middle and late blocks and sample in the
early ones (\cref{fig:supp:schedules}).

\paragraph{Allocation.}
\label{sec:supp:theory:alloc}
For the allocation, each cell $(o,l)$ (operator $o$ in block $l$) removes $r$ tokens on a grid of $\Delta$ tokens, at
most $R_{o,l}$. It saves $w_{o,l}\,r$ of time and, when it is used, costs a fixed $\kappa^0_{o,l}=\kappa^0_o(n_l)$. All
these quantities are evaluated at the token counts $n_l$ of the unreduced model. Here $w_{o,l}=S_l-\kappa^1_o$ is the
net saving per token of \cref{M-prop:threshold}; it is constant when block times are linear in the token count. With
additive damage ($c=0$ in \cref{M-eq:coherent}), the allocation solves
\begin{equation}
  \begin{aligned}
    &\min_{r}\ \sum_{o,l}D_{o,l}(r_{o,l})\quad\text{s.t.}\quad 0\le r_{o,l}\le R_{o,l},\\
    &\sum_{o,l}\big[w_{o,l}\,r_{o,l}-\kappa^0_{o,l}\,\mathbf{1}(r_{o,l}>0)\big]\ge W,
  \end{aligned}
  \label{eq:alloc}
\end{equation}
where $W$ is the time that must be saved to reach the target and $D_{o,l}(r)$ is the damage curve in tokens.
\begin{proposition}[Greedy allocation is Lagrangian]
\label{prop:greedy}
Let $\kappa^0=0$ and $w_{o,l}>0$, and let every $D_{o,l}$ be non-decreasing and convex on the grid. At each step, the
greedy allocation adds the grid step with the smallest damage increase per unit of saved time. Let $s_t$ be its
schedule after $t$ steps and $\eta_t$ the ratio of step $t$. Then $s_t$ minimizes $\sum D-\eta_t\sum w\,r$ over all
schedules on the grid, so no schedule saves at least as much time with less damage. If the greedy stops at the first
$t$ that saves $W$, its damage exceeds the optimum of \cref{eq:alloc} by at most the damage increase of its last step.
\end{proposition}
\begin{proof}
In each cell, convexity makes the damage increments non-decreasing, while the saving per step is constant. The ratios
of a cell's successive steps are therefore non-decreasing. The greedy takes the smallest next ratio over the cells, so
the ratios it takes are non-decreasing in $t$. At step $t$, every step it has taken has a ratio $\le\eta_t$, and every
step it has not taken has a ratio $\ge\eta_t$. In each cell, the prefix taken therefore minimizes
$D_{o,l}(r)-\eta_tw_{o,l}r$: the increments of this function, $w_{o,l}\Delta(\text{ratio}-\eta_t)$, are $\le0$ before
the end of the prefix and $\ge0$ after it. The objective is separable over the cells, so $s_t$ minimizes the Lagrangian
$L(s)=D(s)-\eta_tV(s)$, where $V$ is the saved time. If $V(s)\ge V(s_t)$, then
$D(s)\ge D(s_t)+\eta_t(V(s)-V(s_t))\ge D(s_t)$; this is Everett's theorem. Finally, let $s^\star$ be optimal for the
budget $W$, and let the greedy stop at $t$. Then $V(s^\star)\ge W>V(s_{t-1})$, so $D(s^\star)\ge D(s_{t-1})=D(s_t)-\Delta D_t$.
\end{proof}
\noindent At an interior optimum of the continuous relaxation, all active cells have the same marginal damage per saved
millisecond, $D'_{o,l}/w_{o,l}=\eta$. Thus $\eta$ is the price of time, as in the equal-slope condition of
bit allocation~\cite{everett1963,shoham1988}. The greedy of \method departs from this idealization in several ways: the
token count of a cell depends on the steps before it, the latency table is nonlinear and has fixed costs, the steps are
1, 2, 4 or 8 times 2.5\% of the tokens, and some starts use $c>0$. \Cref{prop:greedy} motivates the greedy start. It
gives no guarantee for the final schedules, which come from the search on the true KL.

\begin{proposition}[Fixed costs make the allocation NP-hard]
\label{prop:nphard}
With fixed costs $\kappa^0>0$, problem \cref{eq:alloc} is NP-hard, even if every damage curve is linear.
\end{proposition}
\begin{proof}
We reduce the minimum-cost covering knapsack problem to \cref{eq:alloc}. Given positive integers $C_i$, $W_i$
($i=1,\dots,N$) and $W$, this problem asks for a set $A$ that minimizes $\sum_{i\in A}C_i$ subject to
$\sum_{i\in A}W_i\ge W$. It is NP-hard, because the complement of $A$ solves a 0/1 knapsack problem. Take $N$ cells with
$D_i(r)=r$, $w_i=1$, $\kappa^0_i=C_i$ and at most $R_i=W_i+C_i$ removable tokens. For a set $A$ of opened cells, the
saving is $\sum_{i\in A}r_i-\sum_{i\in A}C_i$. The set $A$ is feasible iff $\sum_{i\in A}W_i\ge W$, and then the least
damage is exactly $W+\sum_{i\in A}C_i$: fill the open cells until the saving is $W$. This is possible because
$1\le r_i\le R_i$ and the fill $\sum_{i\in A}r_i=W+\sum_{i\in A}C_i$ is reachable. So the optimum of \cref{eq:alloc}
is $W$ plus the optimum of the knapsack.
\end{proof}
\noindent The knapsack is only weakly NP-hard: dynamic programming over the blocks with a discretized time budget solves
\cref{eq:alloc} in pseudo-polynomial time. A prototype on DeiT-B and MAE ViT-L (\gh, batch 256) uses dynamic
programming over the blocks to minimize the predicted damage plus a multiple of the predicted time; the multiple is set
by bisection to meet the budget. The state of the dynamic program holds the token count, a merge flag and two decaying
sums of $\sqrt{D}$, and the damage model has an interference that decays with the block distance (\cref{sec:supp:h1}).
This Lagrangian solution lies on the lower convex hull of the damage--time trade-off. A local search on the model alone
and a pass that gives tokens back then use the time it leaves. The damage model is fitted to the true KL of 41
schedules per target: 30 random ones, plus the ten starts and the final schedule of \method calibrated again on an
RTX~3080. The prototype also takes the damage curves and prune scores of this calibration. At equal measured speed, the
prototype is $0.06\pm0.02$\pp less accurate than \method on DeiT-B and $0.10\pm0.08$\pp more accurate on MAE ViT-L.
Its schedules for the additive, scalar and decaying models, used as three starts of our refinement, are
$0.04\pm0.01$\pp less and $0.02\pm0.10$\pp more accurate. Without fixed costs, the greedy of \cref{prop:greedy} is
optimal for the time it saves and within one step of the optimum. With fixed costs, a single small step in an unused
cell may not pay its fixed cost (\cref{M-prop:threshold}). For this reason our steps remove 2.5\% of the tokens times 1,
2, 4 or 8, and the search adds several starts and a refinement beyond the greedy.

\paragraph{Selection on held-out images.}
\label{sec:supp:theory:select}
\begin{proposition}[Winner's curse and the held-out pick]
\label{prop:holdout}
Let $\bar{\mathcal{K}}_{n_c}(s)=\frac1{n_c}\sum_{i=1}^{n_c}\KL_i(s)$ be the mean KL over $n_c$ i.i.d.\ images, with
$\KL_i(s)\in[0,K_{\max}]$. (a) Take fixed candidates $s_1,\dots,s_M$, and let $\hat\jmath$ be the one with the smallest
$\bar{\mathcal{K}}_{n_c}$. Then
$\mathbb{E}[\bar{\mathcal{K}}_{n_c}(s_{\hat\jmath})]\le\min_j \mathcal{K}(s_j)\le\mathbb{E}[\mathcal{K}(s_{\hat\jmath})]$: the KL on the images used for the
choice underestimates the true KL of the chosen schedule. (b) Suppose that the candidates do not depend on a fresh set
of $n_h$ images, and that $\hat\jmath'$ minimizes the mean KL on this set. Then
$\mathbb{E}[\mathcal{K}(s_{\hat\jmath'})]-\min_j\mathcal{K}(s_j)\le K_{\max}\sqrt{2\log(2M)/n_h}$.
\end{proposition}
\begin{proof}
(a) $\mathbb{E}[\min_j\bar{\mathcal{K}}_{n_c}(s_j)]\le\min_j\mathbb{E}[\bar{\mathcal{K}}_{n_c}(s_j)]=\min_j\mathcal{K}(s_j)$, and $\mathcal{K}(s_{\hat\jmath})\ge\min_j\mathcal{K}(s_j)$.
(b) Let $\epsilon=\max_j|\bar{\mathcal{K}}'(s_j)-\mathcal{K}(s_j)|$ on the fresh images, and let $j^\star$ be the best
candidate. Then
$\mathcal{K}(s_{\hat\jmath'})\le\bar{\mathcal{K}}'(s_{\hat\jmath'})+\epsilon\le\bar{\mathcal{K}}'(s_{j^\star})+\epsilon\le \mathcal{K}(s_{j^\star})+2\epsilon$. By
Hoeffding's lemma, each $\pm(\bar{\mathcal{K}}'(s_j)-\mathcal{K}(s_j))$ is sub-Gaussian with variance proxy
$K_{\max}^2/(4n_h)$, and the maximum of $2M$ such variables has mean at most $\frac{K_{\max}}2\sqrt{2\log(2M)/n_h}$.
\end{proof}
\noindent\emph{Check.} The refinement accepts each move by its KL on the 512 calibration images, so the selection bias
of part (a) acts at every accepted move. Choosing the moves on the same images adds to this bias. Over the 185 final
schedules with both values, the KL on the 2048 held-out images is a median of 1.23 times the KL on the calibration
images (interquartile range 1.11--1.31), and it is larger for 91\% of them (\cref{fig:supp:kl}). This held-out KL is
itself the minimum over the candidates on those images, so the ratio is, if anything, an underestimate. The final pick
on 2048 other images is the remedy of part (b): it uses four times as many images and few candidates. It changes the
choice for 68 of the 185 targets with a nonzero KL. For these targets, it lowers the KL on the images of the pick by a
median of 6\%, which at the slopes of \cref{fig:supp:kl} moves top-1 by a few hundredths of a point. Compared with a
pick on 512 other images, it changes \IN by $+0.01$ and $-0.04$\pp (\cref{tab:ablation}).

\paragraph{KL and the prediction.}
\label{sec:supp:theory:kl}
\begin{proposition}[Small KL keeps the prediction]
\label{prop:pinsker}
Let $m(x)=p_{(1)}-p_{(2)}$ be the gap between the two largest probabilities of the unreduced model on image $x$. If
$\KL(p(\cdot\mid x)\,\|\,p_s(\cdot\mid x))<m(x)^2/2$, the reduced model predicts the same class. Hence the share of
images whose prediction changes, and therefore the drop of top-1, is at most
$\Pr_x[\KL_x\ge m(x)^2/2]\le\mathbb{E}_x[2\KL_x/m(x)^2]$.
\end{proposition}
\begin{proof}
Let $y$ be the class predicted by $p$, and suppose that $q=p_s(\cdot\mid x)$ predicts $y'\neq y$, so $q_{y'}\ge q_y$. Then
$m\le p_y-p_{y'}\le(p_y-q_y)+(q_{y'}-p_{y'})\le|p_y-q_y|+|p_{y'}-q_{y'}|\le2\,\mathrm{TV}(p,q)\le\sqrt{2\KL(p\,\|\,q)}$ by Pinsker's
inequality, so $\KL\ge m^2/2$. The bound on the share follows from Markov's inequality applied to $2\KL_x/m(x)^2$. Top-1
drops only on images whose prediction changes.
\end{proof}
\noindent\emph{Check.} Images with a small margin dominate the bound. The mean KL weights all images equally, yet it
predicts the top-1 drop well. Over the 171 final schedules of the three settings that lose more than 0.02\pp, the
log-log correlation between the KL on held-out training images and the top-1 drop on \IN is 0.96--0.99 per model, with
exponents of 0.75--1.13 (\cref{fig:supp:kl}).

\begin{proposition}[Accuracy change]
\label{prop:accuracy}
Let the unreduced model be correct on a share $a\in(0,1)$ of the images, and let a method fix a share $f$ of its errors
and break a share $b$ of its hits. The accuracy changes by $(1-a)f-ab$. For $b>0$, this change is positive iff
$f/b>a/(1-a)$.
\end{proposition}
\begin{proof}
The method is correct on the $a(1-b)$ hits it keeps and the $(1-a)f$ errors it fixes, so its accuracy is
$a-ab+(1-a)f$.
\end{proof}

\paragraph{\PPS.}
\label{sec:supp:proofs}
\begin{proposition}[What \PPS keeps]
\label{prop:pps}
Let \PPS keep the sample $S$ with inclusion probabilities $\pi_j=\lambda q_j$ (\cref{M-eq:inclusion_prob}, none capped at 1). For
every function $g$ of the tokens and an offset $u$ drawn uniformly,
$\mathbb{E}_u\big[\sum_{j\in S}g(x_j)\big]=k\big[\alpha\,\tfrac1n\sum_jg(x_j)+(1-\alpha)\sum_j\mathrm{softmax}(z)_jg(x_j)\big]$.
On average, the kept tokens are thus a mixture of all tokens, with weight $\alpha$, and of tokens tilted towards a large
size $z$. With Horvitz--Thompson weights $1/\pi_j$, the sum $\sum_{j\in S}g(x_j)/\pi_j$ is unbiased for
$\sum_jg(x_j)$. Under Poisson sampling, its variance is at most $(n/(k\alpha)-1)\sum_jg(x_j)^2$, compared with
$(n/k-1)\sum_jg(x_j)^2$ for a uniform sample.
\end{proposition}
\begin{proof}
We have $\mathbb{E}[\sum_{j\in S}g(x_j)]=\sum_j\pi_jg(x_j)$, and without capping $\lambda=k$ because $\sum_jq_j=1$;
substituting $q_j$ gives the identity. The unbiasedness of the weighted sum follows from the same identity. Under
Poisson sampling the inclusions are independent, so the variance is
$\sum_j\pi_j(1-\pi_j)g(x_j)^2/\pi_j^2=\sum_j(1/\pi_j-1)g(x_j)^2$, and $\pi_j\ge k\alpha/n$ by the coverage bound below.
\end{proof}
\noindent The identity needs $\pi_j<1$. In practice some large tokens are capped at $\pi_j=1$, so the identity holds
only approximately. The variance bound holds for Poisson sampling; our systematic sweep has a variance that depends on
the order of the tokens. The bound also needs the reweighting, which costs 2.2, 1.1 and 0.4\pp of top-1 on DeiT-S,
DeiT-B and MAE ViT-B (\cref{M-sec:operators}). We therefore use \PPS without weights. It deliberately keeps the
tilted mixture, and the defensive term (the uniform component of weight $\alpha$) and the sweep along the Morton curve still
guarantee coverage (see below).

\paragraph{Coverage of \PPS.}
Let $q_j\ge\alpha/n$ with $\sum_j q_j=1$, and let $\pi_j=\min(1,\lambda q_j)$ with $\sum_j\pi_j=k\le n$ (\cref{M-eq:pps_distribution,M-eq:inclusion_prob}).
Since $\min(1,\lambda q_j)\le \lambda q_j$, we have $k=\sum_j\pi_j\le \lambda$.
Hence $\pi_j\ge\min(1,k\,q_j)\ge\min(1,k\alpha/n)=k\alpha/n$, because $k\alpha/n\le1$.

\paragraph{Systematic selection.}
List the tokens along the Morton curve (the Z-order curve over the patch grid), and let $\Pi_j=\sum_{i\le j}\pi_i$, so
that $\Pi_0=0$ and $\Pi_n=k$.
Given an offset $u\in(0,1)$, we keep token $j$ if the cell $(\Pi_{j-1},\Pi_j]$ contains a point $u+t$ with an integer $t$.
The points $u,u+1,\dots,u+k-1$ lie in $(0,k]$, which the cells partition, so exactly $k$ tokens are kept.
A cell has length $\pi_j\le1$, so it holds at most one point; for a uniform $u$, it holds one with probability $\pi_j$.
In practice, $u\in[0.01,0.99)$ is a hash of the sizes $z$ (\cref{sec:supp:impl}), which makes the forward pass
deterministic; the probabilities above are over a uniform $u$.
The sweep also covers the image for every offset: $h$ tokens that are consecutive along the curve span an interval of
length $\sum_j\pi_j\ge hk\alpha/n$, which contains at least $\lfloor hk\alpha/n\rfloor$ of the points. A complete aligned
$4\times4$ square of patches is contiguous along the curve. So when 127 of 196 patches are kept, every such square keeps
at least 3 of its 16.


\end{document}